%% file: main.tex
\documentclass[11pt,letterpaper]{article}
    \usepackage[margin=1in]{geometry}
    \usepackage{times}
\usepackage{mathpazo}

\usepackage{amsmath, amssymb}
\usepackage{amsthm}
\usepackage{mathtools}
\usepackage{nicefrac}
\usepackage{fullpage}

    \usepackage{algorithm}
    \usepackage{algpseudocode}[1]

\usepackage[utf8]{inputenc}
\usepackage[T1]{fontenc} %
\usepackage[dvipsnames]{xcolor}
\usepackage{color} 
\usepackage{cancel}
\usepackage{mathtools}

\usepackage{manfnt} %

\usepackage{figchild}
\usepackage{fontawesome5}

    \usepackage{accents}

\definecolor{niceRed}{RGB}{190,38,38}
\definecolor{niceYellow}{HTML}{f5b400}
\definecolor{blueGrotto}{HTML}{059DC0}
\definecolor{royalBlue}{HTML}{057DCD}
\definecolor{navyBlue}{HTML}{0B579C}
\definecolor{yaleBlue}{HTML}{00356b}
\definecolor{limeGreen}{HTML}{81B622}
\definecolor{nicePurple}{HTML}{9c27b0}
\definecolor{lightRoyalBlue}{HTML}{def2ff}  
\definecolor{gold}{HTML}{ffa300}

\usepackage{color-edits} 
\addauthor[Anay]{am}{gold}
\addauthor[Alkis]{ak}{blue}
\addauthor[Felix]{fz}{Orchid}
\addauthor[GPT]{gpt}{purple}
\addauthor[Claude]{claude}{limeGreen}
\addauthor[Codex]{codex}{niceRed}

\input{macros}

    \usepackage{hyperref}
    \hypersetup{
      colorlinks = true,
      urlcolor = {blueGrotto},
      linkcolor = {royalBlue},
      citecolor = {navyBlue},
      hypertexnames = false, %
    }
    \usepackage[
      backref=true,
      backend=biber,
      natbib=true,
      style=alphabetic,
      sorting=alphabeticlabel,
      sortcites=true,
      minbibnames=3,
      maxbibnames=999,
      mincitenames=4,
      maxcitenames=4,
      minalphanames=4,
      maxalphanames=4,
      doi=false, url=false, eprint=false, isbn=false, 
    ]{biblatex}

    \DeclareSortingTemplate{alphabeticlabel}{
      \sort[final]{%
        \field{labelalpha} %
      }
      \sort{%
        \field{year}   %
      }
      \sort{%
        \field{title}  %
      }
    }
    
    \AtBeginRefsection{\GenRefcontextData{sorting=ynt}}
    \AtEveryCite{\localrefcontext[sorting=ynt]}
\usepackage[most]{tcolorbox}
    
\tcbset{
    mybox/.style={
        colframe=black,        %
        colback=gray!0,       %
        boxrule=0.5mm,         %
        width=\linewidth,      %
        arc=3mm,               %
        left=5mm,              %
        right=5mm,             %
        top=5mm,               %
        bottom=5mm,            %
        enhanced,              %
        halign=center        %
    }
} 

\usepackage{booktabs} 
\usepackage{multicol} 
\usepackage{multirow} 
\usepackage{makecell} 
\usepackage{longtable} 

\usepackage{graphicx}
\usepackage{float} 
\usepackage{tikz}
\usepackage{pgfplots}
\pgfplotsset{compat=1.17}
\usepgfplotslibrary{fillbetween}
\usetikzlibrary{arrows.meta}
\usepackage{setspace}

\tikzset{
  myNodeFlex/.style={
    draw,
    rectangle,
    rounded corners,
    text centered,
    minimum height=1.5em,
  }
}

\tikzset{
  myNode/.style={
    draw,
    rectangle,
    rounded corners,
    text centered,
    minimum height=1.5em,
    minimum width=3cm,
    text width=5cm,    
  }
}

\tikzset{
  myNodeNarrow/.style={
    draw,
    rectangle,
    rounded corners,
    text centered,
    minimum height=1.5em,
    minimum width=1cm,
  }
}

\tikzset{
  myNodeWide/.style={
    draw,
    rectangle,
    rounded corners,
    text centered,
    minimum height=1.5em,
    minimum width=6cm,
  }
}

\usepackage{xinttools} %
\usetikzlibrary{calc}

\def\biglen{20cm} %
\tikzset{
  half plane/.style={ to path={
       ($(\tikztostart)!.5!(\tikztotarget)!#1!(\tikztotarget)!\biglen!90:(\tikztotarget)$)
    -- ($(\tikztostart)!.5!(\tikztotarget)!#1!(\tikztotarget)!\biglen!-90:(\tikztotarget)$)
    -- ([turn]0,2*\biglen) -- ([turn]0,2*\biglen) -- cycle}},
  half plane/.default={1pt}
}

\usepackage{physics} %
\usepackage{amsmath, amssymb}
\usepackage{amsthm}
\usepackage{bbm}
\usepackage{blkarray}
\usepackage{mathtools}
\usepackage{nicefrac}
\usepackage{dsfont}
\usepackage{xspace}
\usepackage{pifont} %

\usepackage{thm-restate} %
    \usepackage[capitalise,noabbrev,nameinlink,sort]{cleveref}

\usepackage{enumerate} 
\usepackage[inline,shortlabels]{enumitem} 
\usepackage{typed-checklist}
\usepackage[framemethod=TikZ]{mdframed}
\mdfsetup{%
backgroundcolor=white, %
roundcorner=4pt,
linewidth=1pt}
\usepackage{changepage} %

\usepackage{algorithm}
\usepackage{algpseudocode}[1]

\usepackage{mathrsfs} %
\usepackage{soul} %

\theoremstyle{plain} 
\newtheorem{theorem}{Theorem}[section]

\newtheorem{proposition}[theorem]{Proposition}
\newtheorem{lemma}[theorem]{Lemma}

\newtheorem{fact}[theorem]{Fact}

\newtheorem{inftheorem}{Informal Theorem}

\newtheorem{definition}{Definition}
\newtheorem*{definition*}{Definition}

\theoremstyle{definition} 

\newtheorem{remark}[theorem]{Remark}

\theoremstyle{remark}

\NewDocumentEnvironment{pf}{o}
  {\IfNoValueTF{#1}{\begin{proof}}{\begin{proof}[Proof of #1.]}}
  {\IfNoValueTF{#1}{\end{proof}}{\end{proof}}}

\AfterEndEnvironment{definition}{\noindent\ignorespaces}
\AfterEndEnvironment{assumption}{\noindent\ignorespaces}
\AfterEndEnvironment{lemma}{\noindent\ignorespaces}
\AfterEndEnvironment{theorem}{\noindent\ignorespaces}
\AfterEndEnvironment{proposition}{\noindent\ignorespaces}
\AfterEndEnvironment{fact}{\noindent\ignorespaces}
\AfterEndEnvironment{question}{\noindent\ignorespaces}
\AfterEndEnvironment{corollary}{\noindent\ignorespaces}
\AfterEndEnvironment{model}{\noindent\ignorespaces}
\AfterEndEnvironment{remark}{\noindent\ignorespaces}
\AfterEndEnvironment{proof}{\noindent\ignorespaces}
\AfterEndEnvironment{fact}{\noindent\ignorespaces}
\AfterEndEnvironment{minftheorem}{\noindent\ignorespaces}
\AfterEndEnvironment{inftheorem}{\noindent\ignorespaces}
\AfterEndEnvironment{maintheorem}{\noindent\ignorespaces}
\AfterEndEnvironment{restatable}{\noindent\ignorespaces}
\AfterEndEnvironment{infassumption}{\noindent\ignorespaces}
\AfterEndEnvironment{conjecture}{\noindent\ignorespaces}
\AfterEndEnvironment{claim}{\noindent\ignorespaces}

\crefname{section}{Section}{Sections}
\crefname{theorem}{Theorem}{Theorems}
\crefname{lemma}{Lemma}{Lemmas}
\crefname{definition}{Definition}{Definitions}
\crefname{conjecture}{Conjecture}{Conjectures}
\crefname{corollary}{Corollary}{Corollaries}
\crefname{construction}{Construction}{Constructions}
\crefname{claim}{Claim}{Claims}
\crefname{observation}{Observation}{Observations}
\crefname{proposition}{Proposition}{Propositions}
\crefname{fact}{Fact}{Facts}
\crefname{question}{Question}{Questions}
\crefname{problem}{Problem}{Problems}
\crefname{remark}{Remark}{Remarks}
\crefname{example}{Example}{Examples}
\crefname{equation}{Equation}{Equations}
\crefname{appendix}{Appendix}{Appendices}
\crefname{algorithm}{Algorithm}{Algorithms}
\crefname{model}{Model}{Models}
\crefname{figure}{Figure}{Figures}
\crefname{inftheorem}{Informal Theorem}{Informal Theorems}
\crefname{infassumption}{Informal Assumption}{Informal Assumptions}
\crefname{minftheorem}{Main Informal Theorem}{Main Informal Theorems}
\crefname{maintheorem}{Main Theorem}{Main Theorems}
\crefname{assumption}{Assumption}{Assumptions}
\crefname{case}{Case}{Cases}
\crefname{program}{Program}{Programs}
\crefname{inequality}{Inequality}{Inequalities}
\crefname{subsubsubappendix}{Appendix}{Appendices}

\newlist{asmpenum}{enumerate}{1} %
\setlist[asmpenum]{label={\arabic*.},ref=\theassumption.{\arabic*}}
\crefname{asmpenumi}{Assumption}{Assumptions}

\usepackage{etoolbox}

\newcommand{\yesnum}{\addtocounter{equation}{1}\tag{\theequation}}  

\makeatletter
\renewcommand{\eqref}[1]{\textup{\eqrefform@{\ref{#1}}}}
\let\eqrefform@\tagform@

\newcommand{\changetag}[1]{%
  \renewcommand\tagform@[1]{\maketag@@@{(\ignorespaces#1\unskip\@@italiccorr)}}%
}
\makeatother

\makeatletter
\newcommand{\tagnum}[2]{%
    \refstepcounter{equation}%
    \tag{#1) \ (\theequation}%
    \protected@write \@auxout {}{%
        \string \newlabel {#2}{{\theequation}{\thepage}{}{equation.\theequation}{}}%
    }%
}
\makeatother

\newcommand{\Stackrel}[2]{\stackrel{\mathmakebox[\widthof{\ensuremath{#2}}]{#1}}{#2}}

\newcommand{\quadtext}[1]{\quad\text{#1}\quad}
\newcommand{\qquadtext}[1]{\qquad\text{#1}\qquad}
 
\newcommand{\quadand}{\quadtext{and}}
\newcommand{\qquadand}{\qquadtext{and}}

\newcommand{\quadwhere}{\quadtext{where}}

\def\abs#1{\left| #1 \right|}
\def\biggabs#1{\bigg| #1 \bigg|}
\def\sabs#1{| #1 |}

\newcommand{\sinparen}[1]{\ensuremath{(#1)}}
\newcommand{\sinbrace}[1]{\ensuremath{\{#1\}}}
\newcommand{\sinsquare}[1]{\ensuremath{[#1]}}
\newcommand{\inbrace}[1]{\ensuremath{\left\{#1\right\}}}
\newcommand{\sinangle}[1]{\ensuremath{\langle#1\rangle}}
\newcommand{\inparen}[1]{\ensuremath{\left(#1\right)}}
\newcommand{\insquare}[1]{\ensuremath{\left[#1\right]}}
\newcommand{\inangle}[1]{\left\langle#1\right\rangle}

\newcommand{\ceil}[1]{\ensuremath{\left\lceil#1\right\rceil}}

\newcommand{\snorm}[1]{\ensuremath{\ensuremath{\| #1 \|}}}
\let\norm\relax
\newcommand{\norm}[1]{\ensuremath{\left\lVert #1 \right\rVert}}

\newcommand{\N}{\mathbb{N}}
\newcommand{\R}{\mathbb{R}}

\newcommand{\evE}{\ensuremath{\mathscr{E}}}

\newcommand{\E}{\operatornamewithlimits{\mathbb{E}}} 
\newcommand{\Ex}{\E}
\newcommand{\Exp}{\Ex}
\newcommand{\Cov}{\ensuremath{\operatornamewithlimits{\rm Cov}}}
\newcommand{\cov}{\ensuremath{\operatornamewithlimits{\rm Cov}}}
\let\var\relax
\newcommand{\var}{\ensuremath{\operatornamewithlimits{\rm Var}}}

\newcommand{\sgn}{\operatornamewithlimits{sgn}}
\newcommand{\argmin}{\operatornamewithlimits{arg\,min}}
\newcommand{\argmax}{\operatornamewithlimits{arg\,max}}

\newcommand{\tv}[2]{\operatorname{d}_{\mathsf{TV}}{\inparen{#1,#2}}}
\newcommand{\kl}[2]{\operatornamewithlimits{\mathsf{KL}}{\inparen{#1\|#2}}}

\newcommand{\sfrac}[2]{{#1/#2}} 
\newcommand{\nfrac}[2]{\nicefrac{#1}{#2}}

\newcommand{\poly}{\mathrm{poly}}
\newcommand{\polylog}{\mathrm{polylog}}

\newcommand{\iid}{i.i.d.}

\newcommand{\eps}{\varepsilon}
\renewcommand{\epsilon}{\varepsilon}
\makeatletter
\newcommand*{\tran}{{\mathpalette\@tran{}}}
\newcommand*{\@tran}[2]{\raisebox{\depth}{$\m@th#1\intercal$}}
\makeatother

\mathchardef\NABLA"272
\newcommand*{\Nabla}{\boldsymbol\NABLA}
\let\nabla\Nabla

\renewcommand{\hat}{\widehat}
\newcommand{\wh}[1]{\widehat{#1}}

\renewcommand{\bar}{\overline}

\renewcommand{\tilde}{\widetilde}

\newcommand{\customcal}[1]{\euscr{#1}}

\newcommand{\cL}{\customcal{L}}
\newcommand{\cM}{\customcal{M}}
\newcommand{\cN}{\customcal{N}}

\newcommand{\cX}{\customcal{X}}
\newcommand{\cY}{\customcal{Y}}

\DeclareMathAlphabet{\mathdutchcal}{U}{dutchcal}{m}{n}
\SetMathAlphabet{\mathdutchcal}{bold}{U}{dutchcal}{b}{n}
\DeclareMathAlphabet{\mathdutchbcal}{U}{dutchcal}{b}{n}

\DeclareMathAlphabet\urwscr{U}{urwchancal}{b}{n}%
\DeclareMathAlphabet\rsfscr{U}{rsfso}{m}{n}
\DeclareMathAlphabet\euscr{U}{eus}{m}{n}
\DeclareFontEncoding{LS2}{}{}
\DeclareFontSubstitution{LS2}{stix}{m}{n}
\DeclareMathAlphabet\stixcal{LS2}{stixcal}{m} {n}

\newcommand{\eg}{\emph{e.g.}}
\newcommand{\ie}{\emph{i.e.}}

\usepackage{upgreek}
\renewcommand{\gamma}{\upgamma}
\renewcommand{\pi}{\uppi}

\newcommand{\eat}[1]{}
\newcommand{\negLL}{\ensuremath{\mathscr{L}}}

\newcommand{\hypo}[1]{\mathdutchcal{#1}}

\newcommand{\hyL}{\hypo{L}}

\newcommand{\hyP}{\hypo{P}}

\renewcommand{\d}{{\rm d}}

\newcommand{\normal}[2]{\cN{\inparen{#1,#2}}}
\newcommand{\coarseNormal}[3]{\cN_{#3}\inparen{#1,#2}}
\newcommand{\truncatedNormal}[3]{\cN\inparen{#1,#2,#3}}
\newcommand{\normalMass}[3]{\cN\inparen{#1, #2; #3}}

\newcommand{\wstar}{{w^\star}}
\newcommand{\Wstar}{{W^\star}}

\newcommand{\muStar}{{\mu^\star}}
\newcommand{\SigmaStar}{{\Sigma^\star}}

\usepackage{caption}
\usepackage{subcaption}
\usepackage{graphicx}
\usepackage{wrapfig}

\newcommand{\ymax}{y_{\max}}

\newcommand{\customcdf}[2]{\Phi\sinparen{#1; \sigma_{#2}^2}}

\renewcommand{\cL}{\negLL}
\renewcommand{\hyL}{\negLL}

\usepackage{array}
\newcolumntype{L}[1]{>{\raggedright\let\newline\\\arraybackslash\hspace{0pt}}m{#1}}
\newcolumntype{C}[1]{>{\centering\let\newline\\\arraybackslash\hspace{0pt}}m{#1}}
\newcolumntype{R}[1]{>{\raggedleft\let\newline\\\arraybackslash\hspace{0pt}}m{#1}}

\usepackage{fancyhdr}

\fancypagestyle{custom}{
  \fancyhf{}                %
  \fancyfoot[C]{\textcolor{black}{\fcFishD{0.025}{black}{0.5}}} %

}

\title{
    Can SGD Select Good Fishermen?\\
    Local Convergence under Self-Selection {Biases}
    }  
    
    \author{
            \begin{tabular}{C{5.1cm}C{5.1cm}C{5.1cm}}
            {\bf Alkis Kalavasis} 
                & {\bf Anay Mehrotra} 
                    & {\bf Felix Zhou}\\[2mm]
            {Yale University} 
                & {Yale University} 
                    & {Yale University}\\[1mm]
            \mbox{\small\texttt{\href{mailto:alkis.kalavasis@yale.edu}{alkis.kalavasis@yale.edu}}} 
                & \mbox{\small\texttt{\href{mailto:anaymehrotra1@gmail.com}{anaymehrotra1@gmail.com}}}
                    & \mbox{\small\texttt{\href{felix.zhou@yale.edu}{felix.zhou@yale.edu}}}
            \\
            \end{tabular}
    }

    \date{}

\begin{document}

\maketitle
 \thispagestyle{empty} 

\begin{abstract}
    We revisit the problem of estimating $k$ linear regressors with self-selection bias in $d$ dimensions with the maximum selection criterion, as introduced by \citet[STOC'23]{cherapanamjeri2023selfselection}. 
    Our main result  
    is a $\poly(d,k,\nfrac{1}{\eps}) + (k\log k)^{O(k)}$ time algorithm for this problem 
    {that improves upon the} running time of the algorithms {by} \citet{cherapanamjeri2023selfselection} and \citet[arXiv]{gaitonde2024selfselection}. We achieve this by providing the first local convergence algorithm for self-selection, thus {resolving one of the main open questions of \citet{cherapanamjeri2023selfselection}.}

    To obtain this algorithm, we 
    reduce self-selection to a seemingly unrelated statistical problem called estimation under coarsening \citep[COLT'21]{fotakis2021coarse}. Coarsening occurs when one does not observe the exact value of the sample but only some
    set (from a partition of the sample space) {containing} the exact value. 
    Inference from coarse samples arises in various real-world applications, including rounding by humans and algorithms, limited precision of instruments, and lag in multi-agent systems.
    The coarse estimation problem arising in our reduction is induced by a non-convex partition, whereas previous works on coarsening exclusively studied convex partitions.
    The resulting estimation algorithm relies on the geometry of the self-selection problem to bypass non-convexity. This geometric approach, in turn, enables us to overcome the limitations of previous analytic approaches and could have applications for designing efficient algorithms for other latent-variable problems.
\end{abstract}

\newpage
    \setcounter{tocdepth}{3}
    {   
        \setstretch{1}
        \tableofcontents
    }
    \thispagestyle{empty}
    \newpage
    
    \pagenumbering{arabic}

\section{Introduction}
    Self-selection bias occurs when data is systematically selected rather than randomly sampled.
    Inference under selection biases has a rich history in Statistics and Econometrics, starting with the foundational works of \citet{roy1951earnings,willis1979education,heckman1979sample}.
    {This framework has since been applied in} various scientific fields, from causal inference and imitation learning \cite{heckman1990varieties}, to learning from strategically reported data \cite{hardt2016strategic,dong2018strategic,kleinberg2020classifiers}, and to learning from auction data \cite{athey2002identification,athey2007nonparametric,cherapanamjeri2022auctionEstimation}.
    
    A concrete application of self-selection bias appears in the work of \citet{fair1972methods}, who studied estimation in markets at disequilibrium, where supply does not match demand.
    Fair and Jaffee modeled the housing market, with $y_S(x)$ representing supply and $y_D(x)$ {denoting} demand, as {functions} of {features} $x$ encoding, \eg{}, location, size, and amenities. 
    For a given $x$, {only the transaction price} $(x,\min\{y_S(x), y_D(x)\})$ {is observed} instead of the complete sample $(x,y_S(x),y_D(x))$, 
    leaving it unclear whether the market imbalance stems from excess supply or demand.

    Another classic model due to \citet{roy1951earnings} examines workers' occupational choices based on potential earnings.
    Suppose there are $k$ occupations (\eg{}, hunting, fishing, and woodcutting) and, for each occupation $i$, the expected income for workers with feature vector $x$ is given by $y_i(x)=x^\top w_i^\star + \xi_i$, 
    where $w_i^\star$ is a parameter vector and $\xi_i \sim \normal{0}{1}$ {captures} the sum of independent market events related to the $i$-th occupation.
    {Workers} select their occupations \textit{strategically} to maximize revenue, considering all potential incomes $y_1(x), y_2(x),\dots, y_k(x)$, and selecting the occupation that offers the highest income.
    From the analyst's perspective, however,
    only the feature vector $x$ and the maximum income achieved $\max \{y_1(x), y_2(x), \dots, y_k(x)\}$ are observed.
    The goal is then to estimate the unknown parameters $w^\star_1,w^\star_2,\dots,w^\star_k$ {which determine the expected income from becoming a hunter, fisherman, woodcutter, and so on.}

    These examples, as many others, are cases of self-selection bias with the \textit{maximum} selection criteria.\footnote{{The minimum in Fair-Jaffee's model can be converted to a maximum by negating the parameter vectors and noise.}}  
    Since the seminal works of \citet{roy1951earnings,fair1972methods}, 
    estimation under self-selection bias has received significant attention from studies of participation in the labor force \cite{heckman1974shadow,hanoch1976multivariate,NELSON1977309,heckman1979sample,Hanoch1980,Cogan1980}, 
    to migration and income \cite{nakosteen1980migration,borjas1987selfSelection}, 
    to effects of unions on wages \cite{lee1978union,abowd1982job}, 
    to returns to education \cite{willis1979education,kenny1979college,griliches1978missing}, 
    and to choices between tenure choice and demand for housing \cite{lee1978estimation,king1980tenure,rosen1979housing}; see \cite{maddala1986limited} for more in-depth {discussion}. 
    {Despite this extensive history, efficient algorithms for estimation under self-selection -- even for natural selectors like $ \min $ and $ \max $ -- were not known until \citet{cherapanamjeri2023selfselection} initiated the design of sample- and computationally-efficient linear regression under self-selection biases.}
    Let us first present the model they studied: linear regression with maximum self-selection bias.
    \begin{restatable}[Max Self-Selection with Unknown Index; \citet{cherapanamjeri2023selfselection}]{definition}{defMaxSelfSelection}
    \label{def:ssb}
    In linear regression with self-selection bias under the maximum selection criterion, a sample $(x,\ymax) \in \R^d \times \R$ is generated as follows:
    \begin{enumerate} %
        \item {The} covariate $x\in \R^d$ is drawn from $\normal{0}{I_d}$, and 
        \item $\ymax$ is the maximum of $k$ different unknown linear functions of $x$ that are independently perturbed by noise, \ie{}, $\ymax = \max_{i \in [k]} \sinbrace{x^\top w_i^\star + \xi_i}$, where $w_1^\star,\dots,w_k^\star$ are the unknown target parameters and $\xi_1,\dots,\xi_k$ are independent $\normal{0}{1}$ random variables.
    \end{enumerate}
    The unknown parameters $w_1^\star,\dots,w_k^\star$ satisfy the following conditions for some constants $\nfrac{1}{c}, C \geq 1$:
    \begin{enumerate} 
        \item[(a)] (Separability)~~  $ \norm{w^\star_i}_2^2 
            \geq c + \max_{j\neq i}~\sabs{\sinangle{w^\star_j, w^\star_i}}.$
        \item[(b)] (Boundedness) ~~$\max_{1\leq i\leq k}\norm{w^\star_i}_2\leq C\,.$
    \end{enumerate}
    When $k=1$, the maximum over the empty set in condition (a) is understood to be zero.
    \end{restatable}
    Both conditions also appear in prior works  \cite{cherapanamjeri2023selfselection,gaitonde2024selfselection}.~Without condition (a), identifying the parameters is information-theoretically impossible \citep[Remark 6.1]{gaitonde2024selfselection} and {condition (b) is a standard boundedness assumption.}
    
    This model captures the aforementioned classical works of \citet{roy1951earnings} and \citet{fair1972methods}.
    Remarkably, \cite{cherapanamjeri2023selfselection} presented an algorithm to estimate regression parameters $w^\star_1,\dots,w^\star_k$ to within $\poly(\nfrac{1}{k})$-error in $\poly(d)\cdot e^{\poly(k)}$ time under \Cref{def:ssb}.
    This result is surprising since even the \emph{identifiability} of the parameters under \Cref{def:ssb} is non-trivial.
    Subsequently, follow-up work by \citet{gaitonde2024selfselection} significantly improved the sample complexity {to the near optimal:} they recover the regression parameters {$w_1,\dots,w_k$ such that the} {maximum $L_2$-}error{, $\max_i \norm{w_i - w^\star_i}_2^2$, is at most} $\eps^2$ with $\widetilde O(d) \cdot \poly(k, \nfrac{1}{\eps})$ samples in {$\poly(d,k,\nfrac{1}{\eps}) + O(\nfrac{(\log{k})}{\eps})^{O(k)}$} time.
    {Both algorithms first identify a $k$-dimensional subspace containing $w_1^\star, \dots, w_k^\star$ by computing the span of the top eigenvectors of some weighted covariance matrix. They then perform a brute-force search within this subspace to identify a regressor $w_i$ (and then iterate to find the remaining regressors). Unfortunately, this brute-force step necessarily introduces an $\eps^{-k}$ dependence in the running time. 
        }

    \subsection{Our Contributions}
    Our main result is a faster algorithm that avoids the $\eps^{-k}$ dependence.  
    \begin{restatable}[Efficient Regression Under Self-Selection]{theorem}{thmSelfSelectionMain}\label{thm:SS}
            \label{infthm:Intro1}
            \label{infthm:Intro:SS}
            Under \Cref{def:ssb}, there is an algorithm that, 
                given $\eps,\delta\in (0,1)$ 
                and {$n=\widetilde O(d) \cdot \poly(k,\nfrac{1}{\eps}, \log(\nicefrac1\delta))$} samples generated by the max-self-selection model with parameters $w^\star_1,w^\star_2,\dots,w^\star_k$,
            outputs a set of estimates $\sinbrace{w_1,w_2,\dots,w_k}$,
            such that, 
            with probability $1-\delta,$
            there is a re-ordering\footnote{{As in previous work, we recover the weights up to permutation, which is information-theoretically optimal.}} of estimates satisfying 
            $\sum_i \norm{w_i-w_i^\star}_2^2\leq \eps^2$.
            The algorithm runs in {$\poly(d,k,\nfrac{1}{\eps},{\log(\nicefrac1\delta)}) + \sinparen{k\log{k}}^{O(k)}$} time.
        \end{restatable}
    This is the first $\poly(d,\nfrac{1}{\eps}) \cdot \sinparen{k\log{k}}^{O(k)}$ time algorithm for the self-selection under maximum selection criterion (henceforth, simply \textit{self-selection}).
    It is based on the first local convergence algorithm for self-selection which also resolves an open problem by \citet{cherapanamjeri2023selfselection}.
    \begin{restatable}[Polynomial Time Local Convergence for Self-Selection]{theorem}{thmSelfSelectionLocalConvergence}
        \label{thm:SS:localConvergence}
        \label{infthm:Intro:SS:localConvergence}
            Under \Cref{def:ssb}, there is an algorithm that, 
            given $\eps, \delta\in (0, 1)$,
                a $D_{\rm SS}$-warm start, where $D_{\rm SS}=(c/(ek))^{O((C/c)^{18})}$ is specified in \cref{eq:SS:DSS},
                and a fresh sample batch, independent of the supplied warm start, of size
                $
                    n=\tilde O{\inparen{\nfrac{d}{\eps^2}\cdot {\log{\nicefrac1\delta}}}}\cdot \poly(k)
                $
                generated by the max-self-selection model with parameters $w^\star_1,w^\star_2,\dots,w^\star_k$,
            outputs a set of estimates $\sinbrace{w_1, w_2, \dots, w_k}$,
            such that, with probability $1-\delta,$ there is a re-ordering of estimates satisfying 
            $\sum_i \norm{w_i-w_i^\star}_2^2\leq \eps^2$.
            The algorithm runs in {$\tilde O{\inparen{\nfrac{d^2}{\eps^2}\cdot {\log{\nicefrac1\delta}}}}\cdot \poly(k)$} time.
        \end{restatable}
        Combining \Cref{thm:SS:localConvergence} with the algorithm of \citet{gaitonde2024selfselection} as a $D_{\rm SS}$-warm start yields the end-to-end guarantee of \Cref{thm:SS}; in particular, \Cref{thm:SS} rests on the local-convergence result, which itself requires a $D_{\rm SS}$-warm start.

        Our approach to {proving} \cref{infthm:Intro:SS,infthm:Intro:SS:localConvergence}  is geometric, rooted in a reduction to a classical statistical problem known as \textit{coarsening} \cite{heitjan1990inference} and complements the important analytic techniques of prior works \cite{cherapanamjeri2023selfselection,gaitonde2024selfselection} 
        by offering a new perspective on the self-selection problem.

        Our idea is extensively described in \Cref{sec:technicalOverview}.
        We believe that our framework will find applications in other problems with exogenous and endogenous selection biases. 
\begin{remark}[Error Metric]\label{rem:error:maximumVsJoint}
    Let $W^\star = [w_1^\star,\dots,w_k^\star]\in \R^{d\times k}$ and $W = [w_1,\dots,w_k]\in \R^{d\times k}$ denote the estimates.
    Since $x\sim \cN(0,I_d)$, $\Exp_x \sinsquare{\snorm{(W^\star-W)^\top x}_2^2} = \snorm{W^\star-W}_F^2$.
    For our algorithms,it is more natural to measure this \textit{joint} $L_2$-error, \ie{}, $\sum_i \norm{w_i-w_i^\star}_2^2 = \norm{W^\star-W}_F^2$,
    whereas prior works use the \emph{maximum} $L_2$-error, \ie{}, $\max_i \norm{w_i-w_i^\star}_2^2$.
    Requiring the stronger $L_2$-error guarantee increases the run time of \citet{gaitonde2024selfselection} to $\poly(d,k,\nfrac{1}{\eps}) + O(\nfrac{(k\log{k})}{\eps})^{O(k)}$.
    Thus, for the joint $L_2$-error metric, we strictly improve their running time.
\end{remark}

\paragraph{Organization of the rest of the paper.}
We present the coarsening problem in \Cref{section:tool} and discuss the related tools.
Translating self-selection to the language of coarsening unlocks these tools.
Specifically, assuming two key properties: \emph{information preservation} (\Cref{def:informationPreservation}) and \emph{local convexity}, this translation allows us to design the SGD-based local convergence algorithm from \Cref{infthm:Intro:SS:localConvergence}.
The main technical challenge is proving that the two properties hold, {which boils down to understanding the underlying optimization landscape}.
We overview our approach in \Cref{sec:Overview}.
We discuss related works in \Cref{sec:relatedWorks} and \Cref{sec:conclusion} concludes with takeaways and an open question.

\section{Technical Overview}\label{sec:technicalOverview}
\subsection{Coarsening}\label{section:tool}
Our main conceptual contribution is a reduction of the self-selection problem to a seemingly unrelated statistical problem known as \emph{coarsening.}
In this section,  we introduce learning from coarse data in its simplest form and the tools we need for the reduction. 
This groundwork is important for \Cref{sec:Overview}, where we show how to {cast the self-selection problem as inference from coarse data and then} employ this framework to obtain faster algorithms for the self-selection problem.

Coarsening occurs when the exact value of a sample is not observed; instead, only a subset of the sample space containing said value is observed.
Coarse data naturally arises in Economics, Engineering, Medical and Biological Sciences, and all areas of the Physical Sciences 
(\eg{}, \citet{heitjan1990inference,heitjan1991ignorability,heitjan1993ignorability,gostic2020practical,leung1997censoring}). 

\begin{figure}[htbp]
        \centering
        \begin{subfigure}[t]{0.3\linewidth}
            \centering
            \includegraphics[width=\linewidth,]{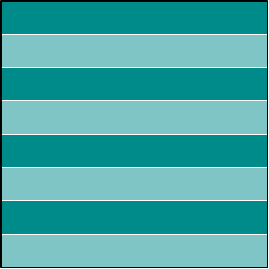}
        \end{subfigure}
        \hfill
        \begin{subfigure}[t]{0.3\linewidth}
            \centering
            \includegraphics[width=\linewidth]{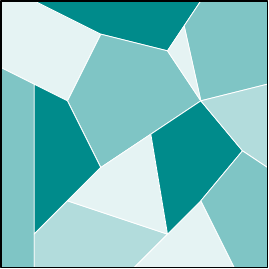}
        \end{subfigure}
        \hfill
        \begin{subfigure}[t]{0.3\linewidth}
            \centering
            \includegraphics[width=\linewidth]{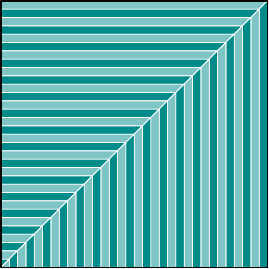}
        \end{subfigure}
        \caption{
            {The figure illustrates different partitions $\hyP$ of $\R^2$.}
            The figure on the left corresponds to a partition that is not identifiable: Given any $\mu^\star \in \R^2$, the vector $\mu_t = \mu^\star + t e_1$ {(for any $t\in \R$)} induces the same coarse Gaussian distribution and so $\coarseNormal{\mu^\star}{I}{\hyP}$ and $\coarseNormal{\mu^\star_t}{I}{\hyP}$ are identical.
            The middle and right figures are identifiable and correspond to convex partitions of the space.
        }
        \label{fig:convex-partitions}
    \end{figure}

The problem of estimation from coarse data, in its simplest form, can be described as follows. Consider the family of normal distributions $\{\normal{\mu}{I}\}_{\mu}$  over the $d$-dimensional Euclidean space $\R^d$ with identity covariance matrix. Suppose that a partition $\hyP$ divides $\R^d$ into sets and our goal is to estimate the unknown parameter $\muStar$ of the target distribution $\normal{\muStar}{I}$.
In the coarse learning problem, the  algorithm has access to an oracle that operates as follows:
\begin{center}
    On each query, the oracle samples $x\sim \normal{\mu^\star}{I}$ and returns the unique set $P$ in $\hyP$ containing $x$.\footnote{In our setting, the set $P$ is succinctly represented as a real number $y\in \R$.}
\end{center}
We denote the distribution on sets $P\in \hyP$ as $\coarseNormal{\mu^\star}{I}{\hyP}.$
Crucially, the {actual} point $x$ itself is not observed.
{The challenge is to design an algorithm that, given \iid{} (set) samples $P_1, \dots, P_n \sim \coarseNormal{\mu^\star}{I}{\hyP}$, accurately estimates $\mu^\star$.}
{We focus on algorithms with time- and sample-complexity polynomial in the dimension and the desired accuracy, achieving an $\ell_2$-norm approximation of the true mean.}
On one extreme, when $\hyP$ consists of singletons, 
{the problem reduces to} ``fine'' Gaussian mean estimation problem where simply computing the empirical average {suffices}.
On the other extreme, when $\hyP = \{\R^d\}$, \ie{}, the only set of the partition is the whole space, parameter recovery {becomes} information-theoretically impossible. 
For further examples, see \Cref{fig:convex-partitions}.

\subsubsection{Coarse Negative Log-Likelihood Objective.}
A classical approach to estimating parameters from coarse samples is to use the negative log-likelihood objective.
This idea is used in the foundational works on learning from coarsened data \cite{heitjan1990inference,heitjan1991ignorability} and has also been recently revisited~\cite{fotakis2021coarse}. 
Specifically, to estimate $\mu^\star$ from coarse samples, one can {minimize the} population negative log-likelihood objective (NLL)
\begin{equation}
      \cL(\mu) = -\E_{P \sim \coarseNormal{\mu^\star}{I}{\hyP}} \log \normalMass{\mu}{I}{P}\,. 
      \label{eq:NLL-intro}
\end{equation}
    Here, $\normalMass{\mu}{I}{P}$ denotes the mass of $P$ under the measure $\normal{\mu}{I}$.
  {As we} will see, {the} identifiability, sample complexity, and computational efficiency {of the problem} all depend on the structure of the sets in the partition $\hyP$.
  This was also observed by \citet{fotakis2021coarse},
  who discovered that (identifiable) \textit{convex} partitions are statistically ``easy.''
  Their proof of identification is based on showing that the log-likelihood function is strictly convex when the partition is convex. 
  On the other hand, even simple non-convex sets can induce computationally intractable instances.
  Our exploration further refines the boundary of tractability by showing that some \emph{non-convex} partitions, such as those resulting from self-selection (\Cref{sec:overview:SS:conversion}), can still be computationally tractable.

\subsubsection{Ingredient I: Information Preservation.}
\label{sec:tool1}
As we have seen, there are simple cases of coarse partitions where inferring the true mean is information-theoretically impossible (left partition in \cref{fig:convex-partitions}). 
\citet{fotakis2021coarse} introduced the notion of \emph{information preservation} to quantify when consistent estimation under coarsening is possible.
We provide a more general formulation of their definition.

\begin{definition}
[{Information Preservation}]
\label{def:informationPreservation}
            Let $\alpha\in [0,1]$. 
            Given a $d$-dimensional Gaussian distribution $\normal{\muStar}{\SigmaStar}$ (with unknown parameters)
            and a partition $\hyP$ of the domain $\R^d$, 
            $\hyP$ is said to be \emph{$\alpha$-information preserving 
            at radius $R>0$
            with respect to $\normal{\muStar}{\SigmaStar}$} if, 
            for any parameters $(\mu,\Sigma)$ 
            {that are $R$-close to $(\muStar,\SigmaStar)$ in $\ell_2$-norm},\footnote{For a matrix $A$, $\norm{A}_F$ denotes its {Frobenius} norm and is defined as $\norm{A}_F^2=\sum_i\sum_j A_{ij}^2$. Here, $(\mu,\Sigma)$ being $R$-close to $(\muStar,\SigmaStar)$ in $\ell_2$-norm means $\sinparen{\norm{\mu-\muStar}_2^2 + \norm{\Sigma-\SigmaStar}_F^2}^{1/2}\le R$.}
            it holds that
            \[
                \tv{
                    \coarseNormal{\mu}{\Sigma}{\hyP}
                }{  
                    \coarseNormal{\muStar}{\SigmaStar}{\hyP}
                }
                \geq 
                \min\sinbrace{
                    1,
                    \alpha \sinparen{
                        \norm{\mu - \muStar}_2^2 
                        +
                        \norm{\Sigma - \SigmaStar}_F^2
                    }^{1/2}
                }
                \,.\footnote{ Taking a minimum on the RHS is somewhat redundant since the TV distance is upper bounded by $1$. However, this serves as a useful notational reminder of this restriction when carrying out calculations.}
            \]
        \end{definition}
        Information preservation is parameterized by $\alpha \in [0,1]$ and $R>0$.
        Intuitively, large values of $\alpha$ correspond to partitions that preserve a lot of information about the original (non-coarsened) probability measures; 
        {while as} $\alpha \to 0$, {the coarse measures become nearly indistinguishable in total variation, making the problem statistically impossible}.
        Note that information preservation is defined with respect to the true parameters $(\mu^\star,\SigmaStar)$, and can vary significantly with different true parameters.
        {Moreover, the radius parameter $R$ imposes a local structure by allowing information to be preserved only in a ball of radius $R$ around the true parameters.}
      {This locality is essential; for example, in problems with permutation invariance} on the true parameters {(\eg{}, the self-selection problem of \Cref{def:ssb} or mixture models)},
      {permuting the true parameters leaves the induced coarse distribution, and hence the total-variation distance on the LHS of \Cref{def:informationPreservation}, unchanged at zero, while the RHS (the parameter distance) stays positive. So information can be preserved only locally}.
        
        Our starting observation is that information preservation {implies ``local growth''} for the negative log-likelihood (due to Pinsker's inequality).
       
    \begin{lemma}
    [Information Preservation Implies Quadratic Growth]
    \label{thm:quadGrowth}
    Consider a partition $\hyP$ of $\R^d$ that is $\alpha$-information preserving at radius $R$ with respect to $\normal{\mu^\star}{\SigmaStar}$. Then,
            \begin{equation}
                 \hyL(\mu, \Sigma)
            -
            \hyL(\mu^\star, \Sigma^\star) 
            \geq
            \min\sinbrace{
                2,
                2\alpha^2\sinparen{\|\mu - \mu^\star\|_2^2 + \|\Sigma - \Sigma^\star\|_{F}^2}
            }\,.
            \label{eq:QuadGrowth}
            \end{equation}
        {for $(\mu, \Sigma)$ that are $R$-close to $(\mu^\star, \SigmaStar)$ in $\ell_2$-norm.}
    \end{lemma}
    Thus, if $\alpha$-information preservation holds, then $\sinparen{\muStar,\SigmaStar}$ is the \textit{unique} minimizer of the negative log-likelihood {\textit{locally}}.
    {However, minimizing the negative log-likelihood can be challenging:}
    Since the distribution of coarse samples does not belong to an exponential family, the negative log-likelihood {is, in general,} non-convex.
    Hence, while the quadratic growth {property} is reminiscent of strong convexity, it does not imply global {or even local} convexity of the negative log-likelihood; in fact, quadratic growth {may hold} even when the NLL is highly non-convex
    (see \Cref{fig:quadratic-growth}).

\subsubsection{Ingredient II: Local Convexity of NLL.}
\label{sec:tool2}
{Assume for now} the partition $\hyP$ is $\alpha$-information preserving with respect to $(\muStar, \SigmaStar)$. The key idea to designing efficient algorithms lies in understanding the \emph{optimization landscape} induced by the coarse likelihood objective of \eqref{eq:NLL-intro}. As shown in \Cref{fig:quadratic-growth}, this objective is, in general, {non-convex} and contains undesirable stationary points.

In such settings, one can hope for a \emph{local convergence} algorithm -- one that converges to the optimum given a sufficiently good warm start.
A natural algorithm of this kind is stochastic gradient descent (SGD) performed on \eqref{eq:NLL-intro}.  
However, proving theoretical guarantees for {SGD} is highly non-trivial because it relies on showing that the negative log-likelihood objective of \eqref{eq:NLL-intro} is convex in a sufficiently large neighborhood (of radius $R$) around the optimum.
{Further, while one might be able to establish convexity for very small radii (\eg{}, on the order of $\poly(\nfrac{1}{d})$),} to get useful and interesting algorithmic guarantees (such as in \Cref{infthm:Intro:SS}), one needs $R$ to be ``non-trivially'' large (\eg{}, have no dependence on $\nfrac{1}{d}$).

\subsubsection{A Recipe for Local Convergence.}\label{sec:localConvReceipe}
    Assuming the ingredients in the previous two sections, we get the following local convergence algorithm for inference from a coarse partition.
   In particular, if we can demonstrate that the partition is information preserving and that the negative log-likelihood (NLL) is locally convex, then SGD (with a warm start for which both properties hold) will converge to the true parameters, given access to stochastic gradients of the NLL objective.
    \begin{mdframed}
        \textbf{A Recipe for Local Convergence.}\quad Suppose the following hold.
        \begin{enumerate}[topsep=0.5em,leftmargin=12pt]
            \item $\hyP$ is an $\alpha$-information preserving partition with respect to $\normal{\mu^\star}{\SigmaStar}$ in a ball of radius $R_{1}.$
            \item The NLL is convex in a ball of radius $R_2$ centered at $(\mu^\star, \SigmaStar)$.
            \item There is an oracle that outputs stochastic gradients $g$ of NLL; with $\Ex[\norm{g}_2^2]\leq G^2$. 
        \end{enumerate}
        Then (projected) stochastic gradient descent on the NLL initialized within radius $\min\{R_1,R_2\}$ of $\inparen{\muStar,\SigmaStar}$
        and run for $O\inparen{\nfrac{G^2}{(\alpha^4\eps^2)}}$ steps
        computes an estimate $\sinparen{\wh{\mu},\wh{\Sigma}}$ such that 
        \[
            \Ex\insquare{\norm{\wh{\mu} - \muStar}_2^2 + \snorm{\wh{\Sigma} - \SigmaStar}_F^2} \leq \eps^2\,.
        \]
    \end{mdframed}

\subsection{From Self-Selection to Coarsening}
\label{sec:Overview}
    In this section, we present the main ideas and challenges for obtaining the local convergence algorithm in \Cref{infthm:Intro:SS:localConvergence} for the self-selection problem (\cref{def:ssb}).  

    In \Cref{sec:overview:SS:conversion},
        We show that linear regression under self-selection with the maximum criterion (\Cref{def:ssb}) can be encoded as a coarse-inference task.
        This gives us access to the local convergence recipe from \Cref{sec:localConvReceipe}.
        However, even when seen through the lens of coarsening, the NLL (see \eqref{eq:NLL-intro}) is \emph{non-convex} under the induced partition.
        The main technical challenges are then to lower bound the \emph{information preservation} of the self-selection coarsening mechanism and show that the resulting NLL satisfies \emph{local convexity} around the optimal parameters. 
        In \Cref{sec:overview:SS:informationPreservation}, we describe our approach to showing information preservation in an $\Omega(\nfrac{1}{\log k})$ radius around a global {minimizer},
        directly gives us a quadratic growth condition for the NLL thanks to \Cref{thm:quadGrowth}.
        Then, we illustrate how to argue a $\poly(\nfrac{1}{k})$ radius of the local convexity in \Cref{sec:overview:SS:localConvexity}.
        Perhaps surprisingly, this is independent of $d$ and is the key to the runtime guarantee in \cref{infthm:Intro:SS:localConvergence}, which resolves an open problem of \citet{cherapanamjeri2023selfselection}. 
        Finally, using the algorithm of \citet{gaitonde2024selfselection} as a warm start, we obtain an end-to-end algorithm in \Cref{sec:overview:SS:summary} that improves on the state of the art running time and gives us \Cref{infthm:Intro:SS}.
        
        The rest of this section describes the above steps in more detail.

\subsubsection{{Conversion to a Coarse Labels Problem}}\label{sec:overview:SS:conversion}
    Our main conceptual idea is to convert the self-selection problem (\cref{def:ssb}) to a coarse labels problem, where the learner observes samples $(x,P_{\ymax})$, where
    \begin{equation}
       P_{y} = \sinbrace{ 
        (y_1,\dots,y_k)\colon \exists i \in [k], ~~~~\text{such that},~~~~ y_i = y 
            \quadand\max\nolimits_{j \neq i} y_{j} \leq y
        }\,,\quad y \in \R \,.
    \label{eq:L-shapeSets} 
    \end{equation} 
    \begin{figure}[htbp]
        \centering
        \hspace{5em}
        \begin{subfigure}[t]{0.3\linewidth}
            \centering
            \includegraphics[width=\linewidth]{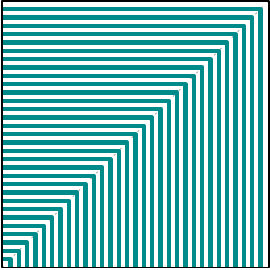}
        \end{subfigure}
        \hfill
        \begin{subfigure}[t]{0.3\linewidth}
            \centering
            \includegraphics[width=\linewidth]{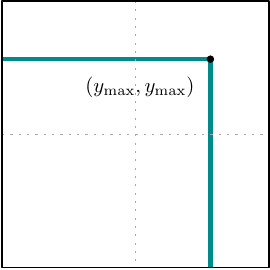}
        \end{subfigure}
        \hspace{5em}
        \caption{The left figure is an approximate illustration of the partition over $\R^2$ that the self-selection mechanism (\cref{def:ssb}) induces
        over the dependent variable space for $k=2$. Each set $P_{y_{\max}}$ corresponds to some green $L$-shape set.
        The true partition covers the entire space with the $2$-dimensional $L$-shapes.
        The right figure is an example of an observation from the self-selection model in the dependent variable.
        {See \cref{fig:self-selection-3D} for an illustration of the partitions with $k=3$.}
        }
        \label{fig:self-selection-partitions}
    \end{figure}
    Here, $P_{\ymax} \subseteq\R^k$ corresponds to the ``$L$-shape sets'' of \Cref{fig:self-selection-partitions} and $\ymax \in \R$ is the observed maximum of the self-selection model given feature $x$.
    For this conversion to be useful, the resulting partition $\{P_{\ymax}\}_{\ymax \in \R}$ of $\R^k$ must be $\alpha$-information preserving with respect to the true parameters $W^\star$ for some non-trivially large $\alpha$ (and within {a} sufficiently large radius {$R$}). Additionally, {(local) convexity must hold within this radius as well}.
    In particular, to get a $\poly(d,\nfrac{1}{\eps})$-time algorithm, the parameter $\alpha$ must be independent of $d$ and to obtain a $\poly(d,k,\nfrac{1}{\eps})$-time local convergence algorithm, {it} must be at least $\poly(\nfrac{1}{k})$.

    \subsubsection{Establishing Information Preservation}\label{sec:overview:SS:informationPreservation}
    \smallskip 
         \noindent First, we discuss several challenges in lower-bounding $\alpha$, and then we present our approach to show \Cref{thm:InfoPresSelfSelection} below. 
         The formal proof is deferred to \Cref{sec:SS:proof:informationPreservation}
    \begin{inftheorem}
        \emph{\textbf{(Information Preservation for Self-Selection; see \Cref{thm:SS:localConvexity})}}
        Consider the model of \Cref{def:ssb}  with true parameters $W^\star = [w_1^\star,w_2^\star,\dots,w_k^\star]$. For any matrix $W\in \R^{d\times k}$ that is $O\inparen{\nfrac{1}{\log{k}}}$-close to $\Wstar$ in Frobenius norm, it holds that
                    $
                        \tv{\cM(W^\star)}{~\cM(W)} \geq 
                            \poly\inparen{\nfrac{1}{k}}
                            \cdot 
                                {\norm{W^\star - W}_F}
                                \,,
                    $
        where $\cM(W)$ (respectively $\cM(W^\star))$ is the distribution of {the observations} ${(x,\max_i y_i)}$ induced by $W$ (respectively $W^\star)$. 
        \label{thm:InfoPresSelfSelection}
    \end{inftheorem}

\begin{figure}
                    \centering
                    \hspace{5em}
                    \begin{subfigure}[t]{0.3\linewidth}
                        \centering
                        \includegraphics[width=\linewidth]{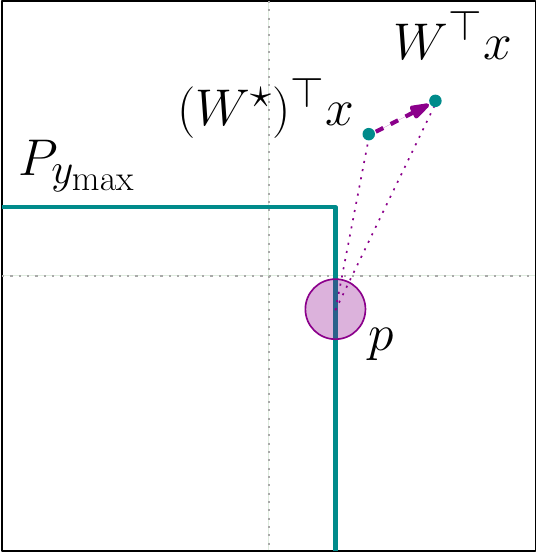}
                    \end{subfigure}
                    \hfill
                    \begin{subfigure}[t]{0.3\linewidth}
                        \centering
                        \includegraphics[width=\linewidth]{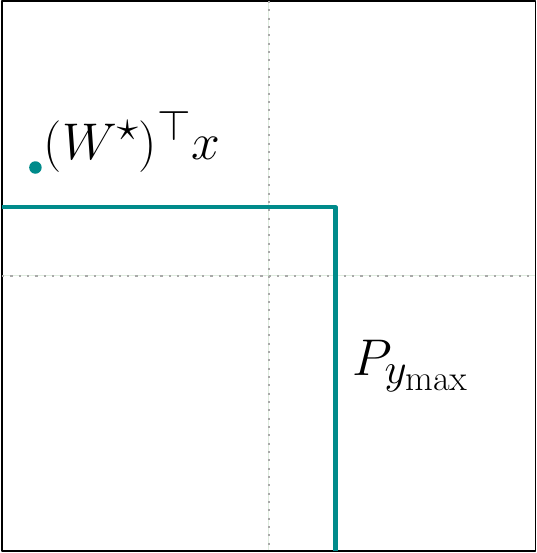}
                    \end{subfigure}
                    \hspace{5em}
                    \caption{
                        The left figure is an illustration of moving $W^\star$ to $W \in \R^{d \times k}$. 
                        In general, it is unclear how the assigned mass on the set $P_{y_{\max}}$ changes. 
                        The depicted direction of change is along an 'easy' direction where the new point $W^\top x \in \R^{k \times 1}$ assigns less mass than $( W^\star)^\top x.$ 
                        This is because any point $p$ on the $L$-shape, is further from $W^\top x$ than from $(W^\star)^\top x$. 
                        The right figure gives an example where the $L$-shape behaves like a convex set since the 
                        random variable $x^\top W^\star$ is ''deep inside'' the $L$-shape, conditioned on the good event $\evE$ and this effectively enables us to ignore all but one part of the $L$-shape.
                    }
                    \label{fig:information-preservation-self-selection}
                    \label{fig:information-preservation-far}
                \end{figure}

    \paragraph{{Challenge I: No Notion of Distance to $P_{\ymax}$.}} %
            {A natural approach to lower bounding $\alpha$ is to use \cref{def:informationPreservation}. This requires showing that}
            if a parameter $W$ is $\eps$-far from $\Wstar$,
            then the distributions $\cM(W)$ and $\cM(\Wstar)$ of the observed maximum $\ymax$ with parameters $W$ and $\Wstar$ respectively are also $\eps$-far from each other (in total variation).
            To show this, {we need} to show that $W$ assigns a higher (or lower) mass to sufficiently many sets $\inbrace{P_{m}}_m$ compared to the mass $\Wstar$ assigns.
            Fix a particular feature $x$ and a value $m\in \R$.
            If the distance to the set $P_{m}$ was well-defined and $W^\top x$ was further from $P_m$ than $( W^\star)^\top x$, then it would be easy to show that $W$ places a lower mass on $P_m$ than $\Wstar$ (and vice versa).
            This follows because the values $y_1,y_2,\dots,y_k$ in \cref{def:ssb} are concentrated around $(W')^\top x$ for each parameter $W'$.
            However, since {the set} $P_m$ is non-convex, we cannot use this argument for a general parameter $W$ close to $\Wstar$.
            The only situation in which this change-in-mass is clear is when $W^\top x$ is a translation of $(W^\star)^\top x$ along the direction of the axis-of-symmetry of the $L$-shapes, \ie{}, along the vector $1_k = (1,1,\dots,1)$
            (\Cref{fig:information-preservation-self-selection} left).

        \paragraph{{Challenge II: Existing Techniques are Insufficient.}}
        While we cannot directly use a notion of distance to the sets $P_m$ (as a notion of distance does not exist), we may still be able to show information preservation via a more complex argument. 
            One idea is to use the analysis in prior works \cite{cherapanamjeri2023selfselection,gaitonde2024selfselection}.
            \citet{cherapanamjeri2023selfselection}, roughly speaking, study the behavior of high-degree moments of $\ymax$ conditioned on certain $e^{-\poly(1/\eps)}$-probability events.
            However, conditioned on an event $\evE$, the largest lower bound on $\alpha$ one can deduce is $\Pr[\evE]$ (\cref{lem:SS:IP:reductionToConditionalIP}) and, hence, \citet{cherapanamjeri2023selfselection}'s approach can only provide a weak lower bound of $\alpha \geq e^{-\poly(1/\eps)}$.
            This is insufficient to obtain a $\poly(d,\nfrac{1}{\eps})$-time algorithm.
            \citet{gaitonde2024selfselection}'s techniques are more promising: they are able to analyze the low-degree moments of $\ymax$ conditioned on an event $\evE$, which happens with a nontrivially large probability, $\Pr[\evE]\geq \poly(\nfrac{1}{k})$.
            Hence, in principle, one might hope to prove a lower bound of $\poly(\nfrac{1}{k})$ conditioned on $\evE$.
            To lower bound $\alpha$, one has to show that the total variation distance between distributions $\cM(W)$ and $\cM(V)$ induced by candidate regressors $W$ and $V$ scales \textit{linearly} with $\norm{W-V}_F$.
            While the moments of $\cM(W)$ and $\cM(V)$ (conditioned on $\evE$), along with other properties of the model, might be sufficient to give some lower bound on $\tv{\cM(W)}{~\cM(V)}$,
            we need significantly tighter analysis to establish this linear dependence for information-preservation. 
            
    \paragraph{Our Approach to Show Information Preservation.}
    One key idea of \citet{gaitonde2024selfselection} is to condition on the event that the covariate $x$ is non-trivially correlated with one of the $k$ regressors $w_1^\star, \dots, w_k^\star$.
    To obtain the $\poly(\nfrac{1}{k})$-information preservation, 
    we draw inspiration from them and craft more involved ``good'' events $\evE_i$ for $i\in [k]$
    over the randomness in $x$ that occur with $\poly(\nfrac1k)$ probability (\cref{def:SS:event}).
    The main conceptual difference in our event is that $\evE_i$ must also capture the interaction between $W^\star, W$.
    Subsequently, we follow a fundamentally different analysis to obtain a sharp linear dependence
    $
        \tv{\cM(\Wstar)}{~\cM(W)}=\Omega(\norm{\Wstar-W}_F)\,.
    $
    
    {The key to obtaining the above linear lower bound is the novel observation that conditioned} on the event $\evE_i$, 
    the $L$-shape $P_y$ {(for certain values $y$)} ``behaves like the convex set'' of the known-index variant of the self-selection problem. 
  {The core intuition} is the following:
    First, $\evE_i$ implies that, for any matrix $A \in \{W^\star, W\}$, 
    the value of the $i$-th model $y_{i, A}$ is far from any other value $y_{j, A}$, $j \neq i$ (\cref{lem:SS:propertiesOfRho}). 
    This means that, conditioned on $\evE_i$, the mass of the measure of $(y_{1,A},\dots,y_{k,A})$ is concentrated ``far'' from the main diagonal of $\R^k$ (where $y_{i,A} \approx y_{j,A}$ for any $j$), thus making the $L$-shape {$P_y$ (for $y\approx y_{i, A}$)}  look like a convex set (see \cref{fig:information-preservation-far}).
    
    {If this was true for all sets $P_y$, then one could hope to reduce our analysis to the analysis of the known index of selection problem studied by \citet{cherapanamjeri2023selfselection}.
    However, because this only holds for sets $P_y$ with  $y\approx y_{i, A}$, we need to do a more careful analysis.}
    {Toward this,} we use the fact that total variation corresponds to the supremum over all events in the $\sigma$-algebra. 
    Hence, {after some calculations,} we get that it suffices to prove that (see \cref{eq:SS:probabilityDiff})
    \[
       \abs{
                \Pr\insquare{y_{\max, W^\star} \leq 0\mid \evE_i}
                -
                \Pr\insquare{y_{\max, W} \leq 0\mid \evE_i}
            }
            \geq 
            \frac{1}{\Pr[\evE_i]}
            \cdot \poly{(\nfrac{1}{k})}
            \cdot {\norm{\Wstar-W}_F}\,.
    \] 
    {Here $y_{\max, W^\star}$ and $y_{\max, W}$ denote the random variable $\ymax$ when the parameter in the self-selection model is $W^\star$ and $W$ respectively.}
    Now, {we use the} structure imposed by the event $\evE_i$ {along with careful} approximations of the ratios of the Gaussian cumulative distribution function (\cref{lem:SS:cdfratios}) to establish the linear dependence in the above inequality.

\subsubsection{Establishing Local Convexity}\label{sec:overview:SS:localConvexity}
Next, we turn to establishing the local convexity result below, whose formal proof is deferred to \Cref{sec:SS:proof:localConvexity}.
    \begin{inftheorem}
    [Local Convexity for Self-Selection; see \Cref{thm:SS:localConvexity}]
        Consider the model of \Cref{def:ssb} with true parameters $W^\star = [w_1^\star,w_2^\star,\dots,w_k^\star]$.
        For any matrix $W\in \R^{d\times k}$ that is $\poly(\nfrac{1}{k})$-close to $\Wstar$ in Frobenius norm, it holds that
                            $
                                \grad^2 \cL(W) \succeq 0\,,
                            $
        where $\cL(\cdot)$ is the population negative log-likelihood objective of the self-selection problem evaluated at $W.$
    \label{thm:LocalConvSelfSelection}
    \end{inftheorem}
A calculation of the Hessian of the negative log-likelihood for the self-selection problem gives that
\[
\nabla^2 \cL(W) = I_{dk} - \Ex_{x,\ymax}\insquare{
            \cov_{z\sim \normal{W^\top x}{I}}\insquare{z\mid z\in P_{\ymax}}
            \otimes xx^\top}\,,
            \yesnum\label{eq:convexity:intro:1}
\]
where $(x,y_{\max})$ has {distribution} induced by \Cref{def:ssb} and the distribution of the covariance matrix is the normal distribution conditioned on the $L$-shape $P_{y_{\max}}$. 
Similar to information preservation, we first discuss various challenges before our approach.

\paragraph{{Challenges in Establishing Local Convexity.}}
To argue about the (local) convexity of $\cL(\cdot)$, we would like to show that certain covariance {matrices} are spectrally upper bounded by the identity (see \eqref{eq:convexity:intro:1}). 
A natural idea is to use classical variance reduction inequalities:
The Brascamp-Lieb inequality states that the variance of the Gaussian distribution reduces when conditioning on a convex set \cite{harge2004convex}. 
For instance, \Cref{fig:convex-partitions} (right) corresponds to a convex partition where each 
$P_{\ymax}$ is replaced by two sets $P_{\ymax}^1$ and $P_{\ymax}^2$ for $\ymax\in \R$.
Unfortunately,
the set $P_{y_{\max}}$ is non-convex,
{making} such inequalities {inapplicable}. 
{Another} idea is to use our ''good'' event $\evE_i$ used before for information preservation in order to reduce the non-convex set $P_{y_{\max}}$ to some convex set. 
However, this idea {also} falls short: 
the event $\evE_i$ has only $\poly(\nfrac1k)$ mass and hence, with the remaining probability,  variance reduction will fail to hold {(this was not an issue for establishing information preservation as the contributions from the part of the event-space where $\evE_i$ does not hold is non-negative)}. %

\medskip 
    
    \noindent\textbf{Our Approach to Show Local Convexity.}
Instead, we follow a more algebraic route. First, thanks to information preservation,  $\cL(W)$  has quadratic growth (\Cref{thm:quadGrowth}).
We hence directly bound the 
change in Hessian value, \ie{}, upper bound the difference $\nabla^2 \cL(W) - \nabla^2 \cL(W^\star)$ for parameters $W$ close to $W^\star$. 
It is indeed not difficult to show that this change is at most of order $\poly(d).$ 
However, this bound gives no algorithmic guarantees. 
A much more delicate analysis, based on the intuition that the population loss is effectively 
$k$-dimensional by the rotation invariance of the covariates, is required to obtain the desired $\poly(k)$ bound.
This yields the $r_{\rm LC}$-sized local-convexity neighbourhood in \Cref{thm:LocalConvSelfSelection}.

More concretely, along $W_u=W^\star+uV$, a score identity writes the derivative of each directional conditional variance as a conditional covariance (\cref{lem:SS:LC:step1}).
Conditional second- and fourth-moment bounds make this variance Lipschitz in $u$.  
To average the resulting pointwise bound, we project $x$ onto the span of the columns of $W^\star$ and $V$.  
This span has dimension at most $2k$; its orthogonal complement remains an independent standard Gaussian.  
Exact Gaussian moments on these two blocks then give the dimension-free population estimate
\[
    \norm{\nabla^2\cL(W)-\nabla^2\cL(W^\star)}_{\rm op}
    \leq O(C^3k^{9/2})\norm{W-W^\star}_F
\]
(\cref{lem:SS:LC:step2,thm:SS:LC:formal}).  
Combining this Hessian stability with the positive curvature at $W^\star$ supplied by information preservation yields the stated $\poly(1/k)$ local-convexity radius.

\subsubsection{Completing the Local-Convergence Algorithm of \Cref{infthm:Intro:SS:localConvergence}}\label{sec:overview:SS:summary} 
The above two results, whose proofs are deferred to \Cref{sec:SS:proof:informationPreservation,sec:SS:proof:localConvexity}, are almost all we need to obtain the local convergence algorithm of \Cref{infthm:Intro:SS:localConvergence}. The algorithm is stochastic gradient descent on the negative log-likelihood objective given a $D_{\rm SS}$-warm-start. To complete the analysis of the algorithm, we need restrict the domain to an appropriate projection set (\Cref{sec:SS:proof:projectionSet}), analyze an unbiased exact-gradient oracle and couple it to an efficient sampler for the Gaussian distribution conditional on $P_{y_{\max}}$, and bound on the second moment of the norm of the gradient (\Cref{sec:SS:proof:gradientSecondMoment}). Along with information preservation and local convexity, these standard steps culminate in the proof of our main \Cref{thm:SS,thm:SS:localConvergence} (see \Cref{sec:SS:proof:PSGD}).

\section{Related Works}\label{sec:relatedWorks}

\noindent\textbf{Learning with Self-Selection Biases.}  
Bias due to outcome self-selection is a
well-documented phenomenon across statistics, econometrics, and
the social sciences \cite{roy1951earnings,fair1972methods,lee1978estimation,willis1979education,olsen1982distributional,maddala1986limited,heckman1979sample}. We refer to the works of \citet{cherapanamjeri2023selfselection,cherapanamjeri2022auctionEstimation} for an exhaustive overview of applications and of related Econometrics works. Compared to our results, the closest works are those of \cite{cherapanamjeri2023selfselection} and \citet{gaitonde2024selfselection}. \cite{cherapanamjeri2023selfselection} provide two sets of results. 
First, they work under \Cref{def:ssb} and give an algorithm estimates the regression parameters to within $\poly(\nfrac{1}{k})$-error in $\poly(d)\cdot e^{\poly(k)}$. One of the most challenging parts of this result is the proof of \emph{identifiability}, which was the inspiration of their moment-based algorithm. Their second set of results concerns the easier problem, which they call the known-index setting. In this case, apart from observing only the maximum of $k$ models, they also observe the \emph{index} of the selected model. This makes the problem more tractable. In the known index setting, \cite{cherapanamjeri2023selfselection} manage to give an efficient algorithm that goes beyond the maximum criterion and works for any \emph{convex-inducing} selection rule. At a technical level, this algorithmic idea is very similar to the idea of \citet{fotakis2021coarse}, who gave a sample-efficient algorithm for coarse Gaussian mean estimation under convex partitions. In fact, under the viewpoint of coarsening, the known index self-selection problem with the maximum criterion (or more generally convex-inducing selection rules) corresponds to a convex partition of the $k$-dimensional Euclidean space.

The follow-up work by \cite{gaitonde2024selfselection}  improved the sample complexity for the maximum self-selection problem to the near-optimal $\widetilde O(d) \cdot \poly(k, \nfrac{1}{\eps})$ and the running time to $\poly(d,k,\nfrac{1}{\eps}) + (\nfrac{(\log{k})}{\eps})^{O(k)}$ {when the error is measured as the maximum $L_2$-error, \ie{}, $\max_i \norm{w_i-w^\star_i}_2^2$. Under the stronger notion of error considered in this work, the joint $L_2$-error, \ie{}, $\sum_i \norm{w_i-w^\star_i}_2^2$, their running time becomes $\poly(d,k,\nfrac{1}{\eps}) + (\nfrac{(k\log{k})}{\eps})^{O(k)}$}.
{Henceforth, we mention their running time for the stronger joint $L_2$-error metric.}
Interestingly, they also provide a moment-based algorithmic approach that works for sub-Gaussian noise. They further provide an algorithm for the related problem of max-affine regression (where the noise is added after taking the maximum). Their result for max-affine regression is an algorithm with sample complexity $\widetilde O(d)\cdot \poly(k,\nfrac{1}{\eps})$ and runtime $\poly(d,k,\nfrac{1}{\eps}) + O(k)^{O(k)},$ improving, in some parameter regime, the result of \citet{ghosh2022maxaffine}. Our techniques could be used to obtain local convergence guarantees for max-affine regression; however, such a result already exists by \citet{kim2024max} (handling sub-Gaussian noise).

            Both \cite{cherapanamjeri2023selfselection} and \cite{gaitonde2024selfselection}'s algorithms for the model in \cref{def:ssb} proceed in two steps. First, they construct a $k$-dimensional subspace containing $w_1^\star,\dots,w_k^\star$ by computing the span of the top eigenvectors of a carefully-constructed weighted covariance matrix. Then, they recover $w_1^\star,\dots,w_k^\star$ from this subspace by relying on conditional moments of the observations.
            These works differ in the specific event upon which they perform the conditioning:
            \cite{cherapanamjeri2023selfselection} condition on the event that the covariate is highly correlated with a guess vector,
            while \cite{gaitonde2024selfselection} choose an event with significantly higher probability
            by only requiring the correlation to be somewhat non-trivial.
            In either case, both approaches explicitly compute estimates of their respective conditional moments and use those estimates to locate $\eps$-close guess vectors to some regressor $w^\star_i$
            {by searching over an $\eps$-net over a $k$-dimensional ball}.
            Our approach of optimizing a suitably chosen likelihood function is quite different from the brute force methods (over carefully chosen $k$-dimensional subspaces) in these works, and is crucial to avoid the $\eps^{-k}$ dependence suffered by them.

\medskip 
    
    \noindent\textbf{Inference with Coarse Data.}
Estimation under coarse examples has a long history in Statistics,
going back to the works of \citet{heitjan1990inference,heitjan1991ignorability,heitjan1993ignorability}, who introduce the coarsening at random framework (CAR), generalizing the missing at random (MAR) model of \citet{rubin1976inference,little1989analysis}. These works and follow-ups \cite{little2019statistical,gill1997coarsening,robins1992recovery,van1996efficient,jacobsen1995coarsening,grunwald2003updating}  focus on statistical aspects of coarsening. In these works, coarsening is based on very simple coarsening mechanisms, such as intervals, motivated by survival analysis problems, and the standard estimation method is based on maximum likelihood estimation, whose computational complexity was not carefully treated in this line of work in Statistics. 

In terms of algorithms in high dimensions, the work of
\cite{fotakis2021coarse} is the closest to our work regarding coarsening. In terms of distribution learning, their work provides a sample-efficient algorithm for coarse Gaussian mean estimation when the sets of the partition are convex. 
In terms of computational hardness, \cite{fotakis2021coarse} show that, under standard computational complexity conditions, there is no efficient  algorithm for coarse Gaussian mean estimation under general non-convex partitions.

    \medskip 
    
    \noindent\textbf{Truncated and Censored Statistics.}
            Our work is also closely related to the literature on learning from censored-truncated data.  
            Truncation and censoring have a long history of work in Statistics \cite{Cohen91} (and \cite{Galton1897,pearson1902v,pearson1908generalised,Lee1915,fisher31,Cohen91,hannon1999estimation,raschke2012inference,maddala1986limited}), Econometrics \cite{maddala1986limited} {(and references therein)}, and a growing recent literature in Computer Science \cite{daskalakis2018efficient,daskalakis2019computationally,daskalakis2020truncated,ilyas2020theoretical,fotakis2020efficient,lee2023learning,lee2024unknown,Kontonis2019EfficientTS,daskalakis2021statistical,plevrakis2021censored,diakonikolas2024statistical,trunc_regression_unknown_var,truncatedDiscrete,truncated_sm,nagarajan2020truncatedMixtureEM,nagarajan2023EMMixtureTruncation,tai2023mixtureCensored}.
            Truncation occurs when samples falling outside of some subset $S^\star$ of the support of the distribution, called \textit{survival set}, are not observed. 
            Truncation arises in a variety of fields from Econometrics \cite{maddala1986limited}, to Astronomy and other physical sciences \cite{woodroofe1985estimating}, to Causal Inference \cite{imbens2015causal,hernan2023causal}.
            Another recent line of work tackles the problem of \textit{testing} whether a given source of data is truncated or not \cite{canonne2020learning, de2023testing, de2024detecting}. 
            The main connection between our algorithms and this line of research is the use of the negative log-likelihood as the key optimization objective. 
            Interestingly, while in the works of \cite{daskalakis2018efficient}, the structure of the truncation set is not related to the computational efficiency of the algorithm (the set should only have mass but can be highly non-convex).
            This is because they assume oracle access to the survival set $S^\star$, whereas when the truncation set is unknown, its low complexity (\eg{}, convexity) becomes important for computational efficiency~\citep{Kontonis2019EfficientTS}.
            In the coarse setting of \Cref{section:tool}, the convexity of the sets in the partition is crucial for a favorable optimization landscape.
 
\section{Conclusion and Discussion}\label{sec:conclusion}
The tractability for convex partitions and intractability for some non-convex partitions positions the problem of self-selection \emph{in between} the algorithms for convex partitions and the hardness result of \citet{fotakis2021coarse} for general non-convex partitions: on the one hand, the self-selection problem of \Cref{def:ssb} gives rise to \emph{non-convex} partitions, and hence the algorithm for convex partitions is not useful. On the other hand, these non-convex sets are \emph{structured} and \emph{well-behaved}, leaving open the possibility of an SGD-based global convergence algorithm.
\begin{description}
    \item[Open Problem.] 
    {\emph{Can an SGD-based algorithm solve self-selection in $\poly(d,k,\nfrac{1}{\eps})$ time?}}
\end{description}

    \section*{Acknowledgements}
        {We thank the anomymous reviewers for comments and suggestions. We also thank an anonymous reviewer for pointing out an error in the proofs of \Cref{eq:SS:probabilityDiff,lem:SS:LC:step1}, which is now fixed.}
        We thank Manolis Zampetakis for feedback on an initial draft of this work which helped improve its presentation, discussions about the literature on estimation from auction data, and for sharing the reference to \citet{meilijson1981autopsy}.
        Alkis Kalavasis was supported by the Institute for Foundations of Data Science at Yale.
        Felix Zhou acknowledges the support of the Natural Sciences and Engineering Research Council of Canada (NSERC).
    \printbibliography

\appendix
\crefalias{section}{appendix}
\crefalias{subsection}{subappendix}
\crefalias{subsubsection}{subsubappendix}

\addtocontents{toc}{\protect\setcounter{tocdepth}{2}}

\newpage
\section{Organization of the {Appendix}} 
We organize the rest of the paper as follows:
\begin{itemize}
    \item \Cref{sec:prelims} introduces some notational conventions and standard preliminaries.
    \item \Cref{apx:models-main-results} restates the maximum self-selection model as well as our main results.
    \item \cref{sec:SS} provides proofs for linear regression with self-selection bias under the maximum selection criterion. In particular,  \cref{sec:SS:proof:informationPreservation} establishes information preservation and \cref{sec:SS:proof:localConvexity} proves local convexity.

    \item \Cref{sec:hessian} fills in some gradient and Hessian calculations deferred from \Cref{sec:SS}. %
    \item \cref{sec:sgd-analysis} offers a general analysis of Projected Stochastic Gradient Descent (PSGD) for functions that are locally convex and satisfy local quadratic growth, which is key to all our local convergence methods.
    \item \Cref{sec:proofsDeferredSS} fills in the full details of technical proofs from \Cref{sec:SS} for completeness.
\end{itemize}

\section{Preliminaries and Notation}\label{sec:prelims}
    In this section, we introduce preliminaries and (standard) notations used throughout this work.
    \input{preliminaries}

\section{Models and Main Results}\label{apx:models-main-results}
   {
    In this section, we outline the results that lead to our main \Cref{thm:SS,thm:SS:localConvergence}.
    To illustrate the core concepts of information preservation, we give a general abstract formulation in \Cref{sec:info-pres-noisyChannel} and present the implications for quadratic growth of the negative log-likelihood objective.
    This is specialized for coarsening in \Cref{sec:info-pres-coarse}.
    \Cref{sec:self-selection} then restates our main results for the reader's convenience.}
    
    \subsection{Information Preservation for General Noisy Channels}
    \label{sec:info-pres-noisyChannel}

    In this section, we present an abstract framework for information preservation.
    Consider a domain $\cX$ and a family of distributions $\{ P(\theta)\colon \theta \in \Theta\}$ over it.
    Let $f\colon \cX \to \cY$ be a  (deterministic) distortion mechanism that transforms each element $x \in \cX$ to some element of $\cY$.
    This mapping induces, for any distribution $P(\theta)$, a new distribution $P_f(\theta)$ over $\cY$, where $P_f(\theta; y) = \int_{x\colon f(x) = y} P(\theta; x) \d x $ for any $y \in \cY$.
    The statistical task of interest is the following: given \iid{} draws $f(x_1),\dots,f(x_n)$ from the distorted distribution $P_f(\theta^\star)$ for some unknown $\theta^\star \in \Theta$ (without observing $x_1,\dots,x_n)$, the goal is to estimate $\theta^\star.$

    \begin{definition}[Information Preserving Distortion Mechanism]\label{def:model:informationPreservingDistortion}
    We say that $f$ is \emph{$\alpha$-information preserving at radius $R>0$ with respect to $P(\theta^\star)$} if, 
    for any $\theta \in \Theta$ satisfying $\norm{\theta-\theta^\star}_2\leq R$,
    \[
        \tv{P_f(\theta)}{P_f(\theta^\star)}
        \geq \min\inbrace{1, \alpha \norm{\theta - \theta^\star}_2}\,.
    \]
    If the radius $R$ is not explicitly mentioned,
    we understand it to be $R=\infty$.
    \end{definition}
    A simple application of Pinsker's inequality shows the following important implication for the growth of the log-likelihood. Let us set
    \[
    \hyL_f(\theta) 
    =
    -\E_{y \sim P_f(\theta^\star)}
    \insquare{\log P_f(\theta; y)}
    =
    -\E_{y \sim P_f(\theta^\star)}
    \insquare{\log \int_{x\colon f(x) = y} P(\theta; x) \d x}\,.
    \]
    
    \begin{lemma}\label{lem:model:informationPreservingImpliesGrowth}
    Consider a distortion mechanism $f$ that is {$\alpha$}-information preserving at radius $R$ with respect to $P(\theta^\star)$. Then, the negative log-likelihood $\hyL_f$ at any $\theta \in \Theta$ where $\norm{\theta-\theta^\star}_2\leq R$ satisfies
    \[
    \hyL_f(\theta)-
    \hyL_f(\theta^\star)
    \geq 
    \min\inbrace{2, 2\alpha^2 \norm{\theta - \theta^\star}_2^2}\,.
    \]
    \end{lemma}
    \begin{proof}
    We have that
    \[
    \hyL_f(\theta)-
    \hyL_f(\theta^\star)
    =
    \E_{y \sim P_f(\theta^\star)}
    \insquare{
    \log(P_f(\theta^\star) / P_f(\theta))
    }
    = \kl{P_f(\theta^\star)}{P_f(\theta)}
    \geq 
    2 \tv{P_f(\theta)}{P_f(\theta^\star)}^2 
    \,.
    \]
    Hence, this implies that
    \[
    \hyL_f(\theta)-
    \hyL_f(\theta^\star)
    \geq  
    \min\inbrace{2, 2\alpha^2 \norm{\theta - \theta^\star}_2^2} 
    \,. \qedhere 
    \]
    \end{proof}
    \Cref{lem:model:informationPreservingImpliesGrowth} shows that it suffices to prove information preservation
    in order to show that the true parameter $\theta^\star$ is a {local minimum} of the negative log-likelihood function.
    
    \subsection{Information Preservation for Coarsening}
    \label{sec:info-pres-coarse}
    In this section, we specialize the previous abstract setting to the coarsening mechanism.
        Let $\hyP$ be a partition of $\R^d$.
        Let $\mu\in\R^d$ and $\Sigma\in \R^{d\times d}$ be the parameters of Gaussian distribution $\normal{\mu}{\Sigma}$.
        Define the coarse Gaussian distribution $\coarseNormal{\mu}{\Sigma}{\hyP}$ as the discrete distribution on $\hyP$ satisfying that, for each set $S \in \hyP$, 
        \[
            \Pr_{P \sim \coarseNormal{\mu}{\Sigma}{\hyP}}\insquare{P= S}
            =
            \normalMass{\mu}{\Sigma}{S}\,.
        \]
        With certain partitions (\eg{}, the singleton partition covering the entire domain, \ie{}, $\hyP = \{\R^d\}$), it is information-theoretically impossible to identify the parameters of the underlying Gaussian distribution.
        To ensure that identification is possible, we assume that the partition preserves information as introduced in \Cref{def:informationPreservation}. Our first observation is that one can employ information preservation of an arbitrary partition $\hyP$ to get quadratic growth guarantees for the negative log-likelihood objective
        \begin{equation}
        \label{eq:NLL}
        \hyL(\mu, \Sigma) = \E_{P \sim \coarseNormal{\mu^\star}{\Sigma^\star}{\hyP}}[- \log \normalMass{\mu}{\Sigma}{P}] \,.
        \end{equation}
        In the above, $(\mu, \Sigma)$ correspond to guesses of the true {unknown} parameters $(\mu^\star, \Sigma^\star).$

       The next result is an immediate consequence of \Cref{sec:info-pres-noisyChannel}.
        \begin{theorem}
        \label{thm:info-pres-implies-strong-conv}
        Consider a partition $\hyP$ of $\R^d$ which is $\alpha$-information preserving at radius $R$ with respect to $\normal{\mu^\star}{\Sigma^\star}$. Then the negative log-likelihood satisfies
        \[
        \hyL(\mu, \Sigma)
        -
        \hyL(\mu^\star, \Sigma^\star) 
        \geq  
        \min \inbrace{
            2, 
            2\alpha^2 \inparen{\|\mu - \mu^\star\|_2^2 + \|\Sigma - \Sigma^\star\|_{F}^2}} 
        \]
        for every $\mu, \Sigma$ such that $\inparen{\norm{\mu-\muStar}_2^2 + \norm{\Sigma-\SigmaStar}_F^2}^{\sfrac12}\leq R$. 
        \end{theorem}

    \subsection{Linear Regression with Self-Selection Bias}
    \label{sec:self-selection}
\begin{figure}[tbh!]
        \centering
        \includegraphics[width=0.5\linewidth]{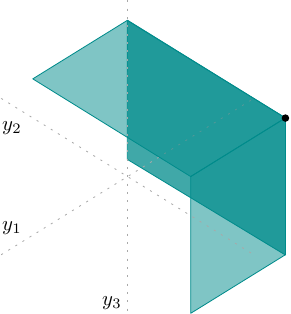}
        \caption{
            \label{fig:self-selection-3D}
            {This figure illustrates one set $P$ in the partition $\hyP$ arising in the self-selection problem with $k=3$. 
            The figure illustrates the set $P$ corresponding to the observation $\ymax=1$.
            $P$ is defined as the set of points $(y_1,y_2,y_3)$ with $\max\inbrace{y_1,y_2,y_3}=1$ and $y_1,y_2,y_3\leq 1$
            In other words, $P = \sinbrace{y_1= 1, y_2\leq 1, y_3\leq 1}\cup \sinbrace{y_1\leq 1, y_2= 1, y_3\leq 1}\cup \sinbrace{y_1\leq 1, y_2\leq 1, y_3= 1}$.
            See \cref{fig:self-selection-partitions} for an illustration of the entire partition $\hyP$ with $k=2$.}
        }
    \end{figure}

    In this section, we formally present our main result for linear regression under self-selection bias \cite{cherapanamjeri2023selfselection,gaitonde2024selfselection}.
    As is usual in multiple linear regression, the goal is to estimate $d\times k$ unknown parameters $\inbrace{w^\star_i\in \R^d\colon 1\leq i\leq k}$ that determine the relation between the independent variable $x\in \R^d$ and the dependent variables $y_1,y_2,\dots,y_k$ as follows 
    \[
        \text{for each $1\leq i\leq k$}\,,\quad 
            y_i = \inangle{x, w_i^\star} + \xi_i 
            \quadwhere \xi_i \sim \normal{0}{1}\,.
    \]
    Equivalently, $(\xi_1,\ldots,\xi_k)\sim\normal{0}{I_k}$ independently of $x$.
    If we observe samples of the form $\inparen{x,y_1,\dots,y_k}$, then this problem reduces to $k$ independent linear regression problems and can be solved using standard methods, \eg{}, least squares regression.
    Under self-selection, instead of observing $\inparen{x,y_1,\dots,y_k}$, one observes $\inparen{x,f(y_1,\dots,y_k)}$ for some known function $f\colon\R^k\to\R$, which prevents us from using learning each regressor $w^\star_i$ separately. %
    We focus on the max-self-selection bias, where $f(\cdot)=\max(\cdot)$ is the maximum function.
    This setting was studied by \citet{cherapanamjeri2023selfselection,gaitonde2024selfselection}.
    In this model, we observe samples of the form (see \cref{def:ssb})
    \[
        \inparen{x\,,~~ y_{\max}\coloneqq \max_{1\leq i\leq k} y_i}\,.
        \yesnum\label{eq:SS:observations}
    \]
    {See \cref{fig:self-selection-partitions,fig:self-selection-3D} for an illustration fo the resulting coarse partitions with $k=2$ and $k=3$ respectively.}
    Since the distribution of these observations is invariant to permutations of  $\inbrace{w^\star_i\colon 1\leq i\leq k}$, the goal is to recover the parameters up to permutation.
    For an arbitrary ordering $w^\star_1,w^\star_2,\dots,w^\star_k$ of the parameters, we define $\Wstar$ to be the following matrix:
    \[
        \Wstar \coloneqq \begin{bmatrix}
            w^\star_1 & w^\star_2 & \dots & w^\star_k
        \end{bmatrix}\,.
    \]
    We study this problem under the following assumptions that match the ones in prior work \cite{cherapanamjeri2023selfselection,gaitonde2024selfselection}. First, we assume that the feature vectors $x$ are independently drawn from $\normal{0}{I_{d}}.$
   \begin{restatable}[Gaussianity]{assumption}{selfSelectionGauss}\label{asmp:SS:gaussianity}
        Covariates $x$ are sampled from the $d$-dimensional Gaussian $\normal{0}{I_d}$. 
    \end{restatable} 
    Next, we impose some separability  and boundedness conditions for the true parameter vectors.
    \begin{restatable}[Separability and Boundedness]{assumption}{selfSelectionSeparation}\label{asmp:SS:separationBoundedness}
        There are known $c\in (0,1]$ and $C\geq 1$ such that 
        $
            \norm{w^\star_i}_2^2 
            \geq c + \max_{j\neq i}~\sabs{\sinangle{w^\star_j, w^\star_i}} 
        $ 
            and
        $
            \max_{1\leq i\leq k}\norm{w^\star_i}_2 \leq C\,.
        $
        For $k=1$, the maximum over the empty set is understood to be zero.
    \end{restatable}

    \noindent Even with this pair of assumptions, it is non-trivial to show that $\Wstar$ is identifiable from the data in \cref{eq:SS:observations} (see \cite{cherapanamjeri2023selfselection} for details).
    The Gaussian prior on $x$ (\cref{asmp:SS:gaussianity}) is a classic assumption that is common in similar problems, such as mixtures of linear regressions (\eg{}, \cite{silvia2020EMmixtures}) and is crucially used to show identifiability. 
    Regarding \cref{asmp:SS:separationBoundedness}, 
    its necessity is shown in \citep[Remark 6.1]{gaitonde2024selfselection},
    which gives a simple example demonstrating that at least $k^{\Omega\inparen{\sfrac{C^4}{c^2}}}$ samples are necessary to estimate $W^\star$ to any non-trivial accuracy. 
    In this sense, some ``sparsity'' of the weight vectors is required to obtain sample-efficient learners under self-selection.

        Our main result, restated below, is a $\poly(d)\cdot \inparen{\nfrac{1}{\eps}}^2 \cdot \sinparen{k\log{k}}^{O(k)}$ time algorithm to estimate the parameters $W^\star$ {up to $\eps$-accuracy} in Frobenius norm error (up to permutation) under the above self-selection model.
        Previously,
        \cite{cherapanamjeri2023selfselection} obtained a $\poly(d)\cdot 2^{\poly(k, {1}/{\eps})}$ time algorithm
        and \cite{gaitonde2024selfselection} obtained a ${\poly(d)\cdot \inparen{\nfrac{k\log k}{\eps}}^{O(k)}}$ time algorithm {to obtain $\eps$-accuracy} in {the Frobenius norm (see \cref{rem:error:maximumVsJoint}).}
        Thus, our algorithm strictly improves the running time of prior works in Frobenius norm.
        \thmSelfSelectionMain*
        \noindent The main new ingredient required to establish \cref{thm:SS} is a proof that the negative log-likelihood arising in self-selection is strongly convex in a $\poly\inparen{\nfrac{1}{k}}$-sized neighborhood of $\Wstar$ (\cref{thm:SS:localConvexity}).
        Note that globally the negative log-likelihood is highly non-convex: it has at least $k!$ distinct stationary points (corresponding to the permutations of $\Wstar$).
        This local strong convexity is sufficient to deduce \cref{thm:SS} as a point within this $\poly\inparen{\nfrac{1}{k}}$-sized neighborhood can be found using existing algorithms for self-selection and, subsequently, one can obtain an estimate $\eps$-close to $\Wstar$ in $\poly(d,k,\nfrac{1}{\eps})$ time using PSGD.
        To formally state this local convergence result, we use the following definition.  A matrix $W$ is an $R$-warm start for an ordered matrix $W^\star$ if $\norm{W-W^\star}_F\leq R$.
        The formal local convergence result is restated below.
        \thmSelfSelectionLocalConvergence*
        As mentioned before, \cref{thm:SS:localConvergence} follows by a local strong-convexity property of the negative log-likelihood. 
        Given parameters $W$, define:
        \[
            \text{
                $\cM(W)\coloneqq$ the distribution of {$(x,y_{\max})$} in the self-selection instance corresponding to $W$\,.
            }
        \]
        Further, define $\cL(\cdot)$ as the negative log-likelihood at point $W$.
        We divide the proof of local strong convexity into two parts (quadratic growth and local convexity), as stated in the next result.
        \begin{restatable}[Information Preservation and Local Convexity for Self-Selection]{theorem}{thmSelfSelectionLocalConvexity} 
            \label{thm:SS:localConvexity}
            Suppose \cref{asmp:SS:gaussianity,asmp:SS:separationBoundedness} hold with constants $c,C>0$.
            Fix an ordering of the true vectors and write $W^\star=[w_1^\star\mid\cdots\mid w_k^\star]$.
            Define
            \[
                r_{\rm IP}
                \coloneqq
                \frac{1}{10^6}
                \inparen{\frac{c}{2C}}^9
                \frac{1}{\sqrt{\log(\nfrac{ek}{c})}}\,.
            \]
            There is also a radius
            $r_{\rm LC}=(c/(ek))^{O((C/c)^{18})}$ for which the local-convexity conclusion below holds.
            The following hold. 
            \begin{enumerate}[leftmargin=15pt]
                \item \textit{(Information Preservation)}
                    For any matrices $V,W\in \R^{d\times k}$ that are $r_{\rm IP}$-warm starts for $\Wstar$,
                    \[
                        \tv{\cM(V)}{~\cM(W)} \geq 
                        \inparen{\frac{c}{ek}}^{{O({C^{18}/c^{18}}){}}}
                        \cdot \norm{V - W}_F\,.
                        \yesnum\label{eq:SS:informationPreservation}
                    \]
                \item \textit{(Local Convexity)} For any matrix $W\in \R^{d\times k}$ that is a $r_{\rm LC}$-warm start for $\Wstar$,
                    \[
                        \grad^2 \cL(W) \succeq 0\,.
                    \]
            \end{enumerate}
            
        \end{restatable}  
        For the proof of \cref{thm:SS:localConvexity}, we refer the reader to \Cref{sec:SS}.

\section{Proofs of Main \texorpdfstring{\Cref{thm:SS,thm:SS:localConvergence}}{Theorems}}
\label{sec:SS}
\input{self_selection}

\section{Gradient Computation for Negative Log-Likelihoods}\label{sec:hessian}
\input{gradient_computation}

\section{PSGD Convergence for Convex Functions Satisfying a Local Growth Condition}
\label{sec:sgd-analysis}
\input{quadratic_growth}

\section{Proofs Deferred From \texorpdfstring{\cref{sec:SS}}{Section 5}}\label{sec:proofsDeferredSS}
\input{self_selection_proofs}

\end{document}

%% file: macros.tex
\DeclarePairedDelimiter{\iprod}{\langle}{\rangle}

\DeclareMathOperator{\proj}{proj}

\newcommand{\sset}{\subseteq}

%% file: preliminaries.tex
\paragraph{Vectors and Matrices.}
We use standard notations for matrix and vector norms:
For a vector $v \in \R^d$, we use $\|v\|_2$ to denote its $\ell_2$-norm and $\|v\|_\infty$ its $\ell_\infty$-norm.
For a matrix $A = (A_{i,j})_{i \in [m], j \in [n]}$, we use $\|A\|_F \triangleq \sqrt{\sum_{i} \sum_{j} |A_{i,j}|^2}$ to denote its Frobenius norm and $\|A\|_2 \triangleq \max_{v \neq 0} \nfrac{\|A v\|_2}{\|v\|_2}$ for its spectral norm. Recall that $\|A\|_2 \leq \|A\|_F.$

\paragraph{Unit Vectors, Balls, and Miscellaneous Notation.}
Moreover, we set $\wh{v}$ to be the unit vector associated with the direction of $v$, \ie{}, $\wh{v} = \nfrac{v}{\|v\|_2}$ for $v \neq 0.$
Given a vector $v$, we use $B(v,R)$ to denote the $\ell_2$-ball centered at $v$ with radius $R$ and $B_{\infty}(v,R)$ for the associated $\ell_\infty$-ball.
Also, for $v\in \R^d$ and $i\in [d]$,
$v_{-i}\in \R^{d-1}$ denotes the vector obtained from $v$ by removing the $i$-th coordinate.

\paragraph{Span, Projection, and Kronecker Product.}
For two vectors $v,w$, we let $\mathrm{span}(v,w)$ denote their span, \ie{}, the set of
all linear combinations of those vectors. We also denote their inner product as $v^\top w.$
A projection of $v$ onto $w$ (formally to the span of $w$) is denoted by $\proj_w(v) \triangleq (v^\top w) \cdot w/{\norm{w}_2^2} = (v^\top\wh{w}) \cdot \wh{w}.$ This can be extended to projections of $v$ onto a subspace $W$ by writing $W$ as the span of $\{w_1,\dots,w_k\}$ and letting $\proj_W(v) = \sum_{i \in [k]} (v^\top \wh{w}_i) \cdot \wh{w}_i.$
Given two matrices $A \in \R^{a} \times \R^b, B \in \R^{c} \times \R^d$, we use $A \otimes B \in \R^{ac} \times \R^{bd}$ to denote the Kronecker product between them. When the identity matrix's dimension is unclear from the context, we use $I_k$ and $I_{k\times k}$ to denote the $k\times k$ identity matrix.
For a matrix $W \in \R^{d \times k}$ and $x \in \R^d$, we use $W^\top x$ to denote the result of the multiplication (which is a column vector).

\paragraph{Partitions.}
We usually denote by $\hyP$ a partition of $\R^d$ and by $P$ an element of this partition. We will call a partition convex if each $P \in \hyP$ is convex. We may also define a distribution $\pi$ over different partitions of $\R^d.$

\paragraph{Gaussians, Coarse Gaussians, and Distances Between Distributions.}
We use $\normal{\mu}{\Sigma}$ to denote the Gaussian distribution with parameters $\mu \in \R^d$ and $\Sigma \in \R^{d \times d}. $
The mass of some set $S \subseteq \R^d$ under this distribution is denoted by $\normalMass{\mu}{\Sigma}{S}.$ More broadly, for a distribution $p$  and set $S$, $p(S)$ is the mass of $S$ under $p$. For some partition $\hyP$, we use the notation $\cN_{\hyP}(\mu, \Sigma)$ to denote the coarse Gaussian distribution, defined as
$\cN_{\hyP}(\mu, \Sigma)(P) = \int_P \normal{\mu}{\Sigma}(x) \d x$ for each $P \in \hyP.$ For two distributions $p,q$ over $\R^d$, we denote their total variation distance as $\tv{p}{q} = \inparen{\sfrac{1}{2}} \int_{\R^d} |p(x) - q(x)| \d x$ and their KL divergence as 
$\kl{p}{q}= \E_{x \sim p}[\log ( \nfrac{p(x)}{q(x)} )].$

\begin{fact}[Gaussian Tail Bound~\citep{duembgen2010boundingstandardgaussiantail}]\label{fact:GaussianTail}
If $X\sim\normal{0}{\sigma^2}$ and $z>0$, then
\[
    \Pr[X\ge z]
    \leq \frac{\sigma}{\sqrt{2\pi}z}
        \exp\!\left(-\frac{z^2}{2\sigma^2}\right).
\]
By symmetry and a union bound, it also follows that
$\Pr[|X|\ge z]\leq 2\exp(-z^2/(2\sigma^2))$.
\end{fact}

%% file: self_selection.tex
In this section, we prove \cref{thm:SS:localConvexity}, which implies our main \Cref{thm:SS,thm:SS:localConvergence}: an efficient local convergence algorithm for the self-selection with max selection rule. 
\noindent We divide the proof of \cref{thm:SS:localConvexity} into two parts corresponding to the two claims.
\subsection{Information Preservation}\label{sec:SS:proof:informationPreservation}
     In this section, we prove the information preservation promised in \cref{thm:SS:localConvexity} (namely \cref{eq:SS:informationPreservation}); we follow the outline in \cref{fig:outline:SS:IP}.
     
     \begin{figure}[!ht]
          \centering
    \vspace{4mm}
    \resizebox{\textwidth}{!}{%
    \begin{tikzpicture}[node distance=2.5cm, auto]
    \node[myNodeNarrow, line width=1pt] (thm) at (0,0cm) 
        {{Information Preservation (Claim 1 of \cref{thm:SS:localConvexity})}};
    \node[myNodeNarrow, text width=8cm, line width=1pt] (subres1) at (-5cm, -3cm) 
        {
        {Reduction to ``Conditional''\\ Information Preservation (\cref{lem:SS:IP:reductionToConditionalIP})}\\[1mm]
        \begin{tabular}{p{0.01cm}@{}p{7cm}@{}}
          &
          \textcolor{gray}{\textit{\small $\beta$-Information Preservation Conditioned on $\evE$}}\\
          &\textcolor{gray}{\textit{\small $\implies\beta\cdot \Pr[\evE]$-Information Preservation}}
        \end{tabular}
        };
    \node[myNodeNarrow, text width=7.25cm, line width=1pt] (subres3) at (5cm, -2.69cm) 
        {
            {Lower bound on Information Preservation Conditioned on Event $\evE$ (\cref{def:SS:event})}
        };

    \node[myNodeNarrow, draw, line width=1pt] (step1) at (-1.5cm, -4.775cm) {Properties of Event $\evE$};
    \node[myNodeNarrow, draw, text width=7cm, line width=1pt] (step2) at (5cm, -5cm) {Elementary Set Witnesses Information Perservation Conditional on $\evE$ {(Eq.~\eqref{eq:SS:probabilityDiff})}};
    
    \node[myNodeNarrow, text width=3cm, dashed] (probOfE) at (-7.5cm, -7.22cm) {$\Pr[\evE] \geq \poly(\nfrac{1}{k})$ (\cref{lem:SS:probability})};
    \node[myNodeNarrow, text width=3.75cm, dashed] (separation) at (-1.5cm, -7.22cm) {
        $\evE$ Implies Separation (\cref{lem:SS:propertiesOfRho})
    }; 

    \node[myNodeNarrow, dashed, text width=2.5cm] (decomposition) at (2.1cm, -7.45cm) {Decomposition Into Terms A, B, and C \eqref{eq:SS:differenceAsProduct}}; 
    \node[myNodeNarrow, dashed, text width=2.5cm] (bounds) at (5cm, -7.45cm) {Bounds on Terms A, B, and C}; 
    \node[myNodeNarrow, dashed, text width=2.5cm] (lowerBound) at (7.9cm, -7.225cm) {Lower Bound on {Eq.~\eqref{eq:SS:probabilityDiff}}}; 
    
    \node[myNodeNarrow, dashed, text width=2cm] (termC) at (2.5cm, -9.5cm) {Bound on Term C}; 
    \node[myNodeNarrow, dashed, text width=2cm] (termB) at (5cm, -9.5cm) {Bound on Term B}; 
    \node[myNodeNarrow, dashed, text width=2cm] (termA) at (7.5cm, -9.5cm) {Bound on Term A}; 

    \node[myNodeNarrow, dashed, text width=2.75cm] (technicalLemma) at (3.75cm, -11cm) {\cref{lem:SS:cdfratios}}; 

    \draw[-stealth, line width=0.5mm] (thm) -- (subres1.north); 
    \draw[-stealth, line width=0.5mm] (thm) -- (subres3.north);

    \draw[-stealth, line width=0.5mm] (subres3) -- (step1);
    \draw[-stealth, line width=0.5mm] (subres3) -- (step2);
    
    \draw[-stealth, line width=0.5mm] (step1) -- (probOfE);
    \draw[-stealth, line width=0.5mm] (step1) -- (separation);  

    \draw[-stealth, line width=0.5mm] (step2) -- (decomposition);  
    \draw[-stealth, line width=0.5mm] (step2) -- (bounds);   
    \draw[-stealth, line width=0.5mm] (step2) -- (lowerBound);  

    \draw[-stealth, line width=0.5mm] (bounds) -- (termA);  
    \draw[-stealth, line width=0.5mm] (bounds) -- (termB);  
    \draw[-stealth, line width=0.5mm] (bounds) -- (termC);  

    \draw[-stealth, line width=0.5mm] (termB) -- (technicalLemma);  
    \draw[-stealth, line width=0.5mm] (termC) -- (technicalLemma);  
    
    \end{tikzpicture}
    }
        \caption{Outline of Proof of Information Preservation for Self-Selection.}
        \label{fig:outline:SS:IP}
    \end{figure}

    \noindent Toward proving  \cref{{eq:SS:informationPreservation}}, we first dispose of two degenerate cases.  When $k=1$, the observation is $y=w_1^\top x+\xi$.  Thus, writing $h=v_1-w_1$ and letting $\Phi$ denote the standard Gaussian CDF, the common-marginal kernel identity gives
    \[
        \tv{\cM(V)}{~\cM(W)}
        =\Ex_x\!\left[2\Phi\!\left(\frac{|h^\top x|}{2}\right)-1\right].
    \]
    For $a\geq0$, $2\Phi(a/2)-1\geq\phi(1/2)\min\{a,1\}$, where $\phi$ is the standard Gaussian density.  The radius in \cref{eq:SS:distance} is at most $10^{-6}$, so $\norm h_2\leq1$; writing $h^\top x\stackrel{\rm d}=\norm h_2G$ for $G\sim\normal{0}{1}$ therefore shows that the last display is at least
    $\phi(1/2)\Ex[|G|\mathds1\{|G|\leq1\}]\norm h_2$.  This proves the claim for $k=1$.  Henceforth assume $k\geq2$.  Also, if $V=W$, the claim is immediate, so assume $V\neq W$.  Finally, for the remainder of this information-preservation proof, we replace the known upper bound $C$ by $2C$ and, for readability, continue to call the enlarged bound $C$.  Thus $\norm{w_j^\star}\leq C/2$ for every $j$; this harmless slack changes only absolute constants in the final $O((C/c)^{18})$ exponent.

    Fix parameters $V=[v_1,\dots,v_k]$ and $W=[w_1,\dots,w_k]$ {close to each other and $\Wstar$ in the following sense}
    \[
        \norm{V - W^\star}_F,\norm{W - W^\star}_F
        \leq  
            \frac{1}{10^6}\cdot 
            \inparen{\frac{c}{C}}^9 \cdot 
            \frac{1}{\sqrt{\log{\nfrac{ek}{c}}}}
            \,.
        \yesnum\label{eq:SS:distance}
    \] 
    The displayed radius is at most $C/2$.  Consequently, the triangle inequality and the slack introduced above imply $\norm{v_j},\norm{w_j}\leq C$ for every $j$, which is the candidate-norm bound used below.
    Fix $i$ to be any index satisfying 
    \[
        i
        \in 
        \argmax_{1\leq j\leq k} \norm{v_j - w_j}_2\,.
        \yesnum\label{eq:SS:i}
    \]
    {Due to the Pigeonhole Principle,
    \[
        \norm{v_i - w_i}_2 \geq \frac{1}{\sqrt{k}}\norm{V-W}_F\,.
        \yesnum\label{eq:SS:lowerboundonI}
    \]
    Consider any event $\evE$ over the draw of the covariate $x$ from a self-selection instance; we will specify $\evE$ later and its choice will depend on $V$ and $W$.
        We will sometimes abuse the notation and also treat $\evE$ as a subset of $\R^d$ which contains all realizations of covariates $x$ for which $\evE$ holds.
    
        \subsubsection{Reduction to ``Conditional'' Information Preservation}
        First, we show that it is sufficient to prove information preservation conditioned on event $\evE$ {(left child of the root in \Cref{fig:outline:SS:IP})}.
        {Recall that $\cM(W)$ denotes the distribution of $(x,\ymax)$ observed by the self-selection model with parameters $W \in \R^{d \times k}$.}
        \begin{lemma}[Reduction to ``Conditional'' Information Preservation]
            \label{lem:SS:IP:reductionToConditionalIP}
            For a measurable set $\evE\subseteq \R^d$ with $\Pr[\evE]>0$,
            \[
                \tv{\cM(V)}{~\cM(W)} 
                \geq 
                \Pr_{x\sim \normal{0}{I}}[x\in\evE]\cdot \tv{\cM(V\mid \evE)}{~\cM(W\mid \evE)}\,,
            \]
            where $\cM(V\mid \evE)$ (respectively $\cM(W\mid \evE)$) is the distribution of $(x,\ymax)\sim \cM(V)$ (respectively $(x,\ymax)\sim \cM(W)$) conditioned on $x\in \evE$. 
        \end{lemma}
        \begin{proof}
            Observe that 
            \begin{align*}
                \tv{\cM(V)}{~\cM(W)}~~
                &=~~ \sup_{{A}\sset \R^d{\times \R}: \text{$A$ measurable}} \abs{\cM(V)({A}) - \cM(W)({A})} \\
                &\geq~~ \sup_{{A}\sset \evE{\times \R}: \text{${A}$ measurable}} \abs{\cM(V)({A}) - \cM(W)({A})}\\ 
                &= ~~\tv{\cM(V\mid \evE)}{~\cM(W\mid \evE)}\cdot \Pr[\evE] \,. \qedhere 
            \end{align*}
        \end{proof}
        Thus, \cref{lem:SS:IP:reductionToConditionalIP} shows that to prove the information preservation claim in \cref{thm:SS:localConvexity} (\ie{}, \cref{eq:SS:informationPreservation}), it is sufficient to find event $\evE$ such that 
        \begin{align*}
            \tv{\cM(V\mid \evE)}{~\cM(W\mid \evE)}~~ 
            &\Stackrel{\eqref{eq:SS:lowerboundonI}}{\geq}~~
            \frac{
                \inparen{\frac{c}{ek}}^{{O({C^{18}/c^{18}}){}}}
            }{\Pr[\evE]}
            \cdot \norm{v_i - w_i}_2
            ~~\geq~~ \frac{1}{\sqrt{k}}\cdot\frac{
                \inparen{\frac{c}{ek}}^{{O({C^{18}/c^{18}}){}}}
            }{\Pr[\evE]} 
            \cdot \norm{V - W}_F \,.
            \yesnum\label{eq:SS:conditionalInfoPreservation}
        \end{align*}
    }
    \subsubsection{Definition of Event $\evE$ and its Properties}
    We now move on to the right child of the root in \Cref{fig:outline:SS:IP}.
    We will condition on the the event specified in \Cref{def:SS:event},
    which controls the length of projections of $x$ along certain directions determined by $V$ and $W$.
    For a matrix $W \in \R^{d \times k}$, we denote its columns by $w_1,\dots,w_k$ with $w_i \in \R^d$.
    \begin{restatable}{definition}{selfSelectionEvent}\label{def:SS:event}
        Fix an index $1\leq i\leq k$, 
        parameters $W,V\in \R^{d\times k}$, 
        constants $\gamma\in (0,\nfrac{1}{2}]$,
        and $R\geq 1$. 
        Let 
        $u\coloneqq v_i - {(v_i^\top \wh{w}_i}) \cdot \wh{w}_i$ be the component of $v_i$ orthogonal to $w_i$.
        When $u\neq0$, let $\wh u$ be its corresponding unit vector and define the tie-broken sign
        \[
            s\coloneqq
            \begin{cases}
                1, & \sinangle{v_i-w_i,\hat w_i}\sinangle{v_i-w_i,\hat u}\geq0,\\
                -1, & \sinangle{v_i-w_i,\hat w_i}\sinangle{v_i-w_i,\hat u}<0.
            \end{cases}
        \]
        In particular, $s\in\{\pm1\}$ even when the product vanishes.
        Define $\evE=\evE_{i,\gamma,R}$ to be the following event 
        \begin{align}
            \Biggl\{ x \in \R^d :\;
            1 ~\le~ \frac{x^\top w_i}{R\sqrt{\log\inparen{\nicefrac1\gamma}}}
            &\le~ \frac{1}{1-\frac{c}{4C^2}} \leq 2\,, 
            \label{eq:SS:event:large-component} \\
            1 ~\le~
            \frac{s\cdot x^\top \wh u}{(12C^3/c)\sqrt{\log\inparen{\nicefrac1\gamma}}}
            &\le~ \frac{1}{1-\frac{c}{4C^2}} \leq 2\,, 
            \label{eq:SS:event:large-component-u} \\
            \frac{\abs{x^\top w_j}}{R\sqrt{\log\inparen{\nicefrac1\gamma}}}
            &\leq 1-\frac{c^2}{12C^4} && \forall j\neq i\,,
            \label{eq:SS:event:small-component-w} \\
            \frac{\abs{x^\top v_j}}{R\sqrt{\log\inparen{\nicefrac1\gamma}}}
            &\leq 1-\frac{c^2}{12C^4} && \forall j\neq i\,,
            \label{eq:SS:event:small-component-v} \\
            \frac{\abs{x^\top (v_j - w_j)}}{(\nfrac{3}{c})R\sqrt{\log\inparen{\nicefrac1\gamma}}}
            &\leq \norm{v_j-w_j}_2 && \forall j\in [k] \Biggr\}\,.
            \label{eq:SS:event:small-difference-norm}
        \end{align}
        Further, in the special case where $v_i$ is parallel to $w_i$ (and, hence, $u=0$), we do not define $\wh u$ or $s$, and the definition of $\evE$ omits the bound on $x^\top \wh{u}$ (\Cref{eq:SS:event:large-component-u}).
    \end{restatable}
    This event, {which is inspired by the work of \citet{gaitonde2024selfselection},} is relevant to us for two reasons {which are outlined in the left sub-branch of the right branch of \cref{fig:outline:SS:IP}.}
    
    \paragraph{Property 1 ($\evE$ occurs with {$\poly(\nfrac{1}{k})$} probability).} 
        First, $\evE$ occurs with a constant probability for any fixed $\gamma$ and $R$.
        \begin{restatable}[]{lemma}{lemSSprobability}
            \label{lem:SS:probability}
            Fix $\gamma\in (0,\nfrac{1}{2}]$,
            $R\geq \max\!\left\{300\frac{C^8}{c^4}\sqrt{\log{k}},\,288\frac{C^8}{c^4}\right\}$,
            and parameters $V,W\in \R^{d\times k}$ satisfying \cref{eq:SS:distance}.
            Let $i$ be the index in \cref{eq:SS:i} maximizing the distance between $w_i$ and $v_i$.
            Then the corresponding event $\evE=\evE_{i,\gamma,R}$ (\cref{def:SS:event}) satisfies 
            \[
                \Pr[\evE] \geq 
                {
                    \frac14 \gamma^{9R^2/c^2}\exp\inparen{-2\ln\frac{24C^2}{c}}
                } 
                \geq 
                \inparen{\frac{c}{10C}}^4
                \gamma^{9R^2/c^2}\,.
            \]
        \end{restatable} 
        Eventually, we will set $\gamma = \nfrac{1}{e}$ and $R$ to be of order $\sqrt{\log k}$, which will imply that the event $\evE$ occurs with $\poly(\nfrac{1}{k})$ probability.
        \Cref{lem:SS:probability} is important as the largest information preservation constant we can deduce from \cref{eq:SS:conditionalInfoPreservation} is $\Pr[\evE]$: this is because when removing the conditioning on $\evE$ to deduce \cref{eq:SS:informationPreservation}, we lose a factor of $\Pr[\evE]$.
        \cref{lem:SS:probability} can be proved using standard tail bounds for Gaussian random variables and the fact that $x\sim \normal{0}{I}$; the proof appears in \cref{sec:proofof:lem:SS:probability}.

    \paragraph{Property 2 ($\evE$ implies ``separation'' between $i$ and other indices).}
            The second reason why $\evE$ is crucial is that, roughly speaking, conditioned on $\evE$, $y_{i,V}$ and $y_{i,W}$ are ``far'' from $y_{j,V}$ and $y_{j,W}$ for any $j\neq i$.
            To make this concrete, consider the following decomposition of $y_{j, V}$ and $y_{j, W}$ (for any $j\in [k]$)
            \begin{equation}
                \begin{split}
                  \hspace{-2.5mm}y_{j, V} 
                    &= \rho_{j,V} + {\xi_j}
                ~~\text{where}~~
                    ~\rho_{j,V}\coloneqq {\inangle{x, v_j}}
                    \,,\\
                \hspace{-2.5mm}y_{j, W} 
                    &= \rho_{j,W} + {\xi_j}
                ~\text{where}~~
                    \rho_{j,W}\coloneqq {\inangle{x, w_j}}
                    \,.
                \end{split}
                \label{eq:SS:decompositionOfY}
            \end{equation} 
            Note that conditioning on the event $\evE=\evE_{i,\gamma,R}$ changes the distribution of $\rho_{j,V}$ and $\rho_{j,W}$, but has no effect on the distribution of $\xi$, which is an independent random variable with $\xi\sim \normal{0}{I_k}$.
            Now, we explain the separation induced by $\evE$ more formally:
            At a high level, we will show that, conditioned on $\evE$, the following holds (where $i$ is the index in \cref{eq:SS:i} and $j$ is any other index) 
            \begin{align*}
                \begin{split}
                    \min_{A\in \inbrace{V,W}}~\rho_{i,A} - \rho_{j,A} &\geq \Omega\inparen{R}\,,\quad \\
                    \max_{\ell} \abs{\rho_{\ell,V} - \rho_{\ell,W}} &\leq O(R)\cdot \norm{V-W}_F\,, \quadand\\
                    \abs{\rho_{i,V} - \rho_{i,W}} &\geq 
                    \Omega(\norm{V-W}_F)\,.
                \end{split}
                \yesnum\label{eq:SS:informalPropertiesOfRho}
            \end{align*}   
            We pause here to make a subtle note; even after conditioning on event $\evE$, $\rho_{j, V}$ and $\rho_{j, W}$ are not deterministic (for all $1\leq j\leq k$).
            We will show that the random variables $\rho_{j, V}$ and $\rho_{j, W}$ satisfy \eqref{eq:SS:informalPropertiesOfRho} with probability 1 when conditioned on $\evE$.
            Next, let us discuss the usefulness of \cref{eq:SS:informalPropertiesOfRho}:
            \begin{itemize}
                \item 
                    For a large value of $R$, the first condition implies that $\rho_{i,V}$ and $\rho_{i,W}$ are significantly larger than $\rho_{j,V}$ and $\rho_{j,W}$ respectively.
                    Combined with properties of the noise distribution $\xi_j\sim \normal{0}{1}$, this implies that for $R\geq \Omega(1)$, $y_{i,V}$ (respectively $y_{i,W}$) is significantly larger than $y_{j,V}$ (respectively $y_{j,W}$) conditioned on $\evE$.
                    We will use this property to show that small changes in $\rho_{j,V}$ and $\rho_{j,W}$ lead to negligible changes in the total variation distance in \cref{eq:SS:conditionalInfoPreservation}.
                    \item The second condition implies that when $V$ and $W$ are close, \ie{}, when $\norm{V-W}_F\ll O(R)$, then $\rho_{j,V}$ and $\rho_{j,W}$ are close to each other for each $j$.
                    Since for any $j\neq i$, small changes between $\rho_{j,V}$ and $\rho_{j,W}$ do not affect the total variation distance in \cref{eq:SS:conditionalInfoPreservation} much, we can just focus on small changes in $\rho_{i,V}$ and $\rho_{i,W}$ (for the index $i$ in \cref{eq:SS:i}).
                    \item So far, we mentioned that due to the first two properties, to lower bound the total variation distance in \cref{eq:SS:conditionalInfoPreservation} it suffices to focus on the $i$-th coordinate.
                    The last condition shows that $\rho_{i,V}$ and $\rho_{i,W}$ are not too close to each other: their distance is at least $\Omega(\norm{V-W}_F)$.
                    This enables us to lower bound the total variation distance in \cref{eq:SS:conditionalInfoPreservation}.
                    It turns out that to obtain a lower bound that is linear in $\norm{V-W}_F$ (which is necessary to imply quadratic growth),
                    the lower bound on $\abs{\rho_{i,V}-\rho_{i,W}}$ must also be linear in $\Omega(\norm{V-W}_F)$.
            \end{itemize}
        The following result formalizes \cref{eq:SS:informalPropertiesOfRho}.
        Its proof relies on carefully applying \cref{asmp:SS:separationBoundedness} and using the properties implied by $\evE$; the complete proof appears in \cref{sec:proofof:lem:SS:propertiesOfRho}. 
        \begin{restatable}[Separation Properties of $\rho$]{lemma}{lemSSpropertiesOfRho}
            \label{lem:SS:propertiesOfRho} 
            Fix $\gamma\in (0,\nfrac{1}{2}]$,
            $R\geq \max\!\left\{300\frac{C^8}{c^4}\sqrt{\log{k}},\,288\frac{C^8}{c^4}\right\}$,
            and parameters $V,W\in \R^{d\times k}$ satisfying \cref{eq:SS:distance}.
            Let $i$ be the index in \cref{eq:SS:i} maximizing the distance between $w_i$ and $v_i$.
            The following guarantees hold with probability 1
            conditioned on the event $\evE= \evE_{i,\gamma,R}$ (\cref{def:SS:event}):
            \begin{enumerate}
                \item 
                    For any $j\neq i$,
                    $\rho_{i,V} - \rho_{j, V}\geq \frac{c^2}{24C^4}R \cdot \sqrt{\log{\nfrac{1}{\gamma}}}$ and 
                    $ \rho_{i, W} - \rho_{j, W}\geq  \frac{c^2}{24C^4}R \cdot  \sqrt{\log{\nfrac{1}{\gamma}}}$;
                \item For each $j$,
                    $\abs{\rho_{j, V} - \rho_{j, W}}\leq (\nfrac{3}{c})\,R \cdot \sqrt{\log{\nfrac{1}{\gamma}}}\cdot \norm{v_j - w_j}_2$; 
                \item Further, for the $i$ in \cref{eq:SS:i}, $\abs{\rho_{i,V} - \rho_{i,W}}\geq \frac{12C^2}{c} \sqrt{\log{\nfrac{1}{\gamma}}}\cdot \norm{v_i - w_i}_2$.
            \end{enumerate}
        \end{restatable}
    Now, we are ready to explain how we prove \cref{eq:SS:conditionalInfoPreservation}. 
    At a high level, {we will show that} the above lemma implies that the event $\ymax\leq \Delta$ (for some constant $\Delta$) has a very different likelihood with respect to $V$ and $W$,
    enabling us to lower bound the total variation distance in \cref{eq:SS:conditionalInfoPreservation}.
    
\subsubsection{Lower bound on Information Preservation Conditioned on Event $\evE$ }\label{sec:proofof:eq:SS:conditionalInfoPreservation}
    In this section, we prove \cref{eq:SS:conditionalInfoPreservation}, which we restate below:  
    \[
        \tv{\cM(V\mid \evE)}{~\cM(W\mid \evE)} \geq 
        \frac{1}{\Pr[\evE]}
        \cdot 
        {\inparen{\frac{c}{ek}}^{{O({C^{18}/c^{18}}){}}}}
        \cdot \norm{v_i - w_i}_2\,,
        \tag{Eq. \eqref{eq:SS:conditionalInfoPreservation} restated}
    \]
    where $\evE$ is the event from \cref{def:SS:event}.
    Recall that $\rho_{j,V}=\inangle{x,v_j}$ and $\rho_{j,W}=\inangle{x,w_j}$.
    Conditioning on $\evE$ bounds the {values $\rho_V =  \inparen{\rho_{1,V}, \dots, \rho_{k,V}}$ and $\rho_W = \inparen{\rho_{1,W}, \dots, \rho_{k,W}}$} to lie in a specific intervals.
    However, it does not fix their values and, therefore, $\rho_V$ and $\rho_W$ are not deterministic even after conditioning on $\evE$.
    To prove \cref{eq:SS:conditionalInfoPreservation}, we will show that for almost every realization of $x\in \evE$, it holds that
    \[
        \tv{\cM(V\mid \evE, x)}{~\cM(W\mid \evE, x)}
        \geq \frac{1}{\Pr[\evE]} \cdot 
        {\inparen{\frac{c}{ek}}^{{O({C^{18}/c^{18}}){}}}}
        \cdot \norm{v_i-w_i}_2\,.
    \]
    This is sufficient to prove \cref{eq:SS:conditionalInfoPreservation} due to the following standard common-marginal kernel equality {(which can be verified using a similar calculation as in the proof of \Cref{lem:SS:IP:reductionToConditionalIP})}.
    \[
        \tv{\cM(V\mid \evE)}{~\cM(W\mid \evE)}
        =
        \Ex_{x\mid \evE}\insquare{
            \tv{\cM(V\mid \evE, x)}{~\cM(W\mid \evE, x)}
        }\,.
    \]
    In the remainder of this section, we fix arbitrary values of $x\in \evE$.
    We remark that fixing $x$ also fixes the value of
    $\rho_V =  \inparen{\rho_{1,V}, \dots, \rho_{k,V}}$ and $\rho_W = \inparen{\rho_{1,W}, \dots, \rho_{k,W}}$ so that the remaining randomness in $\cM(V\mid \evE, x), \cM(W\mid \evE, x)$ is only from the noise $\xi\sim \normal{0}{I_k}$.
    We also do not explicitly mention the conditioning on $x$ and use $\cM(V\mid \evE)$ and $\cM(W\mid \evE)$ to denote $\cM(V\mid \evE,x)$ and $\cM(W\mid \evE,x)$ respectively.

    \subsubsection{Pre-Processing to Simplify Analysis}\label{sec:SS:preprocess}
    Now, observe that if we translate all of $\rho_{1,V},\dots,\rho_{k,V}$ and $\rho_{1,W},\dots,\rho_{k,W}$ by the same constant then the total variation distance between $\cM(V\mid\evE)$ and $\cM(W\mid\evE)$ remains unchanged,
    as the only randomness lies in the noise $\xi\sim \normal{0}{I_k}$.
    The translation may depend measurably on the fixed covariate $x$: it is the same measurable bijection of the response coordinate under both conditional kernels, and hence preserves their total variation distance.
    To simplify the arguments, we translate all $\rho_{1,V},\dots,\rho_{k,V}$ and $\rho_{1,W},\dots,\rho_{k,W}$ to ensure that 
    \[
        \min\inbrace{\rho_{i,V}, \rho_{i,W}} = 0\,.
        \yesnum\label{eq:SS:translation}
    \]
    With some abuse of notation, we continue to use $y_V$, $y_W$, $\rho_V$, and $\rho_W$ to refer to the corresponding vectors in the translated space.
    
    \subsubsection{Elementary Set Witnesses Information Perservation Conditional on $\evE$}
    Next, we move to the right sub-branch of the right branch of \cref{fig:outline:SS:IP}.
    Since $\tv{\cM(V\mid \evE)}{~\cM(W\mid \evE)}$ is defined as a supremum over all measurable sets $S\subseteq\R$
    \[
        \tv{\cM(V\mid \evE)}{~\cM(W\mid \evE)} 
        =
            \sup_{S\subseteq \R\,\text{ measurable}}
            \abs{
                \Pr\insquare{y_{\max, V} \in S\mid \evE}
                -
                \Pr\insquare{y_{\max, W} \in S\mid \evE}
            }\,,
        \yesnum\label{eq:SS:conditionalTVExpression}
    \]
    it suffices to prove that a specific set, in our case $\R_{\leq 0}$, witnesses that $\tv{\cM(V\mid \evE)}{~\cM(W\mid \evE)} $ is large; \ie{}, it is sufficient to prove the following, 
    \[
        \abs{
            \Pr\insquare{y_{\max, V} \leq 0\mid \evE}
            -
            \Pr\insquare{y_{\max, W} \leq 0\mid \evE}
        }
        \geq 
        \frac{1}{\Pr[\evE]}
        \cdot {\inparen{\frac{c}{ek}}^{{O({C^{18}/c^{18}}){}}}}
        \cdot \norm{v_i-w_i}_2\,.
        \yesnum\label{eq:SS:probabilityDiff}
    \]
    \renewcommand{\customcdf}[1]{\Phi(#1)}
    In the above, recall that we have fixed a covariate $x \in \evE$ and the only source of randomness comes from the noise $\xi \in \R^k.$
    To prove the desired bound, we compute expressions for the terms on the left-hand side:
    \begin{align*}
        \Pr\insquare{y_{\max, V} \leq 0\mid \evE}
        &= \prod_{j}~ 
            \Pr\insquare{{\xi_j}\leq -\rho_{j,V}\mid \evE}
        ~\hspace{0.25mm}= 
            \prod_{j}~ \customcdf{-\rho_{j,V}}
        \,,\\
        \Pr\insquare{y_{\max, W} \leq 0\mid \evE}
        &= \prod_{j}~ 
            \Pr\insquare{{\xi_j}\leq -\rho_{j,W}\mid \evE}
        = \prod_{j}~ \customcdf{-\rho_{j,W}}\,.
    \end{align*}
    Here $\Phi(\cdot)$ is the cumulative density function of the one-dimensional standard normal distribution.
    Note that we can factorize the CDF due to the fact that the noise $\xi\sim \normal{0}{I_k}$ has independent coordinates.
    Hence, we get the following decomposition mentioned in \cref{fig:outline:SS:IP}
    \begin{align*}
        \hspace{-2.5mm}
        &\abs{ 
            \Pr\insquare{y_{\max, V} \leq 0\mid \evE}
            - \Pr\insquare{y_{\max, W} \leq 0\mid \evE} 
        } \\
        &= 
        \underbrace{\inparen{
            \prod_{j}~ \customcdf{-\rho_{j,V}}
        }}_{\textsf{A}}
        \cdot  
        \biggabs{
            1 - 
            \underbrace{\frac{
                \customcdf{-\rho_{i,W}}
            }{
                \customcdf{-\rho_{i,V}}
            }}_{\textsf{B}}
            \cdot 
            \underbrace{\prod_{j\neq i}~ \frac{
                \customcdf{-\rho_{j,W}}
            }{
                \customcdf{-\rho_{j,V}}
            }}_{\textsf{C}}
        }\,.
        \yesnum\label{eq:SS:differenceAsProduct}
    \end{align*}
    {Next, following \cref{fig:outline:SS:IP}, we divide the remaining proof into two steps.}
    {The first step has three sub-parts, which} bound the Terms~\textsf{A}, \textsf{B}, and \textsf{C} in \cref{eq:SS:differenceAsProduct}.
    {The second step} uses these bounds to prove \cref{eq:SS:probabilityDiff} which implies \cref{eq:SS:conditionalInfoPreservation}.
    
    To bound Terms~\textsf{B} and \textsf{C}, we use the following result to control the ratios in the above equation.
    \begin{restatable}[Separating CDF Ratios]{lemma}{lemSScdfratios}\label{lem:SS:cdfratios}
        Fix $V,W$ satisfying \cref{eq:SS:distance}, let $i$ be as in \cref{eq:SS:i}, and let $\gamma,R$ satisfy the requirements of \cref{lem:SS:propertiesOfRho}.  For any $x\in\evE_{i,\gamma,R}$, apply the common translation from \cref{eq:SS:translation}.  Then, for every $j\neq i$,
        \begin{align*}
                \frac{
                    \customcdf{-\rho_{i,W}}
                }{
                    \customcdf{-\rho_{i,V}}
                }\,,~~\frac{
                    \customcdf{-\rho_{i,V}}
                }{
                    \customcdf{-\rho_{i,W}}
                }
                &~~\notin~~ 
                \insquare{
                    1 \pm 
                    \frac{1}{5}\min\inbrace{
                        1\,,
                        \frac{12C^2}{c} \sqrt{\log{\nfrac{1}{\gamma}}}\cdot \norm{v_i-w_i}_2 
                    }
                }\,,
                \yesnum\label{eq:SS:cdfratios:lb}\\
                \frac{
                    \customcdf{-\rho_{j,W}}
                }{
                    \customcdf{-\rho_{j,V}}
                }\,,~~
                \frac{\customcdf{-\rho_{j,V}}}{\customcdf{-\rho_{j,W}}}
                &~~\in~~
                \insquare{
                    1
                    \pm \frac{3R}{c} \sqrt{\log{\nfrac{1}{\gamma}}}\cdot 
                    \norm{v_j-w_j}_2
                    \cdot \gamma^{\frac{c^4}{2000C^8}\,R^2}
                }
                \yesnum\label{eq:SS:cdfratios:ub}
                \,.
        \end{align*}
    \end{restatable}
    The proof of \cref{lem:SS:cdfratios} is based on carefully analyzing moments of truncated Gaussians with one-sided truncation and is deferred to 
    \cref{sec:proofof:lem:SS:cdfratios}.
    In the remainder of the proof, we fix the following values of $R$ and $\gamma$ 
    \[
        R = 10^4  \inparen{\frac{C}{c}}^8 \sqrt{\log{\nfrac{k}{c}}}
        \qquadand
        \gamma = \frac{1}{e}\,.
        \yesnum\label{eq:SS:valueRandGamma}
    \] 
    \cref{lem:SS:probability} implies that
    \[
        \Pr[\evE] 
        \geq   \inparen{\frac{c}{10C}}^4 \cdot \gamma^{\frac{9R^2}{c^2}}
        \geq  \inparen{\frac{c}{10C}}^4 \cdot  \inparen{\frac{c}{k}}^{O((\sfrac{C}{c})^{18})}
        \,.
        \yesnum\label{eq:SS:IP:lowerBoundOnProb}
    \]  

    \paragraph{Step 1.1 (Bound on Term~\textsf{C}).}
    Define $D\coloneqq\max_{j\neq i}\norm{v_j-w_j}_2$.  By the choice of $i$, $D\leq\norm{v_i-w_i}_2$.
    First, we will simplify Term~\textsf{C} in \cref{eq:SS:differenceAsProduct}:
    The CDF Ratio Lemma (\cref{lem:SS:cdfratios}) along with the choice $\gamma=\nfrac{1}{e}$, implies that 
    \begin{align*}
        \textsf{C} 
        =
            \prod_{j\neq i}~ \frac{
                \customcdf{-\rho_{j,W}}
            }{
                \customcdf{-\rho_{j,V}}
        }
        \in
        \inparen{
            1
            \pm \frac{3R}{c}\cdot D \cdot \gamma^{\frac{c^4}{2000C^8}R^2}
        }^{k-1}
        \,.
    \end{align*}
    Let $a=c^4/(2000C^8)$.  The chosen value of $R$ satisfies $aR^2/2\geq R$.  Hence $R e^{-aR^2}\leq e^{-aR^2/2}$, and it follows that
    \[
        \textsf{C} 
        =
            \prod_{j\neq i}~ \frac{
                \customcdf{-\rho_{j,W}}
            }{
                \customcdf{-\rho_{j,V}}
        }
        \in
        \inparen{
            1
            \pm \frac{3}{c}~D \cdot e^{-\frac{c^4}{4000C^8}\,R^2}
        }^{k-1}\,.
    \]
    (Here, and in subsequent portions of this proof, we assume that $k\geq 2$. The edge case of $k=1$ can be handled separately: for $k=1$, the problem reduces to a standard linear regression problem for which strong convexity is well-known.)
    Substituting the value of $R$ from \cref{eq:SS:valueRandGamma} and using $\nfrac{c}{k}\leq \nfrac{1}{2}$ implies that
    \[
        \textsf{C} 
        =
            \prod_{j\neq i}~ \frac{
                \customcdf{-\rho_{j,W}}
            }{
                \customcdf{-\rho_{j,V}}
        }
        \in
        \inparen{
            1
            \pm \frac{c}{8k^2}\cdot D
        }^{k-1}\,.
        \yesnum\label{eq:SS:termC}
    \]
    Next, consider following standard inequalities 
    \begin{align*}
        \begin{split}
            \text{for all $k\in \N$ and $z\leq 1$\,,}\qquad 
                (1-z)^{k-1} &\geq 1-(k-1)z\,,\\
            \text{for all $0\leq z\leq \frac{1}{2(k-1)}$\,,}\qquad 
                \inparen{1+z}^{k-1} &\leq \frac{1}{1-(k-1)z}\leq 1+2(k-1)z\,.
        \end{split}
        \yesnum\label{eq:bernouliiInequality}
    \end{align*}
    To use the above inequalities, we need the following observation 
    \[
                D
                \cdot \frac{c}{8k^{2}}
        ~~~~\quad\Stackrel{\eqref{eq:SS:distance},~c\leq1,~k\geq2}{\leq}\quad~~~~ \frac{1}{2k}\,,
    \]
    Substituting the aforementioned inequalities with the above observation in \eqref{eq:SS:termC} implies 
    \[
        \textsf{C} 
            \in 1 \pm D
                \cdot {\frac{c}{4k}}\,.
\yesnum\label{eq:SS:termC:final}
    \] 
    
    \paragraph{Step 1.2 (Bound on Term~\textsf{B}).}
    Substituting the values of $R$ and $\gamma$ (\cref{eq:SS:valueRandGamma}) into the first part of \cref{lem:SS:cdfratios} implies that 
    \[
        \textsf{B} \notin 1 \pm
            \frac{1}{5}\cdot \min\inbrace{1, \frac{12C^2}{c}\sqrt{\log{\nfrac{1}{\gamma}}}\cdot \norm{v_i-w_i}_2}\,.
            \yesnum\label{eq:SS:termB}
    \]
    Observe that \cref{eq:SS:distance} and the value of $R$ and $\gamma$ (\cref{eq:SS:valueRandGamma}) implies that 
    \[
        \frac{12C^2}{c} \sqrt{\log{\nfrac{1}{\gamma}}}\cdot \norm{v_i-w_i}_2\leq 1\,.
    \]
    Therefore, the minimum in \cref{eq:SS:termB} always evaluates to the second term and, hence,  
    \[
        \textsf{B}\notin 1\pm  
            \frac{2C^2}{c}\cdot
                \sqrt{\log{\nfrac{1}{\gamma}}}\cdot \norm{v_i-w_i}_2\,.
    \]
    Since $C,\nfrac{1}{c}\geq 1$, it follows that
    \[
        \textsf{B}\notin 1\pm 2 \cdot \norm{v_i-w_i}_2\,.
        \yesnum\label{eq:SS:termB:final}
    \]
    
    \paragraph{Step 1.3 (Bound on Term~\textsf{A}).}
        Finally, we lower bound Term~\textsf{A} in \cref{eq:SS:differenceAsProduct}.
        Recall that after the translation we described at the start of \cref{sec:proofof:eq:SS:conditionalInfoPreservation}, we have that $\min\inbrace{\rho_{i,V}, \rho_{i,W}}=0$.
        Hence, the separation properties of $\rho$ (\cref{lem:SS:propertiesOfRho}), 
        the choice of $R, \gamma$ (\cref{eq:SS:valueRandGamma}), 
        and the proximity of $V, W, \Wstar$ (\cref{eq:SS:distance}) imply that 
        \begin{align*}
            \rho_{i,V}  
            &\leq  \frac{3R}{c}\sqrt{\log{\nfrac{1}{\gamma}}}\cdot \norm{v_i-w_i}_2
            \leq \frac{1}{10}\,,\\
            \rho_{j,V} &\leq \rho_{i,V} - \frac{c^2}{24C^4}R\cdot \sqrt{\log{\nfrac{1}{\gamma}}}
                \leq \frac{1}{10}
                    - 
                    \frac{c^2}{24C^4}\frac{10^4 C^8}{c^8}\sqrt{\log{\nfrac{k}{c}}}
                \leq \frac{-12C^4}{c^4}\sqrt{\log{\nfrac{k}{c}}}\,.
        \end{align*}
    Further, we have the following standard fact: for all $z>0$, $\Phi(z)\geq 1 - \frac{e^{-z^2/2}}{\sqrt{2\pi}z}$ (\cref{fact:GaussianTail}).
    Combining the above statements, it follows that 
    \begin{align*}
        \customcdf{-\rho_{j,V}}
            &\geq   1 
                    - \frac{1}{\sqrt{2\pi}}\cdot 
                        \frac{
                            e^{-\frac{144C^8}{2c^8}\cdot \log{\nfrac{k}{c}}}
                        }{
                            \frac{12C^4}{c^4} 
                            \sqrt{\log{\nfrac{k}{c}}}
                        }
            \geq 
                1 - 
                e^{-\frac{12C^8}{c^8}\cdot \log{\nfrac{k}{c}}}
            ~~\quad\Stackrel{(C\geq 1,c\leq 1)}{\geq}\quad~~
                1 - {\frac{1}{k^{4}}}\,,
                \yesnum\label{eq:SS:infoPreservation:toReferenceLater1}\\
        \customcdf{-\rho_{i,V}}&
            ~~\quad\Stackrel{(\rho_{i,V}\leq \sfrac{1}{10})}{\geq}\quad~~
            \customcdf{-\frac{1}{10}} 
            ~~>~~  \frac{1}{3}\,.
            \yesnum\label{eq:SS:infoPreservation:toReferenceLater2}
    \end{align*}
    Hence, it follows that 
    \[
        \textsf{A}
        = \prod_{j}~ \customcdf{-\rho_{j,V}}
        \geq 
        \frac{1}{3} 
        \inparen{1 - \frac{1}{k^{4}}}^{k-1}
        ~~\Stackrel{\eqref{eq:bernouliiInequality}}{\geq}~~ 
        \frac{1}{3}\inparen{1 - \frac{1}{k^{3}}}
        \quad\Stackrel{(k\geq 2)}{\geq}\quad \frac{1}{4}\,.
        \yesnum\label{eq:SS:termA:final}
    \]
    
    \paragraph{Step 2 (Completing the proof of \cref{eq:SS:conditionalInfoPreservation}).}
        In this step, we prove \cref{eq:SS:probabilityDiff} which {as proved above} implies \cref{eq:SS:conditionalInfoPreservation} {and, in turn, the information preservation claim in \cref{thm:SS:localConvexity}}.
        We use the following simple fact.
        \begin{fact}\label{fact:simpleInequality}
            Fix any $a,b\in [0,1]$ with $b\leq \nfrac{a}{10}$.
            For any $z_1\notin 1\pm a$ and $z_2\in 1\pm b$, $\abs{1 - z_1z_2}\geq \nfrac{a}{2}$.
        \end{fact}
        \begin{proof}[Proof of \cref{fact:simpleInequality}]
    {Let $\delta \coloneqq \abs{a}$ and $\eps\coloneqq \abs{b}$, so that $\eps\leq \nfrac{\delta}{10}$.
    If $\delta=0$, then the conclusion $\abs{1-z_1z_2}\geq \nfrac{\delta}{2}$ holds trivially, so assume $\delta>0$.
    Toward a contradiction, suppose that $\abs{1-z_1z_2} < \nfrac{\delta}{2}$.
    Since $z_2\in 1\pm b$, we have $z_2\in [1-\eps,\,1+\eps]$; in particular $z_2\geq 1-\eps>0$, so we may divide by $z_2$.
    The assumption $\abs{1-z_1z_2} < \nfrac{\delta}{2}$ implies
        $1-\nfrac{\delta}{2} ~<~ z_1z_2 ~<~ 1+\frac{\delta}{2},$ %
    and therefore
        $\frac{1-\nfrac{\delta}{2}}{1+\eps} ~<~ z_1 ~<~ \frac{1+\nfrac{\delta}{2}}{1-\eps}.$ %
    We now show that the above implies $z_1\in 1\pm a$, contradicting the assumption $z_1\notin 1\pm a$.
    \begin{itemize}
        \item For the upper bound, it suffices to prove
            $\frac{1+\nfrac{\delta}{2}}{1-\eps}\leq 1+\delta$
        This inequality is equivalent (after rearranging) to
            $\nfrac{\delta}{2} ~\ge~ (1+\delta)\eps,$
        which holds since $(1+\delta)\eps \leq 2\cdot (\nfrac{\delta}{10})=\nfrac{\delta}{5} \leq \nfrac{\delta}{2}$ (using $\delta\leq 1$).
        \item For the lower bound, it suffices to prove
            $\frac{1-\nfrac{\delta}{2}}{1+\eps}\geq 1-\delta.$ %
        This inequality is equivalent (after rearranging) to
            $\nfrac{\delta}{2} ~\ge~ (1-\delta)\eps,$ %
        which holds since $(1-\delta)\eps \leq \eps \leq \nfrac{\delta}{10} \leq \nfrac{\delta}{2}$.
    \end{itemize}
    Combining the two bounds yields $z_1\in [1-\delta,\,1+\delta]=1\pm a$, a contradiction.
    Hence $\abs{1-z_1z_2}\geq \nfrac{\delta}{2}=\nfrac{\abs{a}}{2}$.}
\end{proof}

        \noindent{}Apply the above fact with $a=2\norm{v_i-w_i}_2$ and $b=cD/(4k)$.  The warm-start bound gives $a\leq1$, while $D\leq\norm{v_i-w_i}_2$, $c\leq1$, and $k\geq2$ give $b\leq a/10$.  Thus, the fact combined with \cref{eq:SS:termC:final,eq:SS:termB:final} implies that
    \begin{align*}
        \abs{
            1 - 
            \frac{
                \customcdf{-\rho_{i,W}}
            }{
                \customcdf{-\rho_{i,V}}
            }
            \times 
            \prod_{j\neq i}~ \frac{
                \customcdf{-\rho_{j,W}}
            }{
                \customcdf{-\rho_{j,V}}
            }
        }
        \geq  
        \norm{v_i-w_i}_2\,.
        \yesnum\label{eq:SS:secondTermLB}
    \end{align*}
    Substituting this lower bound and the lower bound in \cref{eq:SS:termA:final} into \cref{eq:SS:differenceAsProduct} implies 
    \[
        \abs{
            \Pr\insquare{y_{\max, V} \leq 0\mid \evE}
            -
            \Pr\insquare{y_{\max, W} \leq 0\mid \evE}
        }
        \geq \frac{1}{4} \cdot \norm{v_i-w_i}_2\,.
    \]
    This implies \cref{eq:SS:probabilityDiff} due to \cref{eq:SS:IP:lowerBoundOnProb}.
    Finally, averaging the displayed pointwise bound over $x\mid\evE$ and applying \cref{lem:SS:IP:reductionToConditionalIP,eq:SS:lowerboundonI} gives
    \[
        \tv{\cM(V)}{\cM(W)}
        \geq \frac{\Pr[\evE]}{4}\norm{v_i-w_i}_2
        \geq \frac{\Pr[\evE]}{4\sqrt{k}}\norm{V-W}_F.
    \]
    Substituting \cref{eq:SS:IP:lowerBoundOnProb} and absorbing the factor $k^{-1/2}$ into the absolute constant in the exponent proves \cref{eq:SS:informationPreservation}.

\subsection{Local Convexity}\label{sec:SS:proof:localConvexity}
In this section, we prove the local convexity of the negative log-likelihood (\cref{thm:SS:localConvexity}). We begin by defining the negative log-likelihood and presenting its Hessian. 
Then, we prove local convexity in \cref{sec:SS:proof:localConvexity:main}; see \cref{fig:outline:SS:LC} for an outline of the proof.
\begin{figure}[ht]
    \centering
    \begin{tikzpicture}[node distance=2.5cm, auto]
        \node[myNodeNarrow, line width=1pt] (thm) at (0,0)
            {Local Strong Convexity};
        \node[myNode, text width=6.2cm, line width=1pt] (truth) at (-4cm,-2.2cm)
            {Information Preservation $\Longrightarrow$
             Strong Convexity at $W^\star$ (\cref{eq:SS:LC:hessianLB})};
        \node[myNode, text width=6.2cm, line width=1pt] (stable) at (4cm,-2.2cm)
            {Conditional Covariance Derivative and Moment Bounds
             $\Longrightarrow$ Population Hessian Stability
             (\cref{thm:SS:LC:formal})};
        \draw[-stealth, line width=0.5mm] (thm) -- (truth.north);
        \draw[-stealth, line width=0.5mm] (thm) -- (stable.north);
    \end{tikzpicture}
    \caption{Outline of the proof of local strong convexity for self-selection.}
    \label{fig:outline:SS:LC}
\end{figure}

\subsubsection{Negative Log-Likelihood and Its Hessian}
    Recall that given an observation $\ymax=m$, $y$ lies in the following coarse set (see \cref{eq:L-shapeSets})
    \[
    P_m \coloneqq \inbrace{z \in \R^k : \max_i z_i = m}\,,
    \]
    For an observation $(x,\ymax)$, the sample negative log-likelihood is
    \[
    \cL(W; x,\ymax) = -\log\int_{P_{\ymax}} \exp\Bigl(-\frac{1}{2}\|z-W^\top x\|_2^2\Bigr)\d z\,,
    \]
    and, hence, the (population) negative log-likelihood is given by
    \[
    \cL(W) = \Ex_{(x,\ymax)}\cL(W; x,\ymax)\,.
    \]
    The gradients and Hessians of the likelihood are as follows (see \cref{sec:hessian:SS} for a proof):
    \begin{align}
    \grad \cL(W; x,\ymax) &= xx^\top W - \Ex_{z\sim \normal{W^\top x}{I}}\Bigl[xz^\top \mid z\in P_{\ymax}\Bigr],\\[1mm]
    \grad \cL(W) &= W - {\Ex_{(x,\ymax)}}\Ex_{z\sim \normal{W^\top x}{I}}\bigl[xz^\top \mid z\in P_{\ymax}\bigr]\,,\\[1mm]
    \grad^2 \cL(W; x,\ymax) &= 
            I_{k}\otimes xx^\top
    - \cov_{z\sim \normal{W^\top x}{I}}\Bigl[z \mid z\in P_{\ymax}\Bigr]\otimes xx^\top\,, \label{eq:SS:conditionalHess:mainBody}\\[1mm]
    \grad^2 \cL(W) &= I_{dk} - \Ex_{(x,\ymax)}\inparen{\cov_{z\sim \normal{W^\top x}{I}}\bigl[z \mid z\in P_{\ymax}\bigr]\otimes xx^\top}\,.
    \end{align}

\subsubsection{Proof of Local Convexity}\label{sec:SS:proof:localConvexity:main}
        Our goal is to give an explicit radius $r_{\rm LC}=(c/(ek))^{O((C/c)^{18})}$ such that
        \[
            \norm{W-W^\star}_F\leq r_{\rm LC}
            \qquad\Longrightarrow\qquad
            \nabla^2\cL(W)\succeq0.
            \yesnum\label{eq:SS:LC:sufficientCondition}
        \]
    
    \paragraph{Strong Convexity at $W^\star$.}
    First, we will prove strong convexity of the NLL at the true parameter $W = W^\star$.
    \begin{figure}[bthp]
        \centering
        \includegraphics[width=0.3\linewidth]{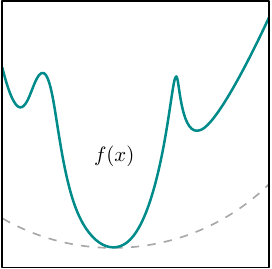}
        \caption{This figure illustrates a function $f$ {that} is {not convex}
        but {does} satisfy a quadratic local growth condition.
        {Indeed, $f$ is lower bounded by a quadratic function (shown by the dotted line).}
        }
        \label{fig:quadratic-growth}
    \end{figure}
    Let $\alpha_{\rm SS}$ denote a valid information-preservation coefficient in \cref{eq:SS:informationPreservation}, decreased if necessary so that $\alpha_{\rm SS}\leq1$; thus
    \[
        \alpha_{\rm SS}=\left(\frac{c}{ek}\right)^{O((C/c)^{18})}.
    \]
    By \cref{lem:model:informationPreservingImpliesGrowth}, throughout the information-preservation neighbourhood the quadratic branch of the minimum applies, and
    \[
        \cL(W)-\cL(W^\star)
        \geq2\alpha_{\rm SS}^2\norm{W-W^\star}_F^2.
    \]
    Fix a matrix $U$ with $\norm{U}_F=1$.  Since $W^\star$ is stationary (\cref{fact:SS:stationaryPoint}) and $\cL$ is twice differentiable,
    \begin{align*}
        \frac12\inangle{\operatorname{vec}(U),
            \nabla^2\cL(W^\star)\operatorname{vec}(U)}
        &=\lim_{s\downarrow0}
            \frac{\cL(W^\star+sU)-\cL(W^\star)}{s^2}\\
        &\geq2\alpha_{\rm SS}^2.
    \end{align*}
    Therefore
    \[
        \nabla^2\cL(W^\star)\succeq4\alpha_{\rm SS}^2I_{dk}.
        \yesnum\label{eq:SS:LC:hessianLB}
    \]
    This completes the left branch of \Cref{fig:outline:SS:LC}.

        We next record two moment estimates used in the Hessian-stability argument.
        \begin{fact}[Moments of Gaussian Maxima]\label{fact:GaussMaxMoments}
            There is an absolute constant $K_{\rm G}$ such that, for every
            $\ell\in\{1,\ldots,6\}$ and $\mu\in\R^k$,
            \[
                \Ex_{v\sim\normal{\mu}{I_k}}\norm{v}_\infty^\ell
                \leq K_{\rm G}\left(
                    (1+\log(2k))^{\ell/2}+\norm{\mu}_\infty^\ell
                \right).
            \]
        \end{fact}
        \begin{proof}
            Write $v=\mu+\xi$, where $\xi\sim\normal{0}{I_k}$, and put
            $M=\norm{\xi}_\infty$.  A union bound gives
            $\Pr[M>r]\leq2k e^{-r^2/2}$.  Let
            $a=2\sqrt{\log(2k)}$.  For $r\geq a$,
            $2k e^{-r^2/2}\leq e^{-r^2/4}$, and hence
            \begin{align*}
                \Ex M^\ell
                &=\int_0^\infty \ell r^{\ell-1}\Pr[M>r]\d r\\
                &\leq a^\ell+
                    \int_a^\infty \ell r^{\ell-1}e^{-r^2/4}\d r\\
                &\leq 2^\ell\log(2k)^{\ell/2}
                    +\ell2^{\ell-1}\Gamma(\ell/2)
                \leq2^\ell\log(2k)^{\ell/2}+384.
            \end{align*}
            The claim follows from
            $(s+t)^\ell\leq2^{\ell-1}(s^\ell+t^\ell)$ and
            $1+\log(2k)\geq1$.
        \end{proof}

        The following conditional-moment lemma is proved in
        \cref{sec:proofof:lem:SS:LC:unconditionalMoments}.
        \begin{lemma}[Conditional Gaussian Moments]
            \label{lem:SS:LC:unconditionalMoments}
            \label{lem:SS:LC:tailBoundII}
            Fix $x\in\R^d$, $m\in\R$, and $W\in\R^{d\times k}$, and let
            \[
                Q=\normal{W^\top x}{I_k}\mid P_m,
                \qquad R=1+|m|+\norm{W^\top x}_\infty.
            \]
            Then
            \[
                \Ex_{z\sim Q}\norm{z}_2^2\leq2kR^2,
                \qquad
                \Ex_{z\sim Q}\norm{z}_2^4\leq3k^2R^4.
            \]
            Consequently, for every unit vector $v\in\R^k$,
            \[
                \Ex_{z\sim Q}(v^\top z)^2\leq2kR^2,
                \qquad
                \Ex_{z\sim Q}(v^\top z)^4\leq3k^2R^4.
            \]
        \end{lemma}

    \begin{lemma}[Conditional Covariance Stability]\label{lem:SS:LC:step1}
        Fix $V\in\R^{d\times k}$ with $\norm{V}_F=1$, let
        $W_u=W^\star+uV$, and fix $0\leq t\leq1$.  Define
        \[
            \gamma_t(x,m)
            \coloneqq
            \sqrt{96}\,t k^{3/2}\norm{V^\top x}_2
            \left(
                1+|m|+
                \max_{0\leq u\leq t}\norm{W_u^\top x}_\infty
            \right)^3.
        \]
        Then, for every $x\in\R^d$ and $m\in\R$,
        \[
            -\gamma_t(x,m)I_k\otimes xx^\top
            \preceq
            \nabla^2\cL(W_t;x,m)-\nabla^2\cL(W^\star;x,m)
            \preceq
            \gamma_t(x,m)I_k\otimes xx^\top.
        \]
    \end{lemma}
    \begin{proof}
        For fixed $x,m$, let
        \[
            Q_u=\normal{W_u^\top x}{I_k}\mid P_m,
            \qquad h=V^\top x.
        \]
        This is an exponential family in $u$ with sufficient statistic
        $h^\top z$.  Hence, for every polynomial $g$,
        \[
            \frac{\d}{\d u}\Ex_{Q_u}g(z)
            =\cov_{Q_u}(g(z),h^\top z).
            \yesnum\label{eq:SS:LC:scoreIdentity}
        \]
        Differentiation under the integral is valid: on a compact interval of
        values of $u$, every differentiated integrand is dominated by a
        polynomial in $\norm z_2$ times $\exp(-\norm z_2^2/4)$.

        Fix a unit vector $a\in\R^k$.  Applying
        \cref{eq:SS:LC:scoreIdentity} to the first and second moments of
        $a^\top z$ gives
        \[
            \frac{\d}{\d u}\var_{Q_u}(a^\top z)
            =\cov_{Q_u}\left(
                (a^\top z-\Ex_{Q_u}a^\top z)^2,h^\top z
            \right).
        \]
        Put $R_u=1+|m|+\norm{W_u^\top x}_\infty$.  By
        \cref{lem:SS:LC:unconditionalMoments} and Jensen's inequality,
        if $h=0$ the desired derivative bound is immediate; otherwise apply the directional moment bounds below to $h/\norm h_2$.
        \begin{align*}
            \Ex_{Q_u}(a^\top z-\Ex_{Q_u}a^\top z)^4
            &\leq16\Ex_{Q_u}(a^\top z)^4
            \leq48k^2R_u^4,\\
            \var_{Q_u}(h^\top z)
            &\leq\Ex_{Q_u}(h^\top z)^2
            \leq2k\norm h_2^2R_u^2.
        \end{align*}
        Cauchy--Schwarz therefore yields
        \[
            \left|\frac{\d}{\d u}\var_{Q_u}(a^\top z)\right|
            \leq\sqrt{96}\,k^{3/2}\norm h_2R_u^3.
        \]
        Integrating from $0$ to $t$ and taking the supremum over unit $a$
        shows that
        \[
            \norm{\cov_{Q_t}(z)-\cov_{Q_0}(z)}_{\rm op}
            \leq\gamma_t(x,m).
        \]
        Finally, the sample Hessian formula gives
        \[
            \nabla^2\cL(W_t;x,m)-\nabla^2\cL(W^\star;x,m)
            =\left(\cov_{Q_0}(z)-\cov_{Q_t}(z)\right)\otimes xx^\top,
        \]
        proving both inequalities.
    \end{proof}

    \begin{lemma}[Population Covariance Bound]\label{lem:SS:LC:step2}
        Under \cref{asmp:SS:gaussianity,asmp:SS:separationBoundedness},
        there is an absolute constant $K_{\rm LC}$ such that, for
        every $V$ with $\norm V_F=1$ and every $0\leq t\leq1$, with
        $W_u=W^\star+uV$ and $\gamma_t$ as in
        \cref{lem:SS:LC:step1},
        \[
            0\preceq
            \Ex_{x,\ymax}\left[
                \gamma_t(x,\ymax)I_k\otimes xx^\top
            \right]
            \preceq K_{\rm LC}C^3k^{9/2}t\,I_{dk}.
        \]
    \end{lemma}
    \begin{proof}
        Let
        \[
            S=\operatorname{span}\left(
                \operatorname{cols}(W^\star)\cup\operatorname{cols}(V)
            \right),
            \qquad r=\dim S\leq\min\{d,2k\},
        \]
        and write $X=P_Sx$ and $Y=P_{S^\perp}x$.  The random vectors $X,Y$
        are independent standard Gaussians on their respective subspaces, and
        every scalar term in $\gamma_t(x,\ymax)$ depends on $x$ only through
        $X$.  Since $\norm V_F=1$, $C\geq1$, and $0\leq u\leq t\leq1$,
        \[
            \norm{V^\top x}_2\leq\norm X_2,
            \qquad
            \norm{W_u^\top x}_\infty\leq2C\norm X_2.
        \]
        Conditionally on $X$, write
        \[
            \ymax=\max_i\left((w_i^\star)^\top X+\xi_i\right),
            \qquad \xi\sim\normal{0}{I_k}.
        \]
        Then $|\ymax|\leq C\norm X_2+\norm\xi_\infty$.  By
        $(a+b+c)^3\leq9(a^3+b^3+c^3)$ and
        \cref{fact:GaussMaxMoments}, there is an absolute $K_0$ such that
        \[
            \Ex\left[
                \left(1+|\ymax|+
                \max_{0\leq u\leq t}\norm{W_u^\top x}_\infty\right)^3
                \middle|X
            \right]
            \leq K_0\left((1+\log(2k))^{3/2}+C^3\norm X_2^3\right).
        \]
        Consequently, $\Ex[\gamma_t(x,\ymax)\mid X]\leq G(X)$, where
        \[
            G(X)=\sqrt{96}K_0tk^{3/2}\norm X_2
            \left((1+\log(2k))^{3/2}+C^3\norm X_2^3\right).
        \]

        Conditionally on $X$, the scalar $\gamma_t(x,\ymax)$ is
        independent of $Y$.  Hence, conditionally on $(X,Y)$,
        \[
            \Ex[\gamma_t(x,\ymax)xx^\top\mid X,Y]
            \preceq G(X)xx^\top.
        \]
        Taking expectations reduces the desired bound to the following block
        calculation.

        In the orthogonal decomposition $S\oplus S^\perp$, independence and
        $\Ex Y=0$ give
        \[
            \Ex[G(X)xx^\top]
            =\begin{bmatrix}
                \Ex[G(X)XX^\top]&0\\
                0&\Ex[G(X)]I_{S^\perp}
            \end{bmatrix}.
        \]
        Its operator norm is at most
        \[
            \max\left\{\Ex[G(X)\norm X_2^2],\,\Ex[G(X)]\right\}.
        \]
        For an $r$-dimensional standard Gaussian,
        \[
            \Ex\norm X_2^4=r(r+2)\leq8k^2,
            \qquad
            \Ex\norm X_2^6=r(r+2)(r+4)\leq48k^3,
        \]
        and Jensen's inequality bounds the lower moments.  Together with
        $(1+\log(2k))^{3/2}\leq(2k)^{3/2}$, substitution gives
        \[
            \Ex[G(X)xx^\top]\preceq
            K_{\rm LC}C^3k^{9/2}t\,I_d.
        \]
        Tensoring with $I_k$ proves the claim.
    \end{proof}

    \begin{theorem}[Hessian Stability Near $W^\star$]
        \label{thm:SS:LC:formal}
        There is an absolute constant $K_{\rm LC}$ such that, for every
        $V\in\R^{d\times k}$ satisfying $\norm V_F=1$ and every $0\leq t\leq1$,
        with $W_t=W^\star+tV$,
        \[
            \norm{\nabla^2\cL(W_t)-\nabla^2\cL(W^\star)}_{\rm op}
            \leq K_{\rm LC}C^3k^{9/2}t.
        \]
    \end{theorem}
    \begin{proof}
        Average the two-sided Loewner bound in
        \cref{lem:SS:LC:step1} over the true distribution of
        $(x,\ymax)$ and apply \cref{lem:SS:LC:step2}.
    \end{proof}

    Define
    \[
        r_{\rm LC}
        \coloneqq
        \min\left\{1,
            \frac{2\alpha_{\rm SS}^2}{K_{\rm LC}C^3k^{9/2}}
        \right\}.
        \yesnum\label{eq:SS:LC:rLC}
    \]
    If $0<\norm{W-W^\star}_F\leq r_{\rm LC}$, apply
    \cref{thm:SS:LC:formal} with
    $t=\norm{W-W^\star}_F$ and
    $V=(W-W^\star)/t$.  Together with
    \cref{eq:SS:LC:hessianLB}, this gives
    \[
        \nabla^2\cL(W)\succeq2\alpha_{\rm SS}^2I_{dk}.
    \]
    The same conclusion is immediate at $W=W^\star$.  Finally,
    $r_{\rm LC}=(c/(ek))^{O((C/c)^{18})}$, so this proves the local-convexity
    claim in \cref{thm:SS:localConvexity}.

    \subsubsection{Proof of \cref{lem:SS:LC:unconditionalMoments} (Conditional Gaussian Moments)}\label{sec:proofof:lem:SS:LC:unconditionalMoments}
        \begin{proof}[Proof of \cref{lem:SS:LC:unconditionalMoments}]
            For consistency with the surrounding notation, write the arbitrary matrix and level in the lemma as $W_t$ and $\ymax$, respectively, and use $\Ex_t$ for expectation under the resulting conditional law.
            Divide $P_{\ymax}$ into $k$ parts $P_1,P_2,\dots,P_k$ where, for each $1\leq i\leq k$,
            \[
                P_i = \inbrace{ z\in \R^k \colon z_i = \ymax ~~\text{and}~~z_{-i}\leq \ymax}\,.
            \]
            See \cref{fig:self-selection-3D-distinct} for an illustration of the partition part of the parts $P_1,P_2,\dots,P_k$ with $k=3$.
            \begin{figure}[tbh!]
                \centering
                \includegraphics[width=0.5\linewidth]{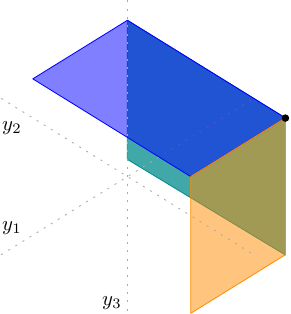}
                \caption{
                    This figure illustrates the partitions $P_1,P_2,\dots,P_k$ of the set $P_{\ymax}$ constructed in the proof of \cref{lem:SS:LC:unconditionalMoments} for $k=3$. 
                    Each colored slab in the figure corresponds to a set $P_i$ for $1\leq i\leq 3$, and the union of them corresponds to $P_{\ymax}$.
                    Compare this figure with \cref{fig:self-selection-3D} which illustrates $P_{\ymax}$.
                    \label{fig:self-selection-3D-distinct}
                }
            \end{figure} 
            Further, to simplify the notation, define 
            \[
                P\coloneqq P_{\ymax} 
                \qquadand
                \mu(t) \coloneqq W_t^\top x\,.
            \]
            Also recall that we use the short hands $\Ex_{t}[\cdot]$ and $\Pr_{t}[\cdot]$ to denote $\Ex_{z\sim \normal{\mu(t)}{I}}[\cdot]$ and $\Pr_{z\sim \normal{\mu(t)}{I}}[\cdot]$ respectively.
            First, observe that Cauchy--Schwarz gives
            \begin{align*}
                \begin{split}
                    \Ex\nolimits_t\insquare{\norm z_2^2\mid z\in P}
                    &=
                    \sum_i \Ex\nolimits_{t}\insquare{z_i^2 \mid z\in P}\,,\\
                    \Ex\nolimits_t\insquare{\norm z_2^4\mid z\in P}
                    &\leq 
                    k \sum_i  \Ex\nolimits_{t}\insquare{z_i^4 \mid z\in P}\,.
                \end{split} 
                \yesnum\label{eq:SS:LC:unconditionalMoment:CS}
            \end{align*}
            Where we use Cauchy--Schwarz in the last inequality. 
            Moreover, for any $1\leq i\leq k$,  
            \begin{align*}
                \begin{split}
                    \Ex\nolimits_{t}\insquare{z_i^2  \mid z\in P}
                &= 
                \sum_j
                \Pr\nolimits_{t}\insquare{z\in P_j\mid z\in P}
                \cdot \Ex\nolimits_{t}\insquare{z_i^2 \mid z\in P_j}
                \leq \max_j \Ex\nolimits_{t}\insquare{z_i^2 \mid z\in P_j}\,,\\
                \Ex\nolimits_{t}\insquare{z_i^4 \mid z\in P}
                &= 
                \sum_j
                \Pr\nolimits_{t}\insquare{z\in P_j\mid z\in P}
                \cdot \Ex\nolimits_{t}\insquare{z_i^4 \mid z\in P_j}
                \leq \max_{j} \Ex\nolimits_{t}\insquare{z_i^4 \mid z\in P_j} \,.
                \end{split}
                \yesnum\label{eq:SS:LC:unconditionalMoment:AcrossLegs}
            \end{align*} 
            Fix any $1\leq i, j\leq k$.
            In the remainder of the proof, we upper bound 
            \[
                \Ex\nolimits_{t}\insquare{z_i^2 \mid z\in P_j}
                \quadand
                \Ex\nolimits_{t}\insquare{z_i^{\smash{4}} \mid z\in P_j}\,.
            \]
            We divide the upper bound into two cases.
            \begin{enumerate}
                \item \textbf{Case A ($i=j$):}~~ In this case, $z_i=\ymax$ and, hence,
                \[
                    \Ex\nolimits_{t}\insquare{z_i^2 \mid z\in P_j} = \ymax^2
                    \quadand
                    \Ex\nolimits_{t}\insquare{z_i^{\smash{4}} \mid z\in P_j} = \ymax^4 \,.
                \]
                \item \textbf{Case B ($i\neq j$):}~~ Conditioned on $z\in P_j$, the variable $z_i$ has law
                \[
                    \truncatedNormal{\mu_i(t)}{1}{(-\infty,\ymax]};
                \]
                that is, $\normal{\mu_i(t)}{1}$ truncated to $(-\infty,\ymax]$.
                Next, we use the following fact which is proved in \cref{sec:proofof:fact:SS:LC:boundOnTruncatedMoments}.
                \begin{restatable}[]{fact}{factboundOnTruncatedMoments}
                    \label{fact:SS:LC:boundOnTruncatedMoments}
                    Fix $\mu,b\in \R$. It holds that 
                    \begin{align*}
                        \Ex_{z\sim \normal{\mu}{1}}\insquare{z^2\mid z\leq b}
                        &\leq 2\inparen{1+\abs{b}+\abs{\mu}}^2,\\
                        \Ex_{z\sim \normal{\mu}{1}}\insquare{z^{\smash{4}}\mid z\leq b}
                        &\leq 3\inparen{1+\abs{b}+\abs{\mu}}^4\,.
                    \end{align*}
                \end{restatable} 
                In our setting this implies the following bounds
                \begin{align*}
                    \Ex\nolimits_{t}\insquare{z_i^2 \mid z\in P_j} &\leq 
                        2 \inparen{1 + \abs{\ymax} + \abs{\mu_i(t)}}^2,\\
                    \Ex\nolimits_{t}\insquare{z_i^4 \mid z\in P_j} &\leq 
                        3 \inparen{1 + \abs{\ymax} + \abs{\mu_i(t)}}^4\,.
                \end{align*}
                Substituting $\mu_i(t)=(W_t^\top x)_i$ and $|\mu_i(t)|\leq \norm{W_t^\top x}_\infty$ implies that
                \begin{align*}
                    \Ex\nolimits_{t}\insquare{z_i^2 \mid z\in P_j} &\leq 
                        2 \inparen{1 + \abs{\ymax} + \snorm{W_t^\top x}_\infty}^2,\\
                    \Ex\nolimits_{t}\insquare{z_i^4 \mid z\in P_j} &\leq 
                        3 \inparen{1 + \abs{\ymax} + \snorm{W_t^\top x}_\infty}^4\,.
                \end{align*} 
            \end{enumerate}
            Substituting the bounds from both cases into \cref{eq:SS:LC:unconditionalMoment:CS,eq:SS:LC:unconditionalMoment:AcrossLegs} implies that 
            \begin{align*}
                 \Ex\nolimits_t\sinsquare{\norm z_2^2\mid z\in P} &\leq
                 2k\inparen{1 + \abs{\ymax} + \snorm{W_t^\top x}_\infty}^2\,,\\
                 \Ex\nolimits_t\sinsquare{\norm z_2^4\mid z\in P} &\leq
                 3k^2\inparen{1 + \abs{\ymax} + \snorm{W_t^\top x}_\infty}^4\,.
            \end{align*}
            The directional conclusions follow from $|v^\top z|\leq\norm z_2$ for every unit $v$.
        \end{proof}

\subsection{Projection Set}\label{sec:SS:proof:projectionSet}
Define
\[
    D_{\rm SS}\coloneqq\frac14\min\{r_{\rm IP},r_{\rm LC},1\}.
    \yesnum\label{eq:SS:DSS}
\]
Since $c,C$ are fixed, $D_{\rm SS}=(c/(ek))^{O((C/c)^{18})}$.
Given a $D_{\rm SS}$-warm start $W^{(0)}$, fix an ordering of
$W^\star$ for which $\norm{W^{(0)}-W^\star}_F\leq D_{\rm SS}$, and set
\[
    K\coloneqq\left\{W\in\R^{d\times k}:
        \norm{W-W^{(0)}}_F\leq D_{\rm SS}\right\}.
    \yesnum\label{eq:SS:projectionSet}
\]
Both $W^{(0)}$ and $W^\star$ belong to $K$.  Moreover, every $W\in K$
satisfies
\[
    \norm{W-W^\star}_F\leq2D_{\rm SS}
    \leq\min\{r_{\rm IP},r_{\rm LC}\},
    \qquad
    \max_i\norm{w_i}_2\leq C+2D_{\rm SS}\leq2C.
\]
Thus information preservation, local strong convexity, and the
stochastic-gradient moment bounds used below all hold throughout $K$.
Projection onto $K$ is explicit: for $A\neq W^{(0)}$,
\[
    \Pi_K(A)=W^{(0)}+
    \min\left\{1,\frac{D_{\rm SS}}{\norm{A-W^{(0)}}_F}\right\}
    (A-W^{(0)}),
\]
while $\Pi_K(W^{(0)})=W^{(0)}$.  It requires $O(dk)$ arithmetic
operations.

\subsection{Stochastic Gradient Oracle and Second Moment}\label{sec:SS:proof:gradientSecondMoment}
    The gradient of the negative log-likelihood function (\Cref{eq:SS:gradientExpression}) at $W\in K$ is given by
    \begin{align*}
        \grad \cL(W)
            = \Ex_x \Ex_{\ymax}
                \inparen{
                        xx^\top W
                        -   
                        \Ex_{z\sim \normal{W^\top x}{I} \mid z\in P(\ymax)}\insquare{xz^\top }
                }
            \,.
    \end{align*}
    
    \begin{lemma}\label{lem:SS:gradientOracle}
        Consider an instance of the max-self-selection problem
        and suppose \Cref{asmp:SS:gaussianity,asmp:SS:separationBoundedness} hold.
        Fix $W$ in the projection set $K$ (\Cref{eq:SS:projectionSet}).
        Suppose $(x, \ymax)$ is drawn from the max-self-selection model
        and $z\sim \normal{W^\top x}{I}\mid z\in P(\ymax)$.
        The following hold:
        \begin{enumerate} %
            \item $g(W)\coloneqq xx^\top W - xz^\top$ is an unbiased estimate of $\grad \negLL(W)$ \label{part:SS:gradientOracle:unbiased}
            \item $\E\insquare{\norm{g(W)}_F^2} \leq d\cdot O\inparen{k^4C^2 \log^3(2k)}$ \label{part:SS:gradientOracle:secondMoment}
            \item For every $\delta\in(0,1)$,
            there is an algorithm that consumes a single observation $(x, \ymax)$ 
            from the self-selection model
            to compute a random vector $\tilde g(W)$ such that $\tv{\tilde g(W)}{g(W)}\leq \delta$
            in time $O(kd\cdot \polylog(\nfrac{k}{\delta}))$. \label{part:SS:gradientOracle:algorithm}
        \end{enumerate}
    \end{lemma}

    \begin{proof}
        \sloppy
        \Cref{part:SS:gradientOracle:unbiased} follows by definition.
        The formal proof of \Cref{part:SS:gradientOracle:secondMoment} is through an involved calculation
        and we defer it to \cref{sec:SS:secondMoment}.

        For \cref{part:SS:gradientOracle:algorithm}, put
        $\mu=W^\top x$ and $a_j=\ymax-\mu_j$.  The exact mixture weight of
        the facet indexed by $i$ is
        \[
            p_i=\frac{q_i}{\sum_{r=1}^kq_r},
            \qquad
            q_i=\varphi(a_i)\prod_{j\neq i}\Phi(a_j),
        \]
        where $\varphi$ and $\Phi$ are the standard Gaussian density and CDF.
        Compute these weights in log space.  In particular, after computing
        $S=\sum_j\log\Phi(a_j)$, use
        \[
            \ell_i=\log q_i=\log\varphi(a_i)+S-\log\Phi(a_i)
        \]
        and stable log-sum-exp normalization.  Evaluating each log-CDF (and
        the elementary log-density) to additive precision
        $\delta/(32k)$ makes
        $|\widehat\ell_i-\ell_i|\leq\delta/8$.  It follows that
        $\widehat p_i/p_i\in[e^{-\delta/4},e^{\delta/4}]$, and hence, for
        $\delta\leq1$,
        \[
            \tv{\widehat p}{p}
            \leq\frac12(e^{\delta/4}-1)\leq\frac\delta4.
        \]
        Standard log-CDF evaluation therefore suffices with
        $O(k\polylog(k/\delta))$ arithmetic operations
        \citep[Proposition~3]{chevillard2012erf}.

        Draw $i$ from $\widehat p$, set $z_i=\ymax$, and independently sample
        each $z_j$, $j\neq i$, from
        $\normal{\mu_j}{1}$ conditioned on $z_j\leq\ymax$.  The following
        rejection sampler has acceptance probability at least $1/2$ uniformly
        in the truncation point.  Write $b=\ymax-\mu_j$.
        If $b\geq0$, draw $T\sim\normal{0}{1}$ until $T\leq b$.  If $b<0$,
        put $a=-b$ and
        \[
            \lambda=\frac{a+\sqrt{a^2+4}}2.
        \]
        Draw $U=a+E$ with $E\sim\operatorname{Exp}(\lambda)$ and accept it
        with probability $\exp(-(U-\lambda)^2/2)$; on acceptance set $T=-U$.
        The accepted density is proportional to $e^{-u^2/2}$ on
        $[a,\infty)$.  Moreover, the Mills-ratio bound
        \[
            \int_a^\infty e^{-u^2/2}\d u
            \geq\lambda^{-1}e^{-a^2/2}
        \]
        and $\lambda^2-a\lambda=1$ show that the acceptance probability is at
        least $e^{-1/(2\lambda^2)}\geq e^{-1/2}>1/2$.
        In either regime, set $z_j=\mu_j+T$.

        Cap each one-dimensional sampler after
        $L=\lceil\log_2(4k/\delta)\rceil$ proposals.  A union bound makes the
        probability that any coordinate reaches the cap at most $\delta/4$;
        on that event output an arbitrary point of $P(\ymax)$.  Apart from
        the categorical error, accepted samples have the exact target law.
        Thus the resulting conditional law of $z$ is within $\delta/2$ (and
        hence within $\delta$) in total variation of the exact conditional
        law, uniformly for every realized $(x,\ymax)$ and $W\in K$.  Computing
        $\mu$, sampling all coordinates, and forming
        $x(x^\top W-z^\top)$ takes
        $O(kd\,\polylog(k/\delta))$ time.
    \end{proof}
    More generally, consider an adaptive execution making at most $N$ oracle
    calls.  If, conditionally on every preceding transcript, the implemented
    oracle kernel is within $\beta$ in total variation of the exact kernel,
    sequential maximal coupling gives a coupling of the complete transcripts
    whose failure probability is at most $N\beta$.  We analyze only the
    conditionally unbiased exact oracle and take $\beta$ inversely
    proportional to the total number of calls; no unbiasedness or moment
    bound is asserted for the approximate oracle.

    \subsubsection{Bound on the Second Moment}\label{sec:SS:secondMoment}
        Fix $W\in K$ and let
        \[
            S=\operatorname{span}\!\left(
                \operatorname{cols}(W^\star)\cup\operatorname{cols}(W)
            \right),
            \qquad r=\dim S\leq\min\{d,2k\}.
        \]
        Write $X=P_Sx$ and $X_\perp=P_{S^\perp}x$.  These are independent
        standard Gaussians on their respective subspaces, and both
        $W^{\star\top}x$ and $W^\top x$ depend only on $X$.

        For fixed $(x,\ymax)$, \cref{lem:SS:LC:unconditionalMoments} gives
        \[
            \Ex_z\!\left[\norm z_2^2\mid z\in P(\ymax)\right]
            \leq2k\left(1+|\ymax|+\norm{W^\top x}_\infty\right)^2.
        \]
        Consequently,
        \begin{align}
            \Ex_z\!\left[\norm{W^\top x-z}_2^2\mid z\in P(\ymax)\right]
            &\leq2\norm{W^\top x}_2^2
                +12k\left(1+\ymax^2+\norm{W^\top x}_\infty^2\right).
            \label{eq:SS:secondMoment:conditional}
        \end{align}

        Conditionally on $X$,
        $\ymax=\max_i((w_i^\star)^\top X+\xi_i)$, where
        $\xi\sim\normal{0}{I_k}$.  Hence \cref{fact:GaussMaxMoments} implies,
        for an absolute constant $K_0$,
        \[
            \Ex[\ymax^2\mid X]
            \leq K_0\left(\log(2k)+C^2\norm X_2^2\right).
        \]
        Every column of $W$ has norm at most $2C$, so
        \[
            \norm{W^\top X}_\infty^2\leq4C^2\norm X_2^2,
            \qquad
            \norm{W^\top X}_2^2\leq4kC^2\norm X_2^2.
        \]
        Averaging \cref{eq:SS:secondMoment:conditional} over $\ymax$ therefore
        gives
        \[
            A(X)\coloneqq
            \Ex_{\ymax,z}\!\left[
                \norm{W^\top x-z}_2^2\mid z\in P(\ymax),X
            \right]
            \leq K_1k\left(1+\log(2k)+C^2\norm X_2^2\right)
        \]
        for an absolute constant $K_1$.

        Finally,
        $\norm{g(W)}_F^2=\norm x_2^2\norm{W^\top x-z}_2^2$.  Using
        \[
            \Ex\norm X_2^2=r,\qquad
            \Ex\norm X_2^4=r(r+2),\qquad
            \Ex\norm{X_\perp}_2^2=d-r,
        \]
        together with independence, we obtain
        \begin{align*}
            \Ex\norm{g(W)}_F^2
            &\leq K_1k\left[
                (1+\log(2k))d
                +C^2\bigl(r(r+2)+(d-r)r\bigr)
            \right]\\
            &=K_1k\left[(1+\log(2k))d+C^2(dr+2r)\right]\\
            &\leq K_2C^2dk^2
            \leq d\cdot O\!\left(k^4C^2\log^3(2k)\right).
        \end{align*}
        This holds uniformly for every $W\in K$ and for all $d,k\geq1$,
        proving \cref{part:SS:gradientOracle:secondMoment}.

\subsection{Projected Stochastic Gradient Descent}\label{sec:SS:proof:PSGD}
    We are now ready to prove \Cref{thm:SS:localConvergence} (restated below) by applying the iterative PSGD algorithm (\Cref{thm:SGD-local-growth-convergence}).
    Our complete algorithm (\Cref{thm:SS}) follows by combining this result with an appropriate warm start.
    \thmSelfSelectionLocalConvergence*

    \begin{proof}
        Let $D_{\rm SS}$ and $K$ be as in
        \cref{eq:SS:DSS,eq:SS:projectionSet}, and fix the ordering of
        $W^\star$ matched by the supplied warm start.  Condition on
        this warm start; throughout the proof, all oracle observations come
        from the fresh independent batch in the theorem statement.  By
        \cref{thm:SS:localConvexity,eq:SS:LC:rLC}, $\cL$ is convex on $K$.
        Let $\alpha_{\rm SS}$ be the information-preservation coefficient in
        \cref{eq:SS:informationPreservation}, chosen no larger than one, and
        set
        \[
            \eta=\sqrt2\,\alpha_{\rm SS}.
        \]
        Every point of $K$ is in the information-preservation neighbourhood
        and is within distance one of $W^\star$.  Thus
        \cref{lem:model:informationPreservingImpliesGrowth} gives, throughout
        $K$,
        \[
            \cL(W)-\cL(W^\star)
            \geq\eta^2\norm{W-W^\star}_F^2.
            \yesnum\label{eq:SS:PSGD:growth}
        \]
        In particular, $\cL$ satisfies an $(\eta,2)$-local growth condition on
        $K$; note that $\eps_{\rm obj}<2$ below.

        Let $G^2=d\cdot O(k^4C^2\log^3(2k))$ be the uniform second-moment
        bound in \cref{lem:SS:gradientOracle}.  The initial objective gap is
        dimension-free.  Indeed, the likelihood identity and data processing
        give
        \begin{align*}
            \cL(W^{(0)})-\cL(W^\star)
            &=\kl{\cM(W^\star)}{\cM(W^{(0)})}\\
            &\leq\Ex_x\kl{
                \normal{W^{\star\top}x}{I_k}}
                {\normal{W^{(0)\top}x}{I_k}}\\
            &=\frac12\Ex_x\norm{(W^{(0)}-W^\star)^\top x}_2^2
            \leq\frac{D_{\rm SS}^2}{2}\eqqcolon\eps_0.
        \end{align*}
        For one base run, set
        \[
            \eps_{\rm obj}=\frac{\eta^2\eps^2}{9}.
        \]
        If $\eps_0\leq\eps_{\rm obj}$, \cref{eq:SS:PSGD:growth} shows that
        the warm start is already $\eps/3$-close to $W^\star$.  Otherwise,
        apply \cref{thm:SGD-local-growth-convergence} with objective accuracy
        $\eps_{\rm obj}$ and the exact oracle.  It uses at most
        \[
            M=O\!\left(
                \left(1+\frac{G^2}{\eta^2\eps_{\rm obj}}\right)
                \log^3\!\left(2+\frac{\eps_0}{\eps_{\rm obj}}\right)
            \right)
            =\widetilde O\!\left(\frac{d\,\poly(k)}{\eps^2}\right)
            =O(d)\poly\!\left(k,\frac1\eps\right)
        \]
        fresh observations and, with probability at least $0.99$, returns
        $\widehat W$ satisfying
        \[
            \norm{\widehat W-W^\star}_F
            \leq\frac{\sqrt{\eps_{\rm obj}}}{\eta}=\frac\eps3.
        \]
        The last equality has no hidden dependence on $d$: both
        $\eps_0=D_{\rm SS}^2/2$ and $\eps_{\rm obj}$ are dimension-free, so
        $\log(2+\eps_0/\eps_{\rm obj})=O(\log(k/\eps))$ for fixed $c,C$.

        Fix a known deterministic integer $M\geq1$ that upper
        bounds the number of calls in a base run and has the displayed
        asymptotic order; use the same bound in the branch that makes no
        oracle calls.
        To obtain confidence $1-\delta$, take
        $q=\Theta(\log(2/\delta))$ independent base runs and implement every
        oracle call to total-variation accuracy
        \[
            \beta=\frac{\delta}{8qM}.
        \]
        The sequential coupling following \cref{lem:SS:gradientOracle}
        couples all $qM$ implemented calls to their exact counterparts except
        with probability at most $\delta/8$.  A Chernoff bound shows that,
        except with probability at most $\delta/4$, strictly more than half of
        the exact-run outputs are $\eps/3$-close to $W^\star$.

        For matrices $A,B$, define
        \[
            d_{\rm perm}(A,B)
            \coloneqq\min_{\pi\in\mathfrak S_k}\norm{A-BP_\pi}_F,
        \]
        which is a metric on matrices modulo column permutations.  Return any
        output having strictly more than $q/2$ outputs within
        $d_{\rm perm}$-distance $2\eps/3$.  A good output has this property,
        and the chosen output's neighbourhood must intersect the set of good
        outputs.  The triangle inequality therefore places the returned
        matrix within $\eps$ of $W^\star$, up to permutation.  If
        no output has the required majority, return an arbitrary base-run
        output; this fallback is used only on the failure event.

        The total number of samples is
        \[
            qM=\widetilde O\!\left(
                \frac{d}{\eps^2}\log\frac1\delta
            \right)\poly(k)
            =O(d)\poly\!\left(k,\frac1\eps,\log\frac1\delta\right).
        \]
        Each implemented query, including projection onto $K$, costs
        $O(kd\,\polylog(qM/\delta))$ time.  Computing $d_{\rm perm}$ is a
        minimum-weight assignment problem and is polynomial in $d,k$.
        Hence the total running time is
        \[
            \widetilde O\!\left(
                \frac{d^2}{\eps^2}\log\frac1\delta
            \right)\poly(k),
        \]
        as claimed.
    \end{proof}

    \begin{proof}[Proof of \cref{thm:SS}]
        On an independent sample batch, run the algorithm of
        \citet{gaitonde2024selfselection} to maximum per-column error
        $D_{\rm SS}/(3\sqrt{k})$.  A successful run has Frobenius error at
        most $D_{\rm SS}/3$, up to a permutation.  Amplify its constant
        success probability to $1-\delta/2$ by independent repetitions and
        the same permutation-invariant clustering argument used above,
        with an arbitrary-output fallback if no majority exists.  The
        resulting estimate is a $D_{\rm SS}$-warm start; fix a matching
        ordering of $W^\star$.

        Conditional on this event, apply
        \cref{thm:SS:localConvergence} with failure probability $\delta/2$ to
        a fresh, independent sample batch.  Fresh samples ensure that,
        conditional on the computed warm start, every exact stochastic
        gradient has the law used in the local analysis.  A union bound gives
        total success probability at least $1-\delta$.

        Because $D_{\rm SS}=k^{-O(1)}$ for fixed $c,C$, the warm-start phase
        uses $\widetilde O(d)\poly(k,\log(1/\delta))$ samples and time
        \[
            \poly(d,k,\log(1/\delta))+(k\log k)^{O(k)}.
        \]
        Adding the local-convergence bounds proves \cref{thm:SS}.  A
        multiplicative $\log(1/\delta)$ from warm-start amplification is
        absorbed into this displayed sum, since
        $A\log(1/\delta)\leq A^2+\log^2(1/\delta)$ for
        $A=(k\log k)^{O(k)}$.
    \end{proof}

%% file: gradient_computation.tex
    \label{sec:hessian:SS}
        Consider an estimate $W\in \R^{d\times k}$ of the parameter $W^\star$ defining the self-selection instance.
        Given an observation $(x,\ymax)$, the likelihood of the estimate $W$ (conditioned on a fixed value of $x$ and $\ymax$) can be written as the likelihood of observing ($x,\ymax$).
        In particular,
        {ignoring normalizing constants,
        we have}
        \[
            {\Pr_{z\sim \normal{W^\top x}{I}} \insquare{\max_i z_i = \ymax}
            \propto} 
            \int_{\max_i {z_i} = \ymax} \exp\inparen{-\frac{1}{2}\snorm{z- W^\top x}_2^2} \d z\,.
        \]
        Here, the integral is over a set of the following form:
        given a value $v\in \R$, define
        \[
            P(v)\coloneqq \inbrace{z\in \R^k\colon \max_i z_i = v}\,.
        \]
        Hence, the negative log-likelihood of observing $(x,\ymax)$ is 
        \[
            \cL\inparen{W; x,\ymax} = -\log\int_{P({\ymax})} \exp\inparen{-\frac{1}{2}\snorm{z- W^\top x}_2^2} \d z\,.
        \]
        Thus, the population negative log-likelihood is
        \[
            \cL\inparen{W} 
            = \Ex_{x,\ymax}\cL\inparen{W; x,\ymax} 
            = -\Ex_{x,\ymax}\log\int_{P({\ymax})} \exp\inparen{-\frac{1}{2}\snorm{z- W^\top x}_2^2} \d z\,.
        \]
        One of the most important properties of the negative log-likelihood is that the true parameters $\Wstar$ is a stationary point of $\cL(\cdot)$.
        \begin{restatable}[]{fact}{SSStationary}\label{fact:SS:stationaryPoint}
            It holds that $\grad \cL(W^\star)=0$.    
        \end{restatable}
        \paragraph{Gradient of $\cL(\cdot)$.}
        Next, we compute $\grad\cL(\cdot)$, which, in particular, enables us to verify \cref{fact:SS:stationaryPoint}.
        \begin{fact}[Direct Calculation]
            It holds that 
            \[
                \grad \cL(W)
                = W - \Ex_{x, \ymax}
                    {
                        \Ex_{z\sim \normal{W^\top x}{I} {\mid z\in P(\ymax)}}\insquare{xz^\top }
                    }
                    \,.
                \yesnum\label{eq:SS:gradientExpression}
            \]
        \end{fact}
        \begin{proof}
            The expression follows by verifying the following expressions via direct computation and using that $\Ex[xx^\top]=I$
            \begin{align*}
                \grad \cL(W; x,\ymax)
                    &~~=~~ \frac{1}{2}\Ex_{z\sim \normal{W^\top x}{I} \mid z\in P(\ymax)}{
                        \grad_W \snorm{z-W^\top x}_2^2
                    }\,,\\
                \grad_W \snorm{z-W^\top x}_2^2 
                    &~~=~~ -2\inparen{xz^\top -xx^\top W}\,.\qedhere
            \end{align*}
        \end{proof}
        Now we are ready to prove \cref{fact:SS:stationaryPoint}.
        \begin{proof}[Proof of \cref{fact:SS:stationaryPoint}]
            This holds because $\Pr[\ymax = m] = \normalMass{{W^\star}^\top x}{I}{{P(m)}}$ and, conditioned on $(x,\ymax)$, 
            \[
                \Pr[z=v\mid x,\ymax]= \frac{\mathds{1}_{P({\ymax})}(v)}{\normalMass{{W^\star}^\top x}{I}{P({\ymax})}}\cdot \normalMass{{W^\star}^\top x}{I}{v}\,.  
            \]
            Therefore, it follows that 
            \[
                \Pr[z=v\mid x] = \normalMass{{W^\star}^\top x}{I}{v}\,.  
            \]
            Substituting this in the expression for $\grad\cL(\cdot)$ implies that
            \[
                \grad \cL(W^\star) 
                = W^\star - \Ex_x \insquare{x \inparen{\Ex_{z\sim \normal{\Wstar^\top x}{I}} z^\top}}
                = \Wstar - \Ex_{x}\inparen{ xx^\top W^\star} = 0\,.\qedhere
            \]
        \end{proof}
        
        \paragraph{Hessian of $\cL(\cdot)$.}
        Next, we compute the Hessian of $\cL(\cdot)$. 
        Differentiating $\grad \cL(\cdot)$ once more, gives us the following expression
        \begin{align*}
            \grad^2 \cL(W)
                &= \grad_W \inparen{
                    W - \Ex_{x, \ymax}
                        {
                            \Ex_{z\sim \normal{W^\top x}{I} \mid z\in P(\ymax)}\insquare{xz^\top }
                        } 
                }
                \\
                &= I_{dk}
                    -
                    \Ex_{x,\ymax}\grad_W \inparen{
                            \Ex_{z\sim \normal{W^\top x}{I} }\insquare{
                        xz^\top\mid z\in P({\ymax})
                    } 
                }
                \,.
                \yesnum\label{eq:hessian:SS:1}
        \end{align*}
        Next, for a fixed $(x,\ymax)$,  we compute the above gradient.
        To simplify the calculations, let us define (for fixed $x$ and $\ymax$)
        \[
            {f(W) = f(W; z)}\coloneqq \snorm{z - W^\top x}_2^2\,.
        \]
        {For the purpose of this calculation,
        we identiy $f\colon \R^{dk}\to \R$ as a function of the flattened vector.}
        Now, we can re-write the aforementioned flattened expectation 
        \[
            \Ex_{z\sim \normal{W^\top x}{I} }\insquare{
                        (xz^\top)^\flat\mid z\in P({\ymax})
            }
            = \frac{
                \int_{P({\ymax})} \inparen{xz^\top}^\flat e^{-\frac{1}{2}f(W)} \d z
            }{
                \int_{P({\ymax})} e^{-\frac{1}{2}f(W)} \d z
            }\,.
        \]
        With this expression in hand, we can compute the gradient as follows
        \begin{align*}
           &\grad_W \frac{
            \int_{P({\ymax})} (xz^\top)^\flat e^{-\frac{1}{2}f(W)} \d z 
           }{
            \int_{P({\ymax})} e^{-\frac{1}{2}f(W)} \d z 
           }\\
           &\quad= 
           -\frac{
               \int_{P({\ymax})} (xz^\top)^\flat \grad f(W)^\top e^{-\frac{1}{2}f(W)} \d z 
           }{
                2\int_{P({\ymax})} e^{-\frac{1}{2}f(W)} \d z 
           }
           +
           \frac{
                \inparen{
                    \int_{P({\ymax})} (xz^\top)^\flat e^{-\frac{1}{2}f(W)} \d z} 
                \inparen{
                    \int_{P({\ymax})} \grad f(W)^\top e^{-\frac{1}{2}f(W)} \d z 
                }
           }{
                2\inparen{\int_{P({\ymax})} e^{-\frac{1}{2}f(W)} \d z }^2
           }
           \,.
        \end{align*}
        Substituting $\grad f(W) = 2(xx^\top W- xz^\top)$ implies that 
        \begin{align*}
           \grad_W \frac{
            \int_{P({\ymax})} {(xz^\top)}^\flat e^{-\frac{1}{2}f(W)} \d z 
           }{
            \int_{P({\ymax})} e^{-\frac{1}{2}f(W)} \d z 
           }
           &= 
           \Ex_z\insquare{{(xz^\top)}^\flat {\sinparen{(xz^\top)^\flat - (xx^\top W)^\flat}}^\top \mid z\in P({\ymax})}\\
           &\quad - \Ex_z\insquare{(xz^\top)^\flat \mid z\in P({\ymax})}
           ~\Ex_z\insquare{
                {(xz^\top)^\flat - (xx^\top W)^\flat}
                \mid z\in P({\ymax})
           }^\top\,.
        \end{align*}
        Since $xx^\top W$ is a constant with respect to $z$, the above simplifies to 
        \begin{align*}
            \grad_W \frac{
             \int_{P({\ymax})} {(xz^\top)}^\flat e^{-\frac{1}{2}f(W)} \d z 
            }{
             \int_{P({\ymax})} e^{-\frac{1}{2}f(W)} \d z 
            }
            &= \Cov\insquare{{(xz^\top)}^\flat\mid z\in P({\ymax})} \\
            &= \Cov\insquare{z\otimes x\mid z\in P({\ymax})} \\
            &= \Cov\insquare{z\mid z\in P({\ymax})}\otimes xx^\top
            \,.
        \end{align*}
        Substituting this in \cref{eq:hessian:SS:1} implies that 
        \[
            \grad^2 \cL(W) =  I_{dk} 
                - \Ex_{x,~\ymax}\insquare{\Cov\insquare{z\mid z\in P({\ymax})}\otimes xx^\top}\,.
        \]

%% file: quadratic_growth.tex
In this section,
we state and prove the analysis for the variant of gradient descent we use to optimize the various likelihood functions in our work.
Consider a lower-semicontinuous convex function $F\colon  K\to \R$ with a global minimizer $w^\star$
on a nonempty, closed, convex subset $K\sset \R^d$.
We write
\[
    S_\rho \coloneqq \inbrace{w\in K\colon F(w) - F(w^\star)\leq \rho}
\]
to denote the \emph{$\eps$-sublevel set} of a function $F$
where $F$ is clear from context.

We now state a useful definition.
\begin{definition}[$\eta$-Local Growth Condition]\label{def:local-growth}
    We say that $F\colon K\to \R$ satisfies an \emph{$(\eta, \rho)$-local growth condition} if
    \[
        \norm{w - w^\star}_2
        \leq \frac{(F(w) - F(w^\star))^\frac12}{\eta}
    \]
    for every $w\in S_{\rho}$.
\end{definition}
We remark that a function satisfying a $(\eta_0, \rho_0)$-local growth condition
also satisfies $(\eta, \rho)$-local growth for every $\eta\in (0, \eta_0], \rho\in (0, \rho_0]$.
We also note that by \Cref{thm:quadGrowth},
    our log-likelihood functions for $\alpha$-information preserving distortion mechanisms at radius $R$ satisfy a $(\sqrt2 \alpha, 2)$-local growth condition
    {if $K\sset B_R(w^\star)$ and $\alpha R\leq1$}.

Recall the following standard convergence result from convex optimization.
\begin{theorem}[Theorem 9.7 in \cite{garrigos2023handbook}]\label{thm:SGD-convergence}
    Let $F\colon S\sset \R^d\to \R$ be differentiable and convex, where $S$ is nonempty, closed, and convex, and let $w\in S$.
    Suppose the oracle uses fresh randomness and, for the natural filtration
    $\mathcal F_t$, satisfies
    $\E[g_t\mid\mathcal F_t]=\nabla F(w_t)$ and $\E[\norm{g_t}_2^2\mid\mathcal F_t]\leq G^2$.
    Then PSGD projected onto $S$ with initial point $w^{(0)}\in S$
    and constant step-size $\gamma > 0$ 
    outputs an average iterate $\bar w$ satisfying
    \[
        \E\insquare{F(\bar w) - F(w)}
        \leq \frac{\gamma G^2}{2} + \frac{\snorm{w^{(0)} - w}_2^2}{2\gamma T}\,.
    \]
\end{theorem}
One straightforward way to recover the parameters from optimizing a function satisfying a $(\eta, \eps)$-local growth condition
is to find an $\eta^2\eps^2$-optimal point using \Cref{thm:SGD-convergence}.
This yields a $O(\nfrac1{\eta^4 \eps^4})$ rate of convergence which may not be desirable.
We design a novel iterative refinement algorithm that is able to leverage the local growth condition in a more clever way.
Our algorithm is inspired by \citet{xu2019accelerate} but their results only hold under the restrictive condition on the gradient oracle that there is a deterministic bound $\norm{g(w)}\leq G$.
This assumption is necessary to obtain high probability bounds in their algorithm.
On the other hand,
our algorithm will converge with constant probability
but relies on a much weaker bound on the second moment of the gradient oracle.

We begin with a key insight due to \citet{yang2018beating} as stated in \cite{xu2019accelerate}.
For the sake of analysis only,
it will be useful to define the following notation with respect to $F$.
\[
    w_\rho^\dagger
    \coloneqq \argmin_{v\in S_\rho} \norm{v-w}_2^2\,.
\]
\begin{proposition}[Lemma 1 in \cite{xu2019accelerate}]\label{prop:level-set-dist}
    Suppose $F\colon K\to\R$, where $K$ is nonempty, closed, and convex, is lower-semicontinuous and convex and satisfies a $(\eta, \rho)$-local growth condition.
    Then for any $w\in K$,
    \[
        \snorm{w-w_\rho^\dagger}_2
        \leq \frac{F(w) - F(w_\rho^\dagger)}{\eta \sqrt{\rho}}\,.
    \]
\end{proposition}
\begin{proof}
    Lower semicontinuity makes $S_\rho$ closed, so the projection exists.
    If $w\in S_\rho$, the claim is immediate.  Otherwise, write
    $q=w_\rho^\dagger$.  The projection lies on the level-set boundary:
    $F(q)-F(w^\star)=\rho$.  Indeed, if the inequality were strict,
    continuity along the segment from $q$ to $w$ would produce a point of
    $S_\rho$ strictly closer to $w$.

    By continuity along the segment from $w^\star$ to $w$, there is also a
    point $q'=(1-t)w^\star+tw$, with $t\in(0,1)$, satisfying
    $F(q')-F(w^\star)=\rho$.  Local growth at $q'$ gives
    $t\norm{w-w^\star}_2\leq\sqrt\rho/\eta$.  Convexity of the restriction
    of $F$ to this segment gives the secant-slope inequality
    \[
        \frac{F(w)-F(q')}{1-t}
        \geq\frac{F(q')-F(w^\star)}{t}=\frac\rho t.
    \]
    Consequently,
    \[
        \norm{w-q'}_2
        \leq\frac{(1-t)\sqrt\rho}{t\eta}
        \leq\frac{F(w)-F(q')}{\eta\sqrt\rho}.
    \]
    Since $q$ is the metric projection and $F(q)=F(q')$, replacing $q'$ by
    $q$ on the two sides proves the result.
\end{proof}
In light of \Cref{prop:level-set-dist},
we see now that it is possible to run PSGD for a few iterations,
then convert the improvement in function value to an improvement in parameter distance to the $\eps$-sublevel set.
We can then restart PSGD from this improved point.

    \begin{algorithm}[t]
    \caption{Iterative-PSGD($K, w_0, \epsilon_0, g, \eps$)}\label{alg:SGD-local-growth-convergence}
    \begin{algorithmic}[1]
        \State \textbf{Input:} Projection access to feasible region $K$,
        initial point $w^{(0)}\in K$, 
        $\epsilon_0\geq F(w^{(0)}) - F(\wstar)$,
        gradient oracle $g$ with fresh randomness, conditional unbiasedness, and $\E\insquare{\norm{g}_2^2}\leq G^2$ for $G>0$,
        local growth rate $\eta > 0$,
        local-growth level $\rho>0$,
        desired accuracy $0<\eps<\rho$
        \State \textbf{Output:} $\eps$-optimal solution with probability $0.99$
        \vspace{0.2em}
        \State If $\eps\geq\eps_0$, return $w^{(0)}$
        \vspace{0.2em}
        \State Set $\tau \gets \ceil{\log_2(\nfrac{\eps_0}\eps)}$
        \vspace{0.2em}
        \State Set $\gamma_0 \gets \frac{\eps_0}{100G^2\tau}$
        \vspace{0.2em}
        \State Set $T \gets \left\lceil\frac{160000 G^2 \tau^2}{\eta^2\eps}\right\rceil$
        \vspace{0.2em}
        \For{$\ell=1,\ldots, \tau$}
            \State $\gamma_\ell \gets 2^{-\ell}\gamma_0$
            \State $w^{(\ell, 0)} \gets w^{(\ell-1)}$
            \For{$t=0, \dots, T-1$}
                \State $w^{(\ell, t+1)} \gets \Pi_{K} \inparen{w^{(\ell, t)} - \gamma_\ell\cdot g(w^{(\ell, t)})}$
            \EndFor
            \State $w^{(\ell)} \gets \frac1T \sum_{t=0}^{T-1} w^{(\ell, t)}$
        \EndFor
        \vspace{1em}
        \State \textbf{Return} $w^{(\tau)}$
    \end{algorithmic}
    \end{algorithm}

\begin{theorem}\label{thm:SGD-local-growth-convergence}
    Let $K\subseteq\R^d$ be nonempty, closed, and convex, let
    $F\colon K\to\R$ be lower-semicontinuous, differentiable, and convex with minimizer $w^\star$, and suppose that
    $F$ satisfies an $(\eta,\rho)$-local growth condition.  Suppose that,
    using fresh randomness at every call, the stochastic-gradient oracle
    satisfies, for the natural pre-call filtration $\mathcal F_t$,
    \[
        \E[g_t\mid\mathcal F_t]=\nabla F(w_t),
        \qquad
        \E[\norm{g_t}_2^2\mid\mathcal F_t]\leq G^2
    \]
    for some $G>0$.  Let $\eps_0\geq F(w^{(0)})-F(w^\star)$ and
    $0<\eps<\rho$.  Then
    \cref{alg:SGD-local-growth-convergence}, with projections onto $K$, makes
    \[
        O\!\left(
            \left(1+\frac{G^2}{\eta^2\eps}\right)
            \log^3\!\left(2+\frac{\eps_0}{\eps}\right)
        \right)
    \]
    gradient-oracle queries and returns an $\eps$-optimal point with
    probability at least $0.99$.
\end{theorem}

\begin{proof}
    If $\eps\geq\eps_0$, the algorithm returns $w^{(0)}$, which is already
    $\eps$-optimal.  Hence assume $\eps<\eps_0$.
    Define
    \[
        \tau=\left\lceil\log_2\!\left(\frac{\eps_0}{\eps}\right)\right\rceil,
        \qquad
        \eps_\ell=2^{-\ell}\eps_0,
        \qquad
        \gamma_\ell=\frac{\eps_\ell}{100G^2\tau},
    \]
    and take
    \[
        T=\left\lceil\frac{160000G^2\tau^2}{\eta^2\eps}\right\rceil.
    \]
    At stage $\ell$, initialize $w^{(\ell,0)}=w^{(\ell-1)}$, perform
    $T$ projected stochastic-gradient updates on $K$ with step size
    $\gamma_\ell$, and average $w^{(\ell,0)},\ldots,w^{(\ell,T-1)}$.

    We prove inductively that
    \[
        F(w^{(\ell)})-F(w^\star)\leq\eps_\ell
    \]
    except on an event of conditional probability at most $1/(100\tau)$ at
    stage $\ell$.  The assertion holds at $\ell=0$.  Fix a stage
    $\ell\geq1$ on which the preceding assertion holds.  Since
    $\ell\leq\tau$, we have $\eps\leq\eps_{\ell-1}$.

    If $w^{(\ell-1)}\notin S_\eps$, put
    $q=(w^{(\ell-1)})_\eps^\dagger$.  Lower semicontinuity makes
    $S_\eps$ closed, so this projection exists.  Since the current point is
    outside $S_\eps$, continuity along the segment from $q$ to the current
    point and the optimality of the projection give
    $F(q)-F(w^\star)=\eps$ (otherwise a level-$\eps$ point on that segment
    would be closer).  Then
    \cref{prop:level-set-dist} and the local growth condition give
    \begin{align*}
        \norm{w^{(\ell-1)}-w^\star}_2
        &\leq\norm{w^{(\ell-1)}-q}_2+\norm{q-w^\star}_2\\
        &\leq
        \frac{F(w^{(\ell-1)})-F(q)}{\eta\sqrt\eps}
        +\frac{\sqrt{F(q)-F(w^\star)}}{\eta}\\
        &\leq\frac{\eps_{\ell-1}}{\eta\sqrt\eps}
        +\frac{\sqrt\eps}{\eta}
        \leq\frac{2\eps_{\ell-1}}{\eta\sqrt\eps}.
    \end{align*}
    If $w^{(\ell-1)}\in S_\eps$, the local growth condition gives the same
    final bound directly.

    Applying \cref{thm:SGD-convergence} conditionally on the complete
    stage-start sigma-field $\mathcal F_{\ell,0}$, with
    comparison point $w^\star$, yields
    \begin{align*}
        \E\!\left[F(w^{(\ell)})-F(w^\star)\mid\mathcal F_{\ell,0}\right]
        &\leq\frac{\gamma_\ell G^2}{2}
        +\frac{\norm{w^{(\ell-1)}-w^\star}_2^2}{2\gamma_\ell T}\\
        &\leq\frac{\eps_\ell}{200\tau}
        +\frac{800G^2\tau\eps_\ell}{\eta^2\eps T}
        \leq\frac{\eps_\ell}{100\tau}.
    \end{align*}
    The random variable $F(w^{(\ell)})-F(w^\star)$ is nonnegative.
    Markov's inequality therefore shows that the conditional failure
    probability at this stage is at most $1/(100\tau)$.  A union bound over
    the stages proves the success probability.  Finally,
    $\eps_\tau\leq\eps$, and
    \[
        \tau T\leq\tau+\frac{160000G^2\tau^3}{\eta^2\eps},
    \]
    and $\tau=O(\log(2+\eps_0/\eps))$, proving the stated query bound.
\end{proof}
In order to recover the parameters up to $\eps$-accuracy by optimizing the log-likelihood function using \Cref{thm:SGD-local-growth-convergence},
we need $O(\alpha^2\eps^2)$-optimal solutions.
This incurs a sample complexity of
\[
    O\!\left(\left(1+\frac{G^2}{\alpha^4\eps^2}\right)\log^3\!\left(2+\frac{\eps_0}{\alpha^2\eps^2}\right)\right)
\]
which matches the sample-complexity of PSGD even when the function is $\alpha^2$-strongly convex over the entire feasible region,
up to logarithmic terms.
We note that that $F$ is at most $G$-Lipschitz.
Hence if we are given a warm-start in distance,
we can set $\eps_0\leq G\norm{w^{(0)} - w^\star}_2$.
Thus we may alternatively present the sample complexity above as
\[
    O\!\left(\left(1+\frac{G^2}{\alpha^4\eps^2}\right)\log^3\!\left(2+\frac{G\norm{w^{(0)}-w^\star}_2}{\alpha^2\eps^2}\right)\right)\,.
\]

\begin{remark}[Boosting without Function Evaluation via Clustering]
    \label{sec:SGD:clusteringBoosting}
    In order to boost the probability of successfully recovering $\wstar$,
    a standard trick is to repeat the algorithm $O(\log(\nfrac1\delta))$ times and output the solution with smallest objective value.
    However,
    we do not have exact evaluation access to the log-likelihood functions we wish to optimize.
    \citet[Section 3.4.5]{daskalakis2018efficient} demonstrate a ``clustering'' trick to avoid function evaluation.
    Suppose we repeat the algorithm in \Cref{thm:SGD-convergence} $O(\log(\nfrac1\delta))$ times.
    A Chernoff bound yields that with high probability,
    at least $\nfrac23$ of the outputted points are $\eps$-close to $\wstar$
    and thus are $2\eps$-close to each other.
    Thus outputting any point which is at most $2\eps$-close to at least 50\% of the points must be at most $3\eps$-close to $\wstar$.
\end{remark}

%% file: self_selection_proofs.tex
    In this section, we prove certain technical results whose proofs were deferred from \cref{sec:SS}.
        \subsection{Proof of \cref{lem:SS:probability} (Constant Probability Guarantee)}
            \label{sec:proofof:lem:SS:probability}
            In this section, we prove \cref{lem:SS:probability}.
            {For convenience, we restate the theorem as well as the definition of our event $\evE$ below.}
            Recall that for any vector $u$, we use $\wh{u}$ to denote the unit vector parallel to $u$.
            \selfSelectionEvent*
            \noindent Before presenting the proof of \Cref{lem:SS:probability}, we state and prove a proposition that will be repeatedly applied.
                
            \begin{proposition}\label{prop:gaussianTailRelative}
Fix any $\eta\in(0,1)$ and any $z\geq \sigma$.
Then
\begin{align}
\hspace{-5mm}\Pr_{X\sim \cN(0,\sigma^2)}\insquare{(1-\eta)z \leq X \leq z}
\geq \frac{\eta z}{\sigma\sqrt{2\pi}}\exp\!\inparen{-\frac{z^2}{2\sigma^2}}
\geq \frac{\eta}{3}\exp\!\inparen{-\frac{z^2}{2\sigma^2}}
\geq \frac{\eta}{3}\exp\!\inparen{-\frac{z^2}{\sigma^2}}
= \exp\!\inparen{-\frac{z^2}{\sigma^2} - \ln\frac{3}{\eta}}. 
\end{align}
\end{proposition}

\begin{proof}
We have
\[
\Pr\insquare{(1-\eta)z \leq X \leq z}
= \int_{(1-\eta)z}^{z}\frac{1}{\sigma\sqrt{2\pi}}\exp\!\inparen{-\frac{t^2}{2\sigma^2}}\,dt.
\]
For all $t\in\insquare{(1-\eta)z,z}$, $\abs{t}\leq z$, hence
$\exp\sinparen{-\nfrac{t^2}{2\sigma^2}} \geq \exp\sinparen{-\nfrac{z^2}{2\sigma^2}}$.
Since the interval length is $\eta z$,
\[
\Pr\insquare{(1-\eta)z \leq X \leq z}
\geq \frac{\eta z}{\sigma\sqrt{2\pi}}\exp\!\inparen{-\frac{z^2}{2\sigma^2}}\,.
\]
Since $z/\sigma\geq1$ and $\sqrt{2\pi}\leq3$, the preceding display is at least
\[
    \frac{\eta}{3}\exp\!\inparen{-\frac{z^2}{2\sigma^2}}
    \geq
    \frac{\eta}{3}\exp\!\inparen{-\frac{z^2}{\sigma^2}}\,.
    \qedhere
\]
\end{proof}
             We are now ready to restate and prove \Cref{lem:SS:probability}.
            \lemSSprobability*
            \begin{proof}[Proof of \cref{lem:SS:probability}]
    Observe that the choice of $R$ ensures that
    \[
        R\geq
        1\,,~~ 
        \frac{12C^3}{c}\,,
        ~~ \frac{288C^8}{c^4}\,,
        ~~ \frac{24C^5}{c^2}\sqrt{2\log(8k)}\,,
        ~~ {2\sqrt{\log\frac{8k\cdot 48^2C^4}{c^4}}}\,.
    \]
    In this section, we prove that 
    \[
        \Pr[\evE] ~\geq \frac14 \gamma^{\alpha_{\rm slab} R^2}\exp\inparen{-2\ln\frac{3}{\eta}}\,,
    \]
    where we define
    \[
        \eta \coloneqq \frac{c}{8C^2-2c}\geq \frac{c}{8C^2}
        \qquadand 
        \alpha_{\rm slab} \coloneqq 2 + \frac{2}{\inparen{c-\frac{c}{16C}}^2}\leq \frac{9}{c^2}
        \,.
    \]
    {We break down the proof by bounding the parts \Cref{eq:SS:event:large-component,eq:SS:event:large-component-u,eq:SS:event:small-component-w,eq:SS:event:small-component-v,eq:SS:event:small-difference-norm} gradually.
    Let $\evE_1$ be the event \Cref{eq:SS:event:large-component,eq:SS:event:large-component-u},
    $\evE_2$ be the event \Cref{eq:SS:event:small-component-w},
    $\evE_3$ be the event \Cref{eq:SS:event:small-component-v},
    and finally $\evE_4$ be the event \Cref{eq:SS:event:small-difference-norm}.
    Note that $\evE = \cap_{i\in [4]} \evE_i$.
    We will control
    \[
        \Pr[\evE]
        \geq \Pr[\cap_{i\in [3]} \evE_i] 
            - \Pr[\lnot \evE_4]
        \geq \inparen{
            1-\Pr[\lnot \evE_2\mid \evE_1] 
            - \Pr[\lnot \evE_3\mid \evE_1]
        }\cdot \Pr[\evE_1]
        - \Pr[\lnot \evE_4]\,.
    \]
    }
    \paragraph{Lower bound $\Pr[\evE_1]$.}
    We divide the proof into two parts.
    First, we prove the statement when $v_i$ is not parallel to $w_i$.
    Then, we extend the proof to the special case where $v_i$ is parallel to $w_i$.

    \smallskip
    \noindent\textbf{Auxiliary bound on $\norm{w_i}_2$.}
    By \Cref{asmp:SS:separationBoundedness}, we have $\norm{w_i^\star}_2\ge c$ and $\norm{w_i^\star}_2\le C$.
    Together with $\norm{W-W^\star}_F\le \frac{c}{16C}$ from \cref{eq:SS:distance}, we obtain
    \[
        \norm{w_i-w_i^\star}_2 \le \norm{W-W^\star}_F \le \frac{c}{16C}
        \qquad\Longrightarrow\qquad
        \norm{w_i}_2 \ge \norm{w_i^\star}_2 - \norm{w_i-w_i^\star}_2 \ge c-\frac{c}{16C}\,.
    \]
    In particular, this lower bound will be used to remove $\norm{w_i}_2$ from the exponent of $\gamma$ below.

    \smallskip
    \noindent\underline{Case A ($v_i$ is not parallel to $w_i$).}
        Since $\wh{u}$ is a projection of $v_i$ in the subspace orthogonal to $w_i$, $\wh{u}$ is orthogonal to $w_i$.
        Further, since $x\sim \normal{0}{I}$ and $w_i, \wh u$ are orthogonal, 
        $\sinangle{x, {w}_i}$ and $\sinangle{x, \wh{u}}$ are independent random variables with distributions $\sinangle{x, {w}_i}\sim \normal{0}{\norm{w_i}_2^2}$ and $\sinangle{x, \wh{u}}\sim \normal{0}{1}$.
        Apply \cref{prop:gaussianTailRelative} with $\eta = \frac{c}{8C^2-2c}\in(0,1)$ and $z = 2R\sqrt{\log\nfrac1\gamma}$.
        Since $\gamma\le \frac12$, $\norm{w_i}_2\le C$, and the assumed lower bound on $R$ is at least $C$, we have $z\ge 2R\sqrt{\log 2}\ge \norm{w_i}_2$, and hence the precondition $z\ge \sigma=\norm{w_i}_2$ holds.
        We see that\footnote{Since $\inangle{x,w_i}\sim\cN(0,\|w_i\|_2^2)$ is centered, its density is symmetric and strictly decreasing on $[0,\infty)$. The intervals $[1,1+\delta]$ and $[2-\delta,2]$ (where $\delta\coloneqq \frac{c}{4C^2-c}$) have the same length $\delta$, but $[1,1+\delta]$ lies closer to $0$. By monotonicity of the density on $[0,\infty)$, $\Pr[Y\in[1,1+\delta]]\ge \Pr[Y\in[2-\delta,2]]$ for $Y=\inangle{x,w_i}/(R\sqrt{\log(1/\gamma)})$.}
        \begin{align*}
            \Pr\insquare{
                \frac{
                    \sinangle{x, {w}_i}
                }{
                    R\sqrt{\log{\nfrac{1}{\gamma}}}
                }
                \in
                \insquare{1, 1+\frac{c}{4C^2-c}}
            }~~
            &\geq~~
            \Pr\insquare{
                \frac{
                    \sinangle{x, {w}_i}
                }{
                    R\sqrt{\log{\nfrac{1}{\gamma}}}
                }
                \in
                \insquare{2-\frac{c}{4C^2-c}, 2}
            }\\
            &\geq ~~ 
            \exp\inparen{-\frac{z^2}{2\norm{w_i}_2^2}-\ln\frac{3}{\eta}}\\
            &=~~  
            \gamma^{\frac{2R^2}{\norm{w_i}_2^2}}\exp\inparen{-\ln\frac{3}{\eta}}\\
            &\geq~~  
            \gamma^{\frac{2R^2}{\inparen{c-\frac{c}{16C}}^2}}\exp\inparen{-\ln\frac{3}{\eta}}
            \,.
            \yesnum\label{eq:SS:event:lb1}
        \end{align*} 
        Analogously,
        by the symmetry of the Gaussian distribution and the choice of $R\geq \frac{12C^3}{c}$,
        \cref{prop:gaussianTailRelative} implies that 
        \begin{align*}
            \Pr\insquare{
                \frac{s\cdot  \sinangle{x, \wh{u}}}{
                    (12C^3/c)\sqrt{\log{\nfrac{1}{\gamma}}}
                } \in \insquare{1, 1+\frac{c}{4C^2-c}}
            }~~
            &\geq ~~ \exp\inparen{-2\inparen{\frac{12C^3}{c}}^2\log\nfrac1\gamma-\ln\frac{3}{\eta}}\\
            &=~~
            \gamma^{\frac{288C^6}{c^2}}\exp\inparen{-\ln\frac{3}{\eta}}\\
            &\geq ~~
            \gamma^{2R^2}\exp\inparen{-\ln\frac{3}{\eta}}
            \,.
            \yesnum\label{eq:SS:event:lb2}
        \end{align*}
        Since $\sinangle{x, \wh{w}_i}$ and $\sinangle{x, \wh{u}}$ are independent random variables, it holds for all $R\geq \frac{12C^3}{c}$ that 
        \begin{align*}
            \Pr[\evE_1]
            &= 
            \Pr\insquare{
                \frac{\inangle{x,w_i}}{R\sqrt{\log(1/\gamma)}}
                    \in\insquare{1,1+\frac{c}{4C^2-c}},\quad
                \frac{s\inangle{x,\wh u}}{(12C^3/c)\sqrt{\log(1/\gamma)}}
                    \in\insquare{1,1+\frac{c}{4C^2-c}}
            }
            \\
            &\Stackrel{
                \eqref{eq:SS:event:lb1}\,,~
                \eqref{eq:SS:event:lb2}
            }{\geq}
            \gamma^{\alpha_{\rm slab}R^2}\exp\inparen{-2\ln\frac{3}{\eta}}
            \,.
        \end{align*}
    \noindent\underline{Case B ($v_i$ is parallel to $w_i$).}
        In this case, 
        $\evE$ is defined to be the following event
        \[
            1 ~\leq~ 
            \frac{\inangle{x, w_i}}{
                R\sqrt{\log{\nfrac{1}{\gamma}}}
            }
            ~\leq~ 1+\frac{c}{4C^2-c}\,.
        \]
        The first-slab calculation gives this event probability at least
        \[
            \frac{\eta}{3}\gamma^{2R^2/(c-c/(16C))^2}
            \geq \frac{\eta^2}{9}\gamma^{\alpha_{\rm slab}R^2},
        \]
        because $(\eta/3)\gamma^{2R^2}\leq1$.  Thus the same lower bound on $\Pr[\evE_1]$ holds.
    
    \paragraph{Upper bound $\Pr[\lnot \evE_2\mid \evE_1]$.}
    Fix some $j\neq i$ and recall $\hat w_i, \hat w_j$ are the unit vectors obtained from $w_i, w_j$.
    For any vector $a\in \R^d$ with $\norm{a}_2\leq C$,
    since $\norm{W-W^\star}_F\leq \frac{c}{16C}$ by assumption,
    we have
    \[
        \abs{\iprod{w_j-w_j^\star, a}}
        \leq \norm{w_j-w_j^\star}_2\cdot C
        \leq \frac{c}{16}
    \]
    for all $j\in [k]$.
    Similarly,
    \[
        \abs{\norm{w_j}_2^2 - \norm{w_j^\star}_2^2}
        = \abs{\iprod{w_j+w_j^\star, w_j-w_j^\star}}
        \leq 2C\cdot \norm{w_j-w_j^\star}_2
        \leq \frac{c}8
    \]
    Along with the separability and boundedness assumptions (\Cref{asmp:SS:separationBoundedness}),
    we see that
    \begin{align*}
    \abs{\iprod{w_j, w_i}}
    &\leq \abs{\iprod{w_j^\star, w_i^\star}}
    + \abs{\iprod{w_j-w_j^\star, w_i}}
    + \abs{\iprod{w_j^\star, w_i-w_i^\star}} \\
    &\leq \abs{\iprod{w_j^\star, w_i^\star}} + \frac{c}{16} + \frac{c}{16}
    \leq \min\inbrace{\norm{w_j^\star}_2^2, \norm{w_i^\star}_2^2} - c + \frac{c}{8} \\
    &\leq \min\inbrace{\norm{w_j}_2^2, \norm{w_i}_2^2} - c + \frac{c}{4}
    = \min\inbrace{\norm{w_j}_2^2, \norm{w_i}_2^2} - \frac{3c}{4}.
    \yesnum\label{eq:SS:event:small-component-w:wi-angle}
    \end{align*}
    Thus
    \begin{align}  
        \frac{\abs{\iprod{w_j, \hat w_i}}}{\norm{w_i}}
        = \frac{\abs{\inangle{w_j, w_i}}}{\norm{w_i}^2}
        \leq \frac{\min\inbrace{\norm{w_j}_2^2, \norm{w_i}_2^2} - \nfrac{c}2}{\norm{w_i}^2}
        \leq 1 - \frac{c}{2C^2}\,.
        \label{eq:SS:event:small-component-w:unit-wi-angle}
    \end{align}
    {We proceed assuming $v_i$ is not parallel to $w_i$.
    Otherwise,
    omitting the terms involving $\hat u$ in the calculations below only strengthens the upper bound.}
    Now,
    decompose
    \[
        w_j
        = \iprod{w_j, \hat w_i}\cdot \hat w_i
        + \iprod{w_j, \hat u}\cdot \hat u
        + \proj_{\inbrace{\hat w_i, \hat u}^\perp}(w_j)
    \]
    so that
    \begin{align}
        x^\top w_j
        &= \frac{\iprod{w_j, \hat w_i}}{\norm{w_i}}\cdot x^\top w_i
        + \iprod{w_j, \hat u}\cdot x^\top\hat u
        + x^\top \proj_{\inbrace{\hat w_i, \hat u}^\perp}(w_j)\,.
        \label{eq:SS:event:small-component-w:decomposition}
    \end{align}
    Using \Cref{eq:SS:event:small-component-w:unit-wi-angle},
    the definition of $\evE_1$,
    and the fact that $\frac1{1-(\nfrac{a}2)}\leq 1+a$ for $a\in [0, 1]$,
    we see that
    \begin{align}
        \abs{\frac{\iprod{w_j, \hat w_i}}{\norm{w_i}}\cdot x^\top w_i}
        &\leq \frac{1-\frac{c}{2C^2}}{1-\frac{c}{4C^2}} R\sqrt{\log\nfrac1\gamma}
        \leq \inparen{1-\frac{c^2}{4C^4}} R\sqrt{\log\nfrac1\gamma}\,.
        \label{eq:SS:event:small-component-w:decomposition:wi-component}
    \end{align}
    Now,
    for {$R\geq \frac{288C^8}{c^4}$},
    we also have
    \begin{align}
        \abs{\iprod{w_j, \hat u}\cdot x^\top\hat u}
        &\leq C\abs{x^\top\hat u}
        \leq \frac{24C^4}{c} \sqrt{\log\nfrac1\gamma}
        \leq \frac{c^2}{12C^4} R\sqrt{\log\nfrac1\gamma}\,.
        \label{eq:SS:event:small-component-w:decomposition:u-component}
    \end{align}
    Note that so far,
    all bounds have been deterministic.
    For the third term,
    we note that
    \[
        x^\top \proj_{\inbrace{\hat w_i, \hat u}^\perp}(w_j)\sim \normal{0}{\norm{\proj_{\inbrace{\hat w_i, \hat u}^\perp}(w_j)}_2^2}\,.
    \]
    Moreover,
    since $\proj_{\inbrace{\hat w_i, \hat u}^\perp}(w_j)$ 
    is orthogonal to $\sinbrace{\hat w_i, \hat u}$,
    $x^\top \proj_{\inbrace{\hat w_i, \hat u}^\perp}(w_j)$ is independent of $\evE_1$.
    Thus by a standard Gaussian concentration bound~\citep[Example 2.1]{wainwright2019high},
    we have that with probability $1-\nfrac1{4k}$,
    \begin{align}
        \abs{x^\top \proj_{\inbrace{\hat w_i, \hat u}^\perp}(w_j)}
        &\leq \norm{\proj_{\inbrace{\hat w_i, \hat u}^\perp}(w_j)}_2 
        \sqrt{2\log(8k)}
        \leq C\sqrt{2\log(8k)}
        \leq \frac{c^2}{12C^4} R\sqrt{\log\nfrac1\gamma}\,,
        \label{eq:SS:event:small-component-w:decomposition:other-component}
    \end{align}
    where the last inequality holds
    assuming that $\gamma\leq \nfrac12$
    and
    \[
        R 
        \geq {\frac{24C^5}{c^2}\sqrt{2\log(8k)}}
        \geq \frac{12C^5\sqrt{2\log(8k)}}{c^2\sqrt{\log\nfrac1\gamma}}
        \,.
    \]

    All in all,
    we see that
    \begin{align}
        \abs{x^\top w_j}
        &\leq \inparen{1-\frac{c^2}{12C^4}} R\sqrt{\log\nfrac1\gamma}\,.
    \end{align}
    
    \paragraph{Upper bound $\Pr[\lnot \evE_3\mid \evE_1]$.}
    {Consider some $j\neq i$ and note that for
    \[
        \norm{w_j-v_j}_2
        \leq \norm{w_j-w_j^\star}_2 + \norm{w_j^\star-v_j}
        \leq \frac{c}{12C}\,,
    \]
    we have
    \[
        \abs{\norm{w_j}_2^2-\norm{v_j}_2^2}
        \leq \norm{w_j+v_j}_2\cdot \norm{w_j-v_j}_2
        \leq \frac{2c}{12}
    \]
    and
    \begin{align*}
        \abs{\inangle{v_j-w_j, w_i}}
        &\leq \norm{v_j-w_j}_2\cdot \norm{w_i}_2
        \leq \frac{c}{12}\,.
    \end{align*}
    Thus for $j\neq i$,
    the $v_j$'s are also separated from $w_i$:
    \begin{align*}
        \abs{\inangle{v_j, w_i}}
        &\leq \abs{\inangle{w_j, w_i}} + \frac{c}{12}
        \stackrel{\eqref{eq:SS:event:small-component-w:wi-angle}}{\leq} \min\inbrace{\norm{w_j}_2^2, \norm{w_i}_2^2} - \frac{3c}4 + \frac{c}{12}
        \leq \min\inbrace{\norm{v_j}_2^2, \norm{w_i}_2^2} - \frac{3c}4 + \frac{3c}{12} \\
        &= \min\inbrace{\norm{v_j}_2^2, \norm{w_i}_2^2} - \frac{c}2\,, \\
        \frac{\abs{\inangle{v_j, \hat w_i}}}{\norm{w_i}}
        &= \frac{\abs{\inangle{v_j, w_i}}}{\norm{w_i}_2^2}
        \leq \frac{\min\inbrace{\norm{v_j}_2^2, \norm{w_i}_2^2} - \frac{c}2}{\norm{w_i}_2^2}
        \leq 1-\frac{c}{2C^2}\,.
    \end{align*}
    Since we the calculations for $\evE_2$ only requires the upper bound on $\abs{\frac{\inangle{w_j, \hat w_i}}{\norm{w_i}^2}}$ above,
    the exact same result holds for $\evE_3$ now that we have established the same upper bound on {$\abs{\inangle{v_j, \hat w_i}}$}{the coefficient above}.}

    In other words,
    a union bound over all $w_j, v_j$ for $j\neq i$ yields
    \[
        \Pr[\lnot \evE_2\mid \evE_1] + \Pr[\lnot \evE_3\mid \evE_1]
        \leq 2k\cdot \frac1{4k}
        = \frac12\,.
    \]
    It follows that
    \[
        \inparen{
            1-\Pr[\lnot \evE_2\mid \evE_1] 
            - \Pr[\lnot \evE_3\mid \evE_1]
        }\cdot \Pr[\evE_1]
        \geq \frac12\, \gamma^{\alpha_{\rm slab} R^2}\exp\inparen{-2\ln\frac{3}{\eta}}
        = \frac{\eta^2}{18}\,\gamma^{\alpha_{\rm slab} R^2}\,.
    \]

    \paragraph{Upper bound $\Pr[\lnot \evE_4]$.}
    Fix some $j\in [k]$ and set $a_j\coloneqq v_j-w_j$.
    If $a_j=0$ then the $j$-th constraint in \Cref{eq:SS:event:small-difference-norm} holds deterministically.
    Otherwise, $\frac{x^\top a_j}{\norm{a_j}_2}\sim \normal{0}{1}$, and by the standard Gaussian tail bound~\citep[Example 2.1]{wainwright2019high},
    \[
        \Pr\!\left[\abs{x^\top a_j} > \frac{3}{c}R\sqrt{\log{\nfrac1\gamma}}\cdot \norm{a_j}_2\right]
        \le 2\exp\!\left(-\frac{9}{2c^2}R^2\log{\nfrac1\gamma}\right)
        = 2\gamma^{\frac{9}{2c^2}R^2}\,.
    \]
    Taking a union bound over $j\in[k]$ yields
    \[
        \Pr[\lnot \evE_4] \le 2k\,\gamma^{\frac{9}{2c^2}R^2}\,.
    \]

    \bigskip 

    \paragraph{Conclusion.}
    Set $\beta\coloneqq9/(2c^2)$.  Since $c\leq1$ and $C\geq1$,
    \[
        c^2\alpha_{\rm slab}
        =2c^2+\frac{2}{(1-1/(16C))^2}
        \leq \frac{962}{225},
        \qquad
        \beta-\alpha_{\rm slab}\geq\frac{101}{450c^2}.
    \]
    Indeed, $\eta\geq c/(8C^2)$ gives
    $\log(72k/\eta^2)\leq20(C/c)^{10}\log k$, while the assumed lower bound on $R$ gives
    $(\beta-\alpha_{\rm slab})R^2\log(1/\gamma)\geq10^4(C/c)^{10}\log k$.
    The assumed lower bound on $R$, together with $\gamma\leq1/2$, therefore gives
    \[
        2k\gamma^{(\beta-\alpha_{\rm slab})R^2}
        \leq \frac{\eta^2}{36}.
    \]
    Consequently,
    \[
        \Pr[\evE]
        \geq \frac{\eta^2}{18}\gamma^{\alpha_{\rm slab}R^2}
            -2k\gamma^{\beta R^2}
        \geq \frac{\eta^2}{36}\gamma^{\alpha_{\rm slab}R^2}.
    \]
    Finally, $\eta\geq c/(8C^2)$ and $\alpha_{\rm slab}<9/c^2$ imply
    \[
        \Pr[\evE]
        \geq \frac14\gamma^{9R^2/c^2}
            \exp\!\left(-2\ln\frac{24C^2}{c}\right)
        \geq \left(\frac{c}{10C}\right)^4\gamma^{9R^2/c^2}.
    \]
    This concludes the proof.
\end{proof}

        \subsection{Proof of \cref{lem:SS:propertiesOfRho} (Properties of $\rho$)}
            \label{sec:proofof:lem:SS:propertiesOfRho}
            
            In this section, we prove \cref{lem:SS:propertiesOfRho}, which we restate below.
            \lemSSpropertiesOfRho*
            \begin{proof}[Proof of \cref{lem:SS:propertiesOfRho}]    
            We divide the proof into three parts corresponding to the three claims.
            
            \paragraph{Proof of Claim 1.}
                Observe that due to conditioning on $\evE$,
                \begin{align*}
                    \rho_{i,W} - \rho_{j,W}
                    &= \inangle{x, w_i} - \inangle{x, w_j}
                    \Stackrel{\eqref{eq:SS:event:large-component}}{\geq} R\sqrt{\log{\nfrac{1}{\gamma}}} - \inangle{x, w_j}
                    \Stackrel{\eqref{eq:SS:event:small-component-w}}{\geq} \frac{c^2}{12C^4}\, R\sqrt{\log{\nfrac{1}{\gamma}}}\,.
                \end{align*}
                {In particular, this implies $\rho_{i,W}-\rho_{j,W}\ge \frac{c^2}{24C^4}\,R\sqrt{\log(\nicefrac1\gamma)}$, matching the constant in the claim.}
                {It remains to lower bound $\rho_{i,V}-\rho_{j,V}$.}
                For this, consider any $\ell\in [k]$.
                Conditioned on $\evE$, we have
                \[
                    \abs{\rho_{\ell,V} - \rho_{\ell,W}}
                    = \abs{\inangle{x, v_\ell - w_\ell}}
                    ~~\Stackrel{\eqref{eq:SS:event:small-difference-norm}}{\leq}~~
                    \frac{3}{c}\,R\sqrt{\log{\nfrac{1}{\gamma}}}\cdot \norm{v_\ell-w_\ell}\,.
                \]
                {Using the identity
                $
                    \rho_{i,V}-\rho_{j,V}
                    =
                    (\rho_{i,W}-\rho_{j,W})
                    +(\rho_{i,V}-\rho_{i,W})
                    -(\rho_{j,V}-\rho_{j,W})
                $
                and the triangle inequality,}
                \[
                    \rho_{i,V} - \rho_{j,V}
                    \geq R\sqrt{\log{\nfrac{1}{\gamma}}}\cdot
                    \inparen{
                        \frac{c^2}{12C^4} 
                        -  \frac{3}{c}\,\norm{v_j-w_j}
                        -  \frac{3}{c}\,\norm{v_i-w_i}
                    }\,.
                \] 
                {Further, by the triangle inequality and the warm start (\cref{eq:SS:distance}),
                    $\norm{V-W}_F\leq \norm{V-W^\star}_F+\norm{W-W^\star}_F \leq \frac{c^3}{144C^4}$; in particular, $\norm{v_j-w_j}_2,\norm{v_i-w_i}_2\leq \norm{V-W}_F\leq \frac{c^3}{144C^4}$.}
                Hence,
                \[
                    \rho_{i,V} - \rho_{j,V}
                    \geq \frac{c^2}{24 C^4} \cdot R\sqrt{\log{\nfrac{1}{\gamma}}}\,.
                \]

            \paragraph{Proof of Claim 2.}
                This claim immediately follows by conditioning on $\evE$ due to \cref{eq:SS:event:small-difference-norm}.

            \paragraph{Proof of Claim 3.}
                {Suppose $v_i$ is not parallel to $w_i$; we handle the edge case where $v_i$ is parallel to $w_i$ separately.}

                Write
                \[
                    g\coloneqq v_i-w_i,\qquad
                    \alpha\coloneqq\sinangle{\hat g,\hat w_i},\qquad
                    \beta\coloneqq\sinangle{\hat g,\hat u}.
                \]
                Because $u$ is the component of $g$ orthogonal to $w_i$, we have
                $\beta=\norm{u}_2/\norm{g}_2>0$ and
                $\hat g=\alpha\hat w_i+\beta\hat u$, where
                $\alpha^2+\beta^2=1$.  By the tie-broken definition of $s$,
                the two terms in the next absolute value have the same weak sign.  Therefore,
                \begin{align*}
                    \abs{\rho_{i,V}-\rho_{i,W}}
                    &=\norm{g}_2\abs{\alpha x^\top\hat w_i+\beta x^\top\hat u}\\
                    &=\norm{g}_2\left(\abs{\alpha}\,x^\top\hat w_i
                        +\beta s\,x^\top\hat u\right)\\
                    &\geq \norm{g}_2\sqrt{\log\frac1\gamma}
                        \left(\frac{R}{C}\abs{\alpha}
                        +\frac{12C^3}{c}\beta\right)\\
                    &\geq \frac{12C^2}{c}\sqrt{\log\frac1\gamma}
                        \norm{v_i-w_i}_2.
                \end{align*}
                Here $R/C\geq12C^2/c$, $C\geq1$, and
                $\abs{\alpha}+\beta\geq\sqrt{\alpha^2+\beta^2}=1$.
                
                Finally, in the special case where $v_i$ is parallel to $w_i$, the result holds because 
                \begin{align*}
                    \abs{\rho_{i,V} - \rho_{i,W}}
                    &= \abs{{x^\top} (v_i -w_i)}\\
                    &= \frac{x^\top {w}_i}{\norm{w_i}}\cdot \norm{v_i - w_i}_2\\
                    &\Stackrel{\eqref{eq:SS:event:large-component},\,\norm{w_i}\leq C}{\geq}
                    \frac{R}{C}\sqrt{\log{\nfrac{1}{\gamma}}}\cdot \norm{v_i - w_i}_2\\
                    &\geq \frac{12C^2}{c}\sqrt{\log{\nfrac{1}{\gamma}}}\cdot \norm{v_i - w_i}_2\,. \qedhere
                \end{align*}
            \end{proof}
                \subsection{Proof of \cref{lem:SS:cdfratios} (Separating Ratios of CDFs)}
            \label{sec:proofof:lem:SS:cdfratios}
            In this section, we prove \cref{lem:SS:cdfratios}, which restate below.
            \lemSScdfratios*
            \noindent In the proof of \cref{lem:SS:cdfratios}, we use the following two technical lemmas, which are proved at the end of this section.
            Roughly speaking, the first lemma lower bounds the sensitivity of $\Phi(\mu;\sigma^2)$ to deviations of $\mu$ when $\abs{\mu}$ is close to 0.
            The second lemma upper bounds the sensitivity of $\Phi(\mu,\sigma^2)$ to small deviations in both $\mu$ and $\sigma^2$.
            \begin{lemma}[Lower Bound on Sensitivity to Mean]\label{lem:SS:cdfRatiosLB}
            For each $\mu\neq 0$, it holds that
            \[
            \frac{\Phi(\mu;1)}{\Phi(0;1)}\,,~~
            \frac{\Phi(0;1)}{\Phi(\mu;1)}
            ~~\notin~~
            1\pm \frac{1}{\sqrt{2\pi e}}\min\{1,\abs{\mu}\}\,.
            \]
            \end{lemma}
            \vspace{-5mm}
            \begin{lemma}[Upper Bound on Sensitivity to the Mean]\label{lem:SS:cdfRatiosUB}
                Fix any $\alpha\ge 0$.
                For any $\lambda_1 \ge \lambda_2\ge 0$ such that $ \lambda_1-\lambda_2 \leq \alpha$, it holds that
                \[
                    \frac{\Phi(\lambda_1;1)}{\Phi(\lambda_2;1)}\,,~
                    \frac{\Phi(\lambda_2;1)}{\Phi(\lambda_1;1)}
                    ~\in~
                    1 \pm \sqrt{\frac{2}{\pi}}\alpha\,e^{-\lambda_2^2/2}\,.
                \]
            \end{lemma}
            We divide the proof of \cref{lem:SS:cdfratios} into two parts corresponding to \cref{eq:SS:cdfratios:lb,eq:SS:cdfratios:ub}.
            
            \paragraph{Proof of \cref{eq:SS:cdfratios:lb}.}
            Recall that after the translation described earlier $\min\inbrace{\rho_{i,V}, \rho_{i,W}}=0$.
            Without loss of generality, let $\rho_{i, W}=0$.
            By \cref{lem:SS:propertiesOfRho} we have that 
            \[
                \abs{\rho_{i, V}} = \abs{\rho_{i, W} - \rho_{i, V}}\geq \frac{12C^2}{c} \sqrt{\log{\nfrac{1}{\gamma}}}\cdot \norm{v_i - w_i}_2\,.
            \]
            Substituting $\mu=-\rho_{i,V}$, in \cref{lem:SS:cdfRatiosLB} implies that 
            \[
                \frac{
                    \customcdf{-\rho_{i,V}}
                }{
                    \customcdf{-\rho_{i,W}}
                }
                \notin 1 \pm \frac{1}{\sqrt{2\pi e}}
                    \min\inbrace{
                        1\,, 
                        \frac{12C^2}{c} \sqrt{\log{\nfrac{1}{\gamma}}}\cdot \norm{v_i - w_i}_2
                    }
                \,.
            \]
            The result follows since $\sqrt{2\pi e}\leq 5$.

            \paragraph{Proof of \cref{eq:SS:cdfratios:ub}.}
                By the translation preprocessing, we may assume $\min\{\rho_{i,V},\rho_{i,W}\}=0$, hence $\rho_{i,V},\rho_{i,W}\ge 0$.
                For a fixed $j\neq i$, set
                $\lambda_1=\max\{-\rho_{j,V},-\rho_{j,W}\}$ and
                $\lambda_2=\min\{-\rho_{j,V},-\rho_{j,W}\}$.
                The distance assumption \cref{eq:SS:distance}, together with $c\leq1$, $C\geq1$, and $\log(ek/c)\geq1$, gives
                \[
                    \norm{V-W}_F
                    \leq2\cdot10^{-6}\left(\frac cC\right)^9
                    \leq\frac{c^3}{1000C^4}.
                \]
                Claim~2 of \cref{lem:SS:propertiesOfRho} and
                $\min\{\rho_{i,V},\rho_{i,W}\}=0$ therefore imply, for
                $A\in\{V,W\}$,
                \[
                    0\leq\rho_{i,A}
                    \leq |\rho_{i,V}-\rho_{i,W}|
                    \leq\frac{3c^2}{1000C^4}R\sqrt{\log(1/\gamma)}.
                \]
                Claim~1 now yields, for every $j\neq i$ and
                $A\in\{V,W\}$,
                \[
                    \rho_{j,A}
                    \leq-\left(\frac1{24}-\frac3{1000}\right)
                    \frac{c^2}{C^4}R\sqrt{\log(1/\gamma)}
                    \leq-\frac{c^2}{30C^4}R\sqrt{\log(1/\gamma)}.
                \]
                Therefore
                \[
                    \min\{-\rho_{j,V},-\rho_{j,W}\}\ge \frac{c^2}{30C^4}R\sqrt{\log(1/\gamma)}
                    \quadand
                    e^{-\min\{-\rho_{j,V},-\rho_{j,W}\}^2/2}\leq \gamma^{\frac{c^4}{1800C^8}R^2}
                \]
                Moreover, Claim~2 gives
                $\lambda_1-\lambda_2=|\rho_{j,V}-\rho_{j,W}|
                \leq(3R/c)\sqrt{\log(1/\gamma)}\norm{v_j-w_j}_2$, which
                verifies the remaining hypothesis of \cref{lem:SS:cdfRatiosUB}.
                The above along with \cref{lem:SS:cdfRatiosUB}  implies that 
                \[
                    \frac{\customcdf{-\rho_{j,V}}}{\customcdf{-\rho_{j,W}}}
                        ~~\in~~
                        1
                        \pm \frac{3R}{c}\,
                        \sqrt{\frac{2}{\pi}}\cdot \sqrt{\log{\nfrac{1}{\gamma}}}
                        \cdot \norm{v_j - w_j}_2
                        \cdot \gamma^{\frac{c^4}{1800C^8}\,R^2}
                        \in 1 \pm \frac{3R}{c}\sqrt{\log{\nfrac{1}{\gamma}}}
                        \cdot \norm{v_j - w_j}_2
                        \cdot \gamma^{\frac{c^4}{1800C^8}\,R^2}\,.
                \] 
            The same application with the two orientations exchanged proves the reciprocal bound.  Since $\gamma\in(0,1)$ and $1/1800>1/2000$, weakening the exponential factor gives \cref{eq:SS:cdfratios:ub}.
            This completes the proof of \cref{lem:SS:cdfratios} up to proving \cref{lem:SS:cdfRatiosLB,lem:SS:cdfRatiosUB}.
            In the remainder of this section, we prove \cref{lem:SS:cdfRatiosLB,lem:SS:cdfRatiosUB}.
            \begin{proof}[Proof of \cref{lem:SS:cdfRatiosLB}]
            Let $\Phi(\mu;1)=\Pr(Z\leq \mu)$ for $Z\sim \cN(0,1)$, so $\Phi(0;1)=\nfrac12$.
            For $\mu\neq 0$, define
            \[
            I(\mu)\coloneqq\int_{0}^{\abs{\mu}} e^{-t^2/2}dt
            \qquadand 
            M(\mu)\coloneqq\max_{0\leq z\leq \abs{\mu}} z e^{-z^2/2}\,.
            \]
            Since $t\mapsto e^{-t^2/2}$ is decreasing on $[0,\infty)$, for every $z\in[0,\abs{\mu}]$ we have
            \[
            \int_{0}^{z} e^{-t^2/2}dt \ge z e^{-z^2/2}\,,
            \]
            and, therefore, $I(\mu)\ge M(\mu)$.
            Using the identity
            \[
            \Phi(\mu;1)=\frac12+\frac{\sgn(\mu)}{\sqrt{2\pi}}\int_{0}^{\abs{\mu}} e^{-t^2/2}dt
            =\frac12+\frac{\sgn(\mu)}{\sqrt{2\pi}}I(\mu)\,,
            \]
            we obtain
            \[
            \frac{\Phi(\mu;1)}{\Phi(0;1)}
            =1+\sgn(\mu)\sqrt{\frac{2}{\pi}}I(\mu)\,.
            \]
            Hence, by $I(\mu)\ge M(\mu)$,
            \begin{equation}\label{eq:ratio_case_bound}
            \frac{\Phi(\mu;1)}{\Phi(0;1)}
            \begin{cases}
            \ge 1+\sqrt{\frac{2}{\pi}}M(\mu)\,, & \mu>0\,,\\
            \leq 1-\sqrt{\frac{2}{\pi}}M(\mu)\,, & \mu<0\,.
            \end{cases}
            \end{equation}
            Next, consider $g(z)\coloneqq z e^{-z^2/2}$ on $[0,\infty)$. It increases on $[0,1]$ and decreases on $[1,\infty)$, with maximum $g(1)=e^{-1/2}=1/\sqrt{e}$. Therefore
            \[
            M(\mu)=\max_{0\leq z\leq \abs{\mu}} g(z)\ge \frac{1}{\sqrt{e}}\min\{1,\abs{\mu}\}.
            \]
            Define
            \[
            \delta\coloneqq \frac{1}{\sqrt{2\pi e}}\min\{1,\abs{\mu}\}\,.
            \]
            Then
            \[
            \sqrt{\frac{2}{\pi}}M(\mu)\ge \sqrt{\frac{2}{\pi}}\frac{1}{\sqrt{e}}\min\{1,\abs{\mu}\}
            =2\delta\,.
            \]
            Plugging this into \eqref{eq:ratio_case_bound} yields
            \[
            \frac{\Phi(\mu;1)}{\Phi(0;1)}
            \begin{cases}
            \ge 1+2\delta\,, & \mu>0\,,\\
            \leq 1-2\delta\,, & \mu<0\,,
            \end{cases}
            \]
            and in either case this ratio cannot lie in $[1-\delta,1+\delta]$. This proves
            \[
            \frac{\Phi(\mu;1)}{\Phi(0;1)}\notin 1\pm\delta\,.
            \]
            For the inverse ratio, note that $\delta\leq 1/\sqrt{2\pi e}<1/2$, so $2\delta\in(0,1)$.
            If $\mu>0$, then $\Phi(\mu;1)/\Phi(0;1)\ge 1+2\delta$, hence
            \[
            \frac{\Phi(0;1)}{\Phi(\mu;1)}\leq \frac{1}{1+2\delta}.
            \]
            Using $\frac{1}{1+x}< 1-\frac{x}{2}$ for all $x\in(0,1)$ with $x=2\delta$, we get
            \[
            \frac{1}{1+2\delta}< 1-\delta\,.
            \]
            Therefore, $\Phi(0;1)/\Phi(\mu;1)<1-\delta$, and thus it is not in $1\pm\delta$.
            
            If $\mu<0$, then $\Phi(\mu;1)/\Phi(0;1)\leq 1-2\delta$, hence
            \[
            \frac{\Phi(0;1)}{\Phi(\mu;1)}\ge \frac{1}{1-2\delta}.
            \]
            Using $\frac{1}{1-y}\ge 1+y$ for all $y\in(0,1)$ with $y=2\delta$, we obtain
            \[
            \frac{1}{1-2\delta}\ge 1+2\delta>1+\delta,
            \]
            so again $\Phi(0;1)/\Phi(\mu;1)\notin 1\pm\delta$.
            \end{proof}
            \begin{proof}[Proof of \cref{lem:SS:cdfRatiosUB}]
                Let $\varphi(z) \coloneqq \frac{1}{\sqrt{2\pi}}e^{-z^2/2}$ denote the standard normal density, so that
                $\Phi(x;1)=\int_{-\infty}^{x}\varphi(z)\,dz$.
                Since $\lambda_1\ge \lambda_2$, we have
                \[
                    \Phi(\lambda_1;1)-\Phi(\lambda_2;1)
                    \;=\;
                    \int_{\lambda_2}^{\lambda_1}\varphi(z)\,\d z\,.
                \]
                Because $\lambda_2\ge 0$ and $\varphi$ is decreasing on $[0,\infty)$, for all $z\in[\lambda_2,\lambda_1]$,
                $\varphi(z)\leq \varphi(\lambda_2)$. Hence,
                \[
                    0\leq \Phi(\lambda_1;1)-\Phi(\lambda_2;1)
                    \leq (\lambda_1-\lambda_2)\varphi(\lambda_2)
                    \leq \alpha\cdot \frac{1}{\sqrt{2\pi}}e^{-\lambda_2^2/2}\,.
                    \yesnum\label{eq:SS:cdfRatiosUB_sigma1:additive}
                \]
                Next, since $\lambda_2\ge 0$ and $\Phi(\cdot;1)$ is monotone increasing,
                \[
                    \Phi(\lambda_2;1)\ge \Phi(0;1)=\frac{1}{2}
                    \qquadand 
                    \Phi(\lambda_1;1)\ge \Phi(\lambda_2;1)\ge \frac{1}{2}\,.
                    \yesnum\label{eq:SS:cdfRatiosUB_sigma1:half}
                \]
                Dividing \eqref{eq:SS:cdfRatiosUB_sigma1:additive} by $\Phi(\lambda_2;1)$ and using
                \eqref{eq:SS:cdfRatiosUB_sigma1:half} gives
                \[
                    1
                    \le
                    \frac{\Phi(\lambda_1;1)}{\Phi(\lambda_2;1)}
                    \le
                    1 + 2\cdot \alpha\cdot \frac{1}{\sqrt{2\pi}}e^{-\lambda_2^2/2}
                    =
                    1 + \sqrt{\frac{2}{\pi}}\alpha\,e^{-\lambda_2^2/2}\,.
                \]
                Similarly, dividing \eqref{eq:SS:cdfRatiosUB_sigma1:additive} by $\Phi(\lambda_1;1)$ and using
                \eqref{eq:SS:cdfRatiosUB_sigma1:half} yields
                \[
                    1 - \sqrt{\frac{2}{\pi}}\alpha\,e^{-\lambda_2^2/2}
                    \le
                    \frac{\Phi(\lambda_2;1)}{\Phi(\lambda_1;1)}
                    \leq 1\,.
                \]
                Combining the two displays implies
                \[
                    \frac{\Phi(\lambda_1;1)}{\Phi(\lambda_2;1)}\,,~
                    \frac{\Phi(\lambda_2;1)}{\Phi(\lambda_1;1)}
                    ~\in~
                    1 \pm \sqrt{\frac{2}{\pi}}\alpha\,e^{-\lambda_2^2/2}\,,
                \]
                completing the proof.
            \end{proof}

    \subsection{Proof of \cref{fact:SS:LC:boundOnTruncatedMoments} (Upper Bounds on Moments of Truncated Gaussian with One-Sided Truncation)}
        \label{sec:proofof:fact:SS:LC:boundOnTruncatedMoments}
        In this section, we prove \cref{fact:SS:LC:boundOnTruncatedMoments}, which we restate below.
        \factboundOnTruncatedMoments*
        \begin{proof}[Proof of \cref{fact:SS:LC:boundOnTruncatedMoments}]
            \citet{orjebin2014recursive} provide the following formulas for the second and fourth moments of the normal distribution truncated to $(-\infty,b]$:
            \begin{align*}
                \Ex_{z\sim \normal{\mu}{1}}\insquare{z^2\mid z\leq b}
                    &= \mu^2 + 1 - \frac{(\mu+b)\,\phi(b-\mu)}{\Phi(b-\mu)}\,,
                    \yesnum\label{eq:SS:LC:momentBound2}\\
                \Ex_{z\sim \normal{\mu}{1}}\insquare{z^{\smash{4}}\mid z\leq b}
                    &=\mu^4 + 6\mu^2 + 3 - \frac{\bigl(b^3+b^2\mu+b\mu^2+3b+5\mu+\mu^3\bigr)\,\phi(b-\mu)}{\Phi(b-\mu)}\,.
                    \yesnum\label{eq:SS:LC:momentBound4}
            \end{align*} 
            Where $\phi(\cdot)$ and $\Phi(\cdot)$ are the probability density function and the cumulative density function of the standard normal distribution.

            Since $\nfrac{\phi(z)}{\Phi(z)}\leq \abs{z}+1$ for $z\in \R$ (see \cref{fact:BoundOnMillsLikeRatio}),  \cref{eq:SS:LC:momentBound2} implies,
            \[
                \Ex_{z\sim \normal{\mu}{1}}\insquare{z^{{2}}\mid z\leq b} 
                \leq \mu^2+1 + \inparen{\abs{b}+\abs{\mu}}\,\inparen{1+\abs{b}+\abs{\mu}}
                \leq 2\inparen{1+\abs{b}+\abs{\mu}}^2 
                \,.
            \] 
            It remains to upper bound $\Ex_{z\sim \normal{\mu}{1}}\insquare{z^{{4}}\mid z\leq b}$.
            To simplify the notation, define,
            \[
                M \coloneqq |b|+|\mu|\,.
            \]
            Since $\left|b^3+b^2\mu+b\mu^2+\mu^3\right|\leq M^3$, $|3b+5\mu|\leq 5M$, and $\nfrac{\phi(z)}{\Phi(z)}\leq \abs{z}+1$ for $z\in \R$ (see \cref{fact:BoundOnMillsLikeRatio}),  \cref{eq:SS:LC:momentBound4} implies,
            \[
                \Ex_{z\sim \normal{\mu}{1}}\insquare{z^{{4}}\mid z\leq b}
                \leq 
                    \mu^4+6\mu^2+3 
                    +
                    \inparen{
                        M^3
                        +
                        5 M
                    } 
                    \inparen{1 + M}
                \,.
            \]
            Using $\mu^4+6\mu^2+3 \leq M^4 + 9 M^2+3$ and $(z^3+5z)(z+1) = z^4 + z^3 + 5z^2 + 5z$ implies 
            \[
                \Ex_{z\sim \normal{\mu}{1}}\insquare{z^{{4}}\mid z\leq b}
                \leq 
                M^4 + 9M^2 + (M^3+5M)(M+1)+3
                \leq 2M^4 + M^3 + 14M^2 + 5M+3\,.
            \]
            This implies the result since $M\geq 0$, $2z^4 + z^3 + 14z^2 + 5z+3\leq 3\inparen{1+z}^4$ for any $z\geq 0$.
        \end{proof}

        \subsubsection*{Proof of Upper Bound on $\sfrac{\phi(\cdot)}{\Phi(\cdot)}$}
        \begin{fact}\label{fact:BoundOnMillsLikeRatio}
            For $z\in \R$,
            \[
                \frac{\phi(z)}{\Phi(z)} \leq |z| + 1\,.
            \]
            Where $\phi(\cdot)$ and $\Phi(\cdot)$ are the probability density function and cumulative density function of the standard normal distribution.
        \end{fact}
        \begin{proof}
            We divide the proof into two cases.
            \begin{itemize}
                \item \textbf{Case A ($z\geq 0$):}~~
                    Since \(\Phi(z)\ge \Phi(0)=\nfrac{1}{2}\) for \(z\ge 0\), $\nfrac{\phi(z)}{\Phi(z)} \leq 2\phi(z)$.
                    Further, $\phi(z)=\inparen{\nfrac{1}{\sqrt{2\pi}}}e^{-z^2/2}$ is maximized at \(z=0\), so that for \(z\ge 0\), $2\phi(z) \leq 2\phi(0)=\sqrt{\sfrac{2}{\pi}} \leq 0.8\,.$
                    But we also have \(|z|+1=z+1\ge 1\) and, hence, $\nfrac{\phi(z)}{\Phi(z)} \leq 0.8 \leq z+1 = |z|+1$.
                \item \textbf{Case B ($z < 0$):}~~
                    Set \(w=-z>0\). By the symmetry of the standard normal density, we have
                    \[
                        \frac{\phi(z)}{\Phi(z)} = \frac{\phi(-w)}{\Phi(-w)} = \frac{\phi(w)}{1-\Phi(w)}\,.
                    \]
                    Define the function
                    \[
                        g(w) = (w+1)(1-\Phi(w)) - \phi(w)\,,\quad w>0\,.
                    \]
                    We claim that \(g(w)\ge 0\) for all \(w>0\): this holds because of the standard lower bound $1-\Phi(x)\geq \phi(x) \cdot \frac{2}{\sqrt{4+x^2}+x}$ for $x\geq 0$ (see  Equation (3) in \citet{duembgen2010boundingstandardgaussiantail}) and the fact that $z+1\geq \frac{1}{2}\sinparen{\sqrt{4+z^2}+z}$ for $z\geq 0$.
                    Thus, we conclude $(w+1)(1-\Phi(w)) \ge \phi(w)$ or equivalently,
                    \[
                    \frac{\phi(w)}{1-\Phi(w)} \leq w+1\,.
                    \]
                    Returning to the variable \(z\) (with \(w=-z\)), we deduce
                    \[
                    \frac{\phi(z)}{\Phi(z)} \leq |z| + 1\,.\qedhere
                    \]
            \end{itemize}
        \end{proof}

%% file: refs.bib
@misc{duembgen2010boundingstandardgaussiantail,
      title={Bounding Standard Gaussian Tail Probabilities}, 
      author={Lutz Duembgen},
      year={2010},
      eprint={1012.2063},
      archivePrefix={arXiv},
      primaryClass={math.ST},
      url={https://arxiv.org/abs/1012.2063}, 
}

@InProceedings{tai2023mixtureCensored,
  title = 	 {Learning Mixtures of Gaussians with Censored Data},
  author =       {Tai, Wai Ming and Aragam, Bryon},
  booktitle = 	 {Proceedings of the 40th International Conference on Machine Learning},
  pages = 	 {33396--33415},
  year = 	 {2023},
  editor = 	 {Krause, Andreas and Brunskill, Emma and Cho, Kyunghyun and Engelhardt, Barbara and Sabato, Sivan and Scarlett, Jonathan},
  volume = 	 {202},
  series = 	 {Proceedings of Machine Learning Research},
  month = 	 {7},
  publisher =    {PMLR},
  url = 	 {https://proceedings.mlr.press/v202/tai23a.html}, 
}

@article{nagarajan2023EMMixtureTruncation, title={Mean Estimation of Truncated Mixtures of Two Gaussians: A Gradient Based Approach}, volume={37}, url={https://ojs.aaai.org/index.php/AAAI/article/view/26110}, DOI={10.1609/aaai.v37i8.26110}, number={8}, journal={Proceedings of the AAAI Conference on Artificial Intelligence}, author={Nagarajan, Sai Ganesh and Palaiopanos, Gerasimos and Panageas, Ioannis and Vaidya, Tushar and Yu, Samson}, year={2023}, month={6}, pages={9260-9267} }

@InProceedings{nagarajan2020truncatedMixtureEM,
  title = 	 {On the Analysis of EM for Truncated Mixtures of Two Gaussians},
  author =       {Nagarajan, Sai Ganesh and Panageas, Ioannis},
  booktitle = 	 {Proceedings of the 31st International Conference  on Algorithmic Learning Theory},
  pages = 	 {634--659},
  year = 	 {2020},
  editor = 	 {Kontorovich, Aryeh and Neu, Gergely},
  volume = 	 {117},
  series = 	 {Proceedings of Machine Learning Research},
  month = 	 {2},
  publisher =    {PMLR},
  url = 	 {https://proceedings.mlr.press/v117/nagarajan20a.html}
}

@article{meilijson1981autopsy,
 ISSN = {00219002},
 URL = {http://www.jstor.org/stable/3213058},
 author = {Isaac Meilijson},
 journal = {Journal of Applied Probability},
 number = {4},
 pages = {829--838},
 publisher = {Applied Probability Trust},
 title = {Estimation of the Lifetime Distribution of the Parts from the Autopsy Statistics of the Machine},
 volume = {18},
 year = {1981}
}

@article{orjebin2014recursive,
  title={A Recursive Formula for the Moments of a Truncated Univariate Normal Distribution},
  author={Orjebin, Eric and Liquet, B. and Nazarathy, Y.},
  journal={Manuscript, The University of Queensland},
  year={2014},
  url={https://people.smp.uq.edu.au/YoniNazarathy/teaching_projects/studentWork/EricOrjebin_TruncatedNormalMoments.pdf}
}

@article{lee1978estimation,
    title = {Estimation of Some Limited Dependent Variable Models with Application to Housing Demand},
    journal = {Journal of Econometrics},
    volume = {8},
    number = {3},
    pages = {357-382},
    year = {1978},
    issn = {0304-4076},
    doi = {https://doi.org/10.1016/0304-4076(78)90052-0},
    url = {https://www.sciencedirect.com/science/article/pii/0304407678900520},
    author = {Lung-Fei Lee and Robert P. Trost}, 
}

@article{king1980tenure,
    title = {An Econometric Model of Tenure Choice and Demand for Housing as a Joint Decision},
    journal = {Journal of Public Economics},
    volume = {14},
    number = {2},
    pages = {137-159},
    year = {1980},
    issn = {0047-2727},
    doi = {https://doi.org/10.1016/0047-2727(80)90038-9},
    url = {https://www.sciencedirect.com/science/article/pii/0047272780900389},
    author = {Mervyn A. King}, 
}

@article{rosen1979housing,
    title = {Housing Decisions and the U.S. Income Tax: An Econometric Analysis},
    journal = {Journal of Public Economics},
    volume = {11},
    number = {1},
    pages = {1-23},
    year = {1979},
    issn = {0047-2727},
    doi = {https://doi.org/10.1016/0047-2727(79)90042-2},
    url = {https://www.sciencedirect.com/science/article/pii/0047272779900422},
    author = {Harvey S. Rosen}, 
}

@article{kenny1979college,
 ISSN = {00206598, 14682354},
 URL = {http://www.jstor.org/stable/2526272},
 author = {Lawrence W. Kenny and Lung-Fei Lee and G. S. Maddala and R. P. Trost},
 journal = {International Economic Review},
 number = {3},
 pages = {775--789},
 publisher = {[Economics Department of the University of Pennsylvania, Wiley, Institute of Social and Economic Research, Osaka University]},
 title = {Returns to College Education: An Investigation of Self-Selection Bias Based on the Project Talent Data},
 volume = {20},
 year = {1979}
}

@article{griliches1978missing,
 ISSN = {00190209},
 URL = {http://www.jstor.org/stable/20075289}, 
 author = {Zvi Griliches and Bronwyn H. Hall and Jerry A. Hausman},
 journal = {Annales de l'inséé},
 number = {30/31},
 pages = {137--176},
 publisher = {[GENES, ADRES]},
 title = {Missing Data and Self-Selection in Large Panels},
 year = {1978}
}

@article{lee1978union,
 ISSN = {00206598, 14682354},
 URL = {http://www.jstor.org/stable/2526310},
 author = {Lung-Fei Lee},
 journal = {International Economic Review},
 number = {2},
 pages = {415--433},
 publisher = {[Economics Department of the University of Pennsylvania, Wiley, Institute of Social and Economic Research, Osaka University]},
 title = {Unionism and Wage Rates: A Simultaneous Equations Model with Qualitative and Limited Dependent Variables},
 volume = {19},
 year = {1978}
}

@article{abowd1982job,
    author = {John M. Abowd and Henry S. Farber},
    title ={Job Queues and the Union Status of Workers},
    journal = {ILR Review},
    volume = {35},
    number = {3},
    pages = {354-367},
    year = {1982},
    doi = {10.1177/001979398203500305},
    URL = {https://doi.org/10.1177/001979398203500305},
    eprint = {https://doi.org/10.1177/001979398203500305} 
}

@article{leung1997censoring,
   author = "Leung, Kwan-Moon and Elashoff, Robert M. and Afifi, Abdelmonem A.",
  title={Censoring Issues in Survival Analysis},
   journal= "Annual Review of Public Health",
   year = "1997",
   volume = "18",
   number = "Volume 18, 1997",
   pages = "83-104",
   doi = "https://doi.org/10.1146/annurev.publhealth.18.1.83",
   url = "https://www.annualreviews.org/content/journals/10.1146/annurev.publhealth.18.1.83",
   publisher = "Annual Reviews",
   issn = "1545-2093",
   type = "Journal Article",
  }

@article{nakosteen1980migration,
 ISSN = {00384038},
 URL = {http://www.jstor.org/stable/1057152},
 author = {Robert A. Nakosteen and Michael Zimmer},
 journal = {Southern Economic Journal},
 number = {3},
 pages = {840--851},
 publisher = {Southern Economic Association},
 title = {Migration and Income: The Question of Self-Selection},
 volume = {46},
 year = {1980}
}

@article{harge2004convex,
  title={A Convex/Log-Concave Correlation Inequality for Gaussian Measure and an Application to Abstract Wiener Spaces},
  author={Harg{\'e}, Gilles},
  journal={Probability theory and related fields},
  volume={130},
  pages={415--440},
  year={2004},
  publisher={Springer}
}

@incollection{athey2007nonparametric,
title = {Chapter 60 Nonparametric Approaches to Auctions},
editor = {James J. Heckman and Edward E. Leamer},
series = {Handbook of Econometrics},
publisher = {Elsevier},
volume = {6},
pages = {3847-3965},
year = {2007},
issn = {1573-4412},
doi = {https://doi.org/10.1016/S1573-4412(07)06060-6},
url = {https://www.sciencedirect.com/science/article/pii/S1573441207060606},
author = {Susan Athey and Philip A. Haile}
}

@article{kim2024max,
author = {Kim, Seonho and Lee, Kiryung},
title = {Max-Affine Regression via First-Order Methods},
journal = {SIAM Journal on Mathematics of Data Science},
volume = {6},
number = {2},
pages = {534-552},
year = {2024},
doi = {10.1137/23M1594662},

URL = { 
    
        https://doi.org/10.1137/23M1594662
    
    

},
eprint = { 
    
        https://doi.org/10.1137/23M1594662
    
    

}
}

@article{olsen1982distributional,
 ISSN = {00206598, 14682354},
 URL = {http://www.jstor.org/stable/2526473},
 author = {Randall J. Olsen},
 journal = {International Economic Review},
 number = {1},
 pages = {223--240},
 publisher = {[Economics Department of the University of Pennsylvania, Wiley, Institute of Social and Economic Research, Osaka University]},
 title = {Distributional Tests for Selectivity Bias and a More Robust Likelihood Estimator},
 volume = {23},
 year = {1982}
}

@techreport{borjas1987selfSelection,
 title = "Self-Selection and the Earnings of Immigrants",
 author = "Borjas, George J.",
 institution = "National Bureau of Economic Research",
 type = "Working Paper",
 series = "Working Paper Series",
 number = "2248",
 year = "1987",
 doi = {10.3386/w2248},
 URL = "http://www.nber.org/papers/w2248",
}

@article{heckman1974shadow,
 ISSN = {00129682, 14680262},
 URL = {http://www.jstor.org/stable/1913937},
 author = {James Heckman},
 journal = {Econometrica},
 number = {4},
 pages = {679--694},
 publisher = {[Wiley, Econometric Society]},
 title = {Shadow Prices, Market Wages, and Labor Supply},
 volume = {42},
 year = {1974}
}

@article{heckman1979sample,
 ISSN = {00129682, 14680262},
 URL = {http://www.jstor.org/stable/1912352},
 author = {James J. Heckman},
 journal = {Econometrica},
 number = {1},
 pages = {153--161},
 publisher = {[Wiley, Econometric Society]},
 title = {Sample Selection Bias as a Specification Error},
 volume = {47},
 year = {1979}
}

@book{hanoch1976multivariate,
    author="Hanoch, Giora",
    title="A Multivariate Model of Labor Supply: Methodology for Estimation",
    address="Santa Monica, CA",
    year="1976",
    doi="",
    url={https://www.rand.org/pubs/reports/R1869.html#citation},
    publisher="RAND Corporation"
}

@inbook{Hanoch1980,
url = {https://doi.org/10.1515/9781400856992.119},
title = {Chapter 3. Hours And Weeks In The Theory Of Labor Supply},
booktitle = {Female Labor Supply},
booktitle = {Theory and Estimation},
author = {Giora Hanoch},
publisher = {Princeton University Press},
address = {Princeton},
pages = {119--165},
doi = {doi:10.1515/9781400856992.119},
isbn = {9781400856992},
year = {1980},
lastchecked = {2024-10-22}
}

@inbook{Cogan1980,
url = {https://doi.org/10.1515/9781400856992.327},
title = {Chapter 7 Labor Supply With Costs Of Labor Market Entry},
booktitle = {Female Labor Supply},
booktitle = {Theory and Estimation},
author = {John Cogan},
publisher = {Princeton University Press},
address = {Princeton},
pages = {327--364},
doi = {doi:10.1515/9781400856992.327},
isbn = {9781400856992},
year = {1980},
lastchecked = {2024-10-22}
}

@article{NELSON1977309,
title = {Censored Regression Models with Unobserved, Stochastic Censoring Thresholds},
journal = {Journal of Econometrics},
volume = {6},
number = {3},
pages = {309-327},
year = {1977},
issn = {0304-4076},
doi = {https://doi.org/10.1016/0304-4076(77)90003-3},
url = {https://www.sciencedirect.com/science/article/pii/0304407677900033},
author = {Forrest D. Nelson}
}

@ARTICLE{ghosh2022maxaffine,
  author={Ghosh, Avishek and Pananjady, Ashwin and Guntuboyina, Adityanand and Ramchandran, Kannan},
  journal={IEEE Transactions on Information Theory}, 
  title={Max-Affine Regression: Parameter Estimation for Gaussian Designs}, 
  year={2022},
  volume={68},
  number={3},
  pages={1851-1885},
  doi={10.1109/TIT.2021.3130717}}

@InProceedings{silvia2020EMmixtures,
  title = 	 {EM Converges for a Mixture of Many Linear Regressions},
  author =       {Kwon, Jeongyeol and Caramanis, Constantine},
  booktitle = 	 {Proceedings of the Twenty Third International Conference on Artificial Intelligence and Statistics},
  pages = 	 {1727--1736},
  year = 	 {2020},
  editor = 	 {Chiappa, Silvia and Calandra, Roberto},
  volume = 	 {108},
  series = 	 {Proceedings of Machine Learning Research},
  publisher =    {PMLR},
  url = 	 {https://proceedings.mlr.press/v108/kwon20a.html}
}

@article{hannon1999estimation,
	author = {Patrick M. Hannon and Ram C. Dahiya},
	doi = {10.1080/03610929908832440},
	journal = {Communications in Statistics - Theory and Methods},
	number = {11},
	pages = {2591--2612},
	publisher = {Taylor \& Francis},
	title={Estimation of Parameters for the Truncated Exponential Distribution},
	volume = {28},
	year = {1999},
}

@article{heitjan1990inference,
author = {Daniel F. Heitjan and Donald B. Rubin},
title = {Inference from Coarse Data via Multiple Imputation with Application to Age Heaping},
journal = {Journal of the American Statistical Association},
volume = {85},
number = {410},
pages = {304--314},
year = {1990},
publisher = {ASA Website},
doi = {10.1080/01621459.1990.10476202},


URL = { 
    
    
        https://www.tandfonline.com/doi/abs/10.1080/01621459.1990.10476202
    

},
eprint = { 
    
    
        https://www.tandfonline.com/doi/pdf/10.1080/01621459.1990.10476202
    

}
}

@article{fisher31,
  title={{Properties and Applications of HH Functions}},
  author={Fisher, Ronald A.},
  journal={Mathematical Tables},
  volume={1},
  pages={815--852},
  year={1931}
}

@article{Lee1915,
 ISSN = {00063444},
 URL = {http://www.jstor.org/stable/2331782},
 author = {Alice Lee},
 journal = {Biometrika},
 number = {2/3},
 pages = {208--214},
 publisher = {[Oxford University Press, Biometrika Trust]},
 title = {Table of the Gaussian `Tail' Functions; When the `Tail' is Larger than the Body},
 volume = {10},
 year = {1914}
}

@book{hernan2023causal,
  title={{Causal Inference: What If}},
  author={Hernan, Miguel A. and Robins, James M.},
  isbn={9781315374932},
  lccn={2022050840},
  series={Chapman \& Hall/CRC Monographs on Statistics \& Applied Probability},
  url={https://books.google.co.in/books?id=FPkN0AEACAAJ},
  year={2023},
  publisher={Taylor \& Francis}
}

@article{van1996efficient,
author = {{van der Laan}, Mark J.},
title = {{Efficient Estimation in the Bivariate Censoring Model and Repairing NPMLE}},
volume = {24},
journal = {The Annals of Statistics},
number = {2},
publisher = {Institute of Mathematical Statistics},
pages = {596 -- 627},
year = {1996},
doi = {10.1214/aos/1032894454},
URL = {https://doi.org/10.1214/aos/1032894454}
}

@inproceedings{hardt2016strategic,
author = {Hardt, Moritz and Megiddo, Nimrod and Papadimitriou, Christos and Wootters, Mary},
title = {Strategic Classification},
year = {2016},
isbn = {9781450340571},
publisher = {Association for Computing Machinery},
address = {New York, NY, USA},
url = {https://doi.org/10.1145/2840728.2840730},
doi = {10.1145/2840728.2840730},
booktitle = {Proceedings of the 2016 ACM Conference on Innovations in Theoretical Computer Science},
pages = {111–122},
numpages = {12},
location = {Cambridge, Massachusetts, USA},
series = {ITCS '16}
}

@article{athey2002identification,
  title={Identification of standard auction models},
  author={Athey, Susan and Haile, Philip A},
  journal={Econometrica},
  volume={70},
  number={6},
  pages={2107--2140},
  year={2002},
  publisher={Wiley Online Library}
}

@article{heckman1990varieties,
 ISSN = {00028282},
 URL = {http://www.jstor.org/stable/2006591},
 author = {James J. Heckman},
 journal = {The American Economic Review},
 number = {2},
 pages = {313--318},
 publisher = {American Economic Association},
 title = {Varieties of Selection Bias},
 volume = {80},
 year = {1990}
}

@article{fair1972methods,
 ISSN = {00129682, 14680262},
 URL = {http://www.jstor.org/stable/1913181},
 author = {Ray C. Fair and Dwight M. Jaffee},
 journal = {Econometrica},
 number = {3},
 pages = {497--514},
 publisher = {[Wiley, Econometric Society]},
 title = {Methods of Estimation for Markets in Disequilibrium},
 volume = {40},
 year = {1972}
}

@article{willis1979education,
  title={Education and Self-selection},
  author={Willis, Robert J and Rosen, Sherwin},
  journal={Journal of political Economy},
  volume={87},
  number={5, Part 2},
  pages={S7--S36},
  year={1979},
  publisher={The University of Chicago Press}
}

@article{kleinberg2020classifiers,
author = {Kleinberg, Jon and Raghavan, Manish},
title = {How Do Classifiers Induce Agents to Invest Effort Strategically?},
year = {2020},
issue_date = {November 2020},
publisher = {Association for Computing Machinery},
address = {New York, NY, USA},
volume = {8},
number = {4},
issn = {2167-8375},
url = {https://doi.org/10.1145/3417742},
doi = {10.1145/3417742},
journal = {ACM Trans. Econ. Comput.},
month = oct,
articleno = {19},
numpages = {23}
}

@inproceedings{dong2018strategic,
author = {Dong, Jinshuo and Roth, Aaron and Schutzman, Zachary and Waggoner, Bo and Wu, Zhiwei Steven},
title = {Strategic Classification from Revealed Preferences},
year = {2018},
isbn = {9781450358293},
publisher = {Association for Computing Machinery},
address = {New York, NY, USA},
url = {https://doi.org/10.1145/3219166.3219193},
doi = {10.1145/3219166.3219193},
booktitle = {Proceedings of the 2018 ACM Conference on Economics and Computation},
pages = {55–70},
numpages = {16},
location = {Ithaca, NY, USA},
series = {EC '18}
}

@article{grunwald2003updating,
  title={Updating Probabilities},
  author={Grunwald, Peter D. and Halpern, Joseph Y.},
  journal={Journal of Artificial Intelligence Research},
  volume={19},
  pages={243--278},
  year={2003},
  doi={https://doi.org/10.1613/jair.1164}
}

@article{jacobsen1995coarsening,
author = {Martin Jacobsen and Niels Keiding},
title = {{Coarsening at Random in General Sample Spaces and Random Censoring in Continuous Time}},
volume = {23},
journal = {The Annals of Statistics},
number = {3},
publisher = {Institute of Mathematical Statistics},
pages = {774 -- 786},
year = {1995},
doi = {10.1214/aos/1176324622},
URL = {https://doi.org/10.1214/aos/1176324622}
}

@Inbook{robins1992recovery,
author="Robins, James M.
and Rotnitzky, Andrea",
editor="Jewell, Nicholas P.
and Dietz, Klaus
and Farewell, Vernon T.",
title="Recovery of Information and Adjustment for Dependent Censoring Using Surrogate Markers",
bookTitle="AIDS Epidemiology: Methodological Issues",
year="1992",
publisher="Birkh{\"a}user Boston",
address="Boston, MA",
pages="297--331",
isbn="978-1-4757-1229-2",
doi="10.1007/978-1-4757-1229-2_14",
url="https://doi.org/10.1007/978-1-4757-1229-2_14"
}

@InProceedings{gill1997coarsening,
    title={Coarsening at Random: Characterizations, Conjectures, Counter-Examples},
    author={Gill, Richard D. and Van Der Laan, Mark J. and Robins, James M.},
    editor="Lin, D. Y.
    and Fleming, T. R.", 
    booktitle="Proceedings of the First Seattle Symposium in Biostatistics",
    year="1997",
    publisher="Springer US",
    address="New York, NY",
    pages="255--294",
    isbn="978-1-4684-6316-3"
}

@inproceedings{truncatedDiscrete,
  title={Learning Hard-Constrained Models with One Sample},
  author={Galanis, Andreas and Kalavasis, Alkis and Kandiros, Anthimos Vardis},
  booktitle={Proceedings of the 2024 Annual ACM-SIAM Symposium on Discrete Algorithms (SODA)},
  pages={3184--3196},
  url={https://arxiv.org/pdf/2311.03332},
  year={2024},
  organization={SIAM}
}

@article{gostic2020practical,
    doi = {10.1371/journal.pcbi.1008409},
    author = {Gostic, Katelyn M. AND McGough, Lauren AND Baskerville, Edward B. AND Abbott, Sam AND Joshi, Keya AND Tedijanto, Christine AND Kahn, Rebecca AND Niehus, Rene AND Hay, James A. AND De Salazar, Pablo M. AND Hellewell, Joel AND Meakin, Sophie AND Munday, James D. AND Bosse, Nikos I. AND Sherrat, Katharine AND Thompson, Robin N. AND White, Laura F. AND Huisman, Jana S. AND Scire, Jérémie AND Bonhoeffer, Sebastian AND Stadler, Tanja AND Wallinga, Jacco AND Funk, Sebastian AND Lipsitch, Marc AND Cobey, Sarah},
    journal = {PLOS Computational Biology},
    publisher = {Public Library of Science},
    title={Practical Considerations for Measuring the Effective Reproductive Number, $R_t$},
    year = {2020},
    month = {12},
    volume = {16},
    url = {https://doi.org/10.1371/journal.pcbi.1008409},
    pages = {1-21},
    number = {12},

}

@book{little2019statistical,
  title={Statistical Analysis with Missing Data},
  author={Little, Roderick J.A. and Rubin, Donald B.},
  volume={793},
  year={2019},
  doi={10.1002/9781119482260},
  publisher={John Wiley \& Sons}
}

@article{little1989analysis,
  author={Little, Roderick J.A. and Rubin, Donald B.},
title ={The Analysis of Social Science Data with Missing Values},

journal = {Sociological Methods \& Research},
volume = {18},
number = {2-3},
pages = {292-326},
year = {1989},
doi = {10.1177/0049124189018002004},

URL = { 
    
        https://doi.org/10.1177/0049124189018002004
    
    

},
eprint = { 
    
        https://doi.org/10.1177/0049124189018002004
    
    

}
}

@article{rubin1976inference,
      title={Inference and Missing Data},
      author={Rubin, Donald B.},
	doi = {10.1093/biomet/63.3.581},
	eprint = {https://academic.oup.com/biomet/article-pdf/63/3/581/756166/63-3-581.pdf},
	issn = {0006-3444},
	journal = {Biometrika},
	month = {12},
	number = {3},
	pages = {581-592},
	url = {https://doi.org/10.1093/biomet/63.3.581},
	volume = {63},
	year = {1976}}

@article{heitjan1991ignorability,
 ISSN = {00905364, 21688966},
 URL = {http://www.jstor.org/stable/2241929},
 author = {Daniel F. Heitjan and Donald B. Rubin},
 journal = {The Annals of Statistics},
 number = {4},
 pages = {2244--2253},
 publisher = {Institute of Mathematical Statistics},
 title = {Ignorability and Coarse Data},
 volume = {19},
 year = {1991}
}

@article{heitjan1993ignorability,
 ISSN = {0006341X, 15410420},
 URL = {http://www.jstor.org/stable/2532251},
 author = {Daniel F. Heitjan},
 journal = {Biometrics},
 number = {4},
 pages = {1099--1109},
 publisher = {International Biometric Society},
 title = {Ignorability and Coarse Data: Some Biomedical Examples},
 volume = {49},
 year = {1993}
}

@article{fotakis2020efficient,
author = {Fotakis, Dimitris and Kalavasis, Alkis and Tzamos, Christos},
title = {Efficient Parameter Estimation of Truncated Boolean Product Distributions},
year = {2022},
issue_date = {Aug 2022},
publisher = {Springer-Verlag},
address = {Berlin, Heidelberg},
volume = {84},
number = {8},
issn = {0178-4617},
url = {https://doi.org/10.1007/s00453-022-00961-9},
doi = {10.1007/s00453-022-00961-9},
journal = {Algorithmica},
month = aug,
pages = {2186–2221},
numpages = {36}
}

@InProceedings{ilyas2020theoretical,
  title = 	 {A Theoretical and Practical Framework for Regression and Classification from Truncated Samples},
  author =       {Ilyas, Andrew and Zampetakis, Emmanouil and Daskalakis, Constantinos},
  booktitle = 	 {Proceedings of the Twenty Third International Conference on Artificial Intelligence and Statistics},
  pages = 	 {4463--4473},
  year = 	 {2020},
  editor = 	 {Chiappa, Silvia and Calandra, Roberto},
  volume = 	 {108},
  series = 	 {Proceedings of Machine Learning Research},
  month = 	 {8},
  publisher =    {PMLR},
  url = 	 {https://proceedings.mlr.press/v108/ilyas20a.html}
}

@InProceedings{daskalakis2021statistical,
  title = 	 {A Statistical Taylor Theorem and Extrapolation of Truncated Densities},
  author =       {Daskalakis, Constantinos and Kontonis, Vasilis and Tzamos, Christos and Zampetakis, Emmanouil},
  booktitle = 	 {Proceedings of Thirty Fourth Conference on Learning Theory},
  pages = 	 {1395--1398},
  year = 	 {2021},
  editor = 	 {Belkin, Mikhail and Kpotufe, Samory},
  volume = 	 {134},
  series = 	 {Proceedings of Machine Learning Research},
  month = 	 {8},
  publisher =    {PMLR},
  url = 	 {https://proceedings.mlr.press/v134/daskalakis21a.html}
}

@article{raschke2012inference,
	author = {Raschke, Mathias},
	date = {2012-01-01},
	doi = {10.1007/s00477-011-0458-8},
	id = {Raschke2012},
	isbn = {1436-3259},
	journal = {Stochastic Environmental Research and Risk Assessment},
	number = {1},
	pages = {127--138},
  title={Inference for the Truncated Exponential Distribution},
	url = {https://doi.org/10.1007/s00477-011-0458-8},
	volume = {26},
	year = {2012}}

@article{pearson1908generalised,
 ISSN = {00063444, 14643510},
 URL = {http://www.jstor.org/stable/2331556},
 author = {Karl Pearson and Alice Lee},
 journal = {Biometrika},
 number = {1},
 pages = {59--68},
 publisher = {[Oxford University Press, Biometrika Trust]},
 title = {On the Generalised Probable Error in Multiple Normal Correlation},
 volume = {6},
 year = {1908}
}

@article{pearson1902v,
  title={V. On the Methematical Theory of Errors of Judgement, with Special Reference to the Personal Equation},
  author={Pearson, Karl},
  journal={Philosophical Transactions of the Royal Society of London. Series A, Containing Papers of a Mathematical or Physical Character},
  volume={198},
  number={300-311},
  pages={235--299},
  year={1902},
  doi={https://doi.org/10.1098/rsta.1902.0005},
  url={https://royalsocietypublishing.org/doi/abs/10.1098/rsta.1902.0005},
  publisher={The Royal Society London}
}

@book{imbens2015causal,
  author={Imbens, Guido W. and Rubin, Donald B.},
	isbn = {9780521885881},
	lccn = {2014020988},
	publisher = {Cambridge University Press},
	series = {Causal Inference for Statistics, Social, and Biomedical Sciences: An Introduction},
	title = {Causal Inference in Statistics, Social, and Biomedical Sciences},
	url = {https://books.google.com/books?id=Bf1tBwAAQBAJ},
	year = {2015}}

@article{woodroofe1985estimating,
author = {Michael Woodroofe},
title = {{Estimating a Distribution Function with Truncated Data}},
volume = {13},
journal = {The Annals of Statistics},
number = {1},
publisher = {Institute of Mathematical Statistics},
pages = {163 -- 177},
year = {1985},
doi = {10.1214/aos/1176346584},
URL = {https://doi.org/10.1214/aos/1176346584}
}

@inproceedings{daskalakis2020truncated,
 author = {Daskalakis, Constantinos and Rohatgi, Dhruv and Zampetakis, Emmanouil},
 booktitle = {Advances in Neural Information Processing Systems},
 editor = {H. Larochelle and M. Ranzato and R. Hadsell and M.F. Balcan and H. Lin},
 pages = {10338--10347},
 publisher = {Curran Associates, Inc.},
 title = {Truncated Linear Regression in High Dimensions},
 url = {https://proceedings.neurips.cc/paper_files/paper/2020/file/751f6b6b02bf39c41025f3bcfd9948ad-Paper.pdf},
 volume = {33},
 year = {2020}
}

@inbook{de2023testing, 
	author = {Anindya De and Shivam Nadimpalli and Rocco A. Servedio},
	booktitle = {Proceedings of the 2023 Annual ACM-SIAM Symposium on Discrete Algorithms (SODA)},
	doi = {10.1137/1.9781611977554.ch155},
	pages = {4050-4082},
    year={2023},
	title = {Testing Convex Truncation},
}

@INPROCEEDINGS{lee2024unknown,
  author={Lee, Jane H. and Mehrotra, Anay and Zampetakis, Manolis},
  booktitle={2024 IEEE 65th Annual Symposium on Foundations of Computer Science (FOCS)}, 
  title={Efficient Statistics With Unknown Truncation, Polynomial Time Algorithms, Beyond Gaussians}, 
  year={2024},
  volume={},
  number={},
  pages={988-1006},
  doi={10.1109/FOCS61266.2024.00066}}

@article{truncated_sm,
  author  = {Song Liu and Takafumi Kanamori and Daniel J. Williams},
  title   = {Estimating Density Models with Truncation Boundaries using Score Matching},
  journal = {Journal of Machine Learning Research},
  year    = {2022},
  volume  = {23},
  number  = {186},
  pages   = {1--38},
  url     = {http://jmlr.org/papers/v23/21-0218.html}
}

@inproceedings{trunc_regression_unknown_var,
 author = {Daskalakis, Constantinos and Stefanou, Patroklos and Yao, Rui and Zampetakis, Emmanouil},
 booktitle = {Advances in Neural Information Processing Systems},
 editor = {M. Ranzato and A. Beygelzimer and Y. Dauphin and P.S. Liang and J. Wortman Vaughan},
 pages = {1952--1963},
 publisher = {Curran Associates, Inc.},
 title = {Efficient Truncated Linear Regression with Unknown Noise Variance},
 url = {https://proceedings.neurips.cc/paper_files/paper/2021/file/0ed8861dc36bee580d100f91283d0559-Paper.pdf},
 volume = {34},
 year = {2021}
}

@InProceedings{diakonikolas2024statistical,
  title = 	 {Statistical Query Lower Bounds for Learning Truncated Gaussians},
  author =       {Diakonikolas, Ilias and Kane, Daniel M. and Pittas, Thanasis and Zarifis, Nikos},
  booktitle = 	 {Proceedings of Thirty Seventh Conference on Learning Theory},
  pages = 	 {1336--1363},
  year = 	 {2024},
  editor = 	 {Agrawal, Shipra and Roth, Aaron},
  volume = 	 {247},
  series = 	 {Proceedings of Machine Learning Research},
  month = 	 {6},
  publisher =    {PMLR},
  url = 	 {https://proceedings.mlr.press/v247/diakonikolas24b.html}
}

@InProceedings{plevrakis2021censored,
  title = 	 {{Learning From Censored and Dependent Data: The Case of Linear Dynamics}},
  author =       {Plevrakis, Orestis},
  booktitle = 	 {Proceedings of Thirty Fourth Conference on Learning Theory},
  pages = 	 {3771--3787},
  year = 	 {2021},
  editor = 	 {Belkin, Mikhail and Kpotufe, Samory},
  volume = 	 {134},
  series = 	 {Proceedings of Machine Learning Research},
  month = 	 {8},
  publisher =    {PMLR},
  url = 	 {https://proceedings.mlr.press/v134/plevrakis21a.html},
}

@inproceedings{cherapanamjeri2022auctionEstimation,
author = {Cherapanamjeri, Yeshwanth and Daskalakis, Constantinos and Ilyas, Andrew and Zampetakis, Manolis},
title = {Estimation of Standard Auction Models},
year = {2022},
isbn = {9781450391504},
publisher = {Association for Computing Machinery},
address = {New York, NY, USA},
url = {https://doi.org/10.1145/3490486.3538284},
doi = {10.1145/3490486.3538284},
booktitle = {Proceedings of the 23rd ACM Conference on Economics and Computation},
pages = {602–603},
numpages = {2},
location = {Boulder, CO, USA},
series = {EC '22}
}

@article{roy1951earnings,
	author = {Roy, Andrew D.},
	doi = {10.1093/oxfordjournals.oep.a041827},
	eprint = {https://academic.oup.com/oep/article-pdf/3/2/135/6966554/3-2-135.pdf},
	issn = {0030-7653},
	journal = {Oxford Economic Papers},
	month = {06},
	number = {2},
	pages = {135-146},
	title = {Some Thoughts on the Distribution of Earnings},
	url = {https://doi.org/10.1093/oxfordjournals.oep.a041827},
	volume = {3},
	year = {1951}}

@book{Cohen91,
	author = {Cohen, A. Clifford},
	isbn = {9781420052114},
	publisher = {Taylor \& Francis},
	series = {Statistics: A Series of Textbooks and Monographs},
	title = {Truncated and Censored Samples: Theory and Applications},
	url = {https://books.google.com/books?id=ZRs86JTMdogC},
	year = {1991}}

@misc{gaitonde2024selfselection,
      title={Sample-Efficient Linear Regression with Self-Selection Bias}, 
      author={Jason Gaitonde and Elchanan Mossel},
      year={2024},
      eprint={2402.14229},
      archivePrefix={arXiv},
      primaryClass={math.ST},
      url={https://arxiv.org/abs/2402.14229}, 
}

@InProceedings{fotakis2021coarse,
  title = 	 {{Efficient Algorithms for Learning from Coarse Labels}},
  author =       {Fotakis, Dimitris and Kalavasis, Alkis and Kontonis, Vasilis and Tzamos, Christos},
  booktitle = 	 {Proceedings of Thirty Fourth Conference on Learning Theory},
  pages = 	 {2060--2079},
  year = 	 {2021},
  editor = 	 {Belkin, Mikhail and Kpotufe, Samory},
  volume = 	 {134},
  series = 	 {Proceedings of Machine Learning Research},
  publisher =    {PMLR},
  url = 	 {https://proceedings.mlr.press/v134/fotakis21a.html}, 
}

@book{maddala1986limited,
	author = {Maddala, Gangadharrao S.},
	isbn = {9780521338257},
	lccn = {82009554},
	publisher = {Cambridge University Press},
	series = {Econometric Society Monographs},
	title = {Limited-Dependent and Qualitative Variables in Econometrics},
	url = {https://books.google.com/books?id=-Ji1ZaUg7gcC},
	year = {1983}}

@article{Galton1897,
  title={An Examination Into the Registered Speeds of American Trotting Horses, With Remarks on Their Value as Hereditary Data},
  author={Galton, Francis},
  doi = {http://doi.org/10.1098/rspl.1897.0115},
  journal={Proceedings of the Royal Society of London},
  volume={62},
  number={379-387},
  pages={310--315},
  year={1897},
  publisher={The Royal Society}
}

@inbook{canonne2020learning,
author = {Clément L. Canonne and Anindya De and Rocco A. Servedio},
title = {Learning from satisfying assignments under continuous distributions},
booktitle = {Proceedings of the 2020 ACM-SIAM Symposium on Discrete Algorithms (SODA)},
year = {2020},
chapter = {},
pages = {82-101},
doi = {10.1137/1.9781611975994.6},
URL = {https://epubs.siam.org/doi/abs/10.1137/1.9781611975994.6},
eprint = {https://epubs.siam.org/doi/pdf/10.1137/1.9781611975994.6}
}

@inproceedings{de2024detecting,
author = {De, Anindya and Li, Huan and Nadimpalli, Shivam and Servedio, Rocco A.},
title = {Detecting Low-Degree Truncation},
year = {2024},
isbn = {9798400703836},
publisher = {Association for Computing Machinery},
address = {New York, NY, USA},
url = {https://doi.org/10.1145/3618260.3649633},
doi = {10.1145/3618260.3649633},
booktitle = {Proceedings of the 56th Annual ACM Symposium on Theory of Computing},
pages = {1027–1038},
numpages = {12},
location = {Vancouver, BC, Canada},
series = {STOC 2024}
}

@inproceedings{cherapanamjeri2023selfselection,
    author = {Cherapanamjeri, Yeshwanth and Daskalakis, Constantinos and Ilyas, Andrew and Zampetakis, Manolis},
    title = {{What Makes a Good Fisherman? Linear Regression under Self-Selection Bias}},
    year = {2023},
    isbn = {9781450399135},
    publisher = {Association for Computing Machinery},
    address = {New York, NY, USA},
    url = {https://doi.org/10.1145/3564246.3585177},
    doi = {10.1145/3564246.3585177}, 
    booktitle = {Proceedings of the 55th Annual ACM Symposium on Theory of Computing},
    pages = {1699–1712},
    numpages = {14},
    location = {Orlando, FL, USA},
    series = {STOC 2023}
}

@INPROCEEDINGS{daskalakis2018efficient,
  author={Daskalakis, Constantinos and Gouleakis, Themis and Tzamos, Christos and Zampetakis, Manolis},
  booktitle={2018 IEEE 59th Annual Symposium on Foundations of Computer Science (FOCS)}, 
  title={{Efficient Statistics, in High Dimensions, from Truncated Samples}}, 
  year={2018},
  volume={},
  number={},
  pages={639-649},
  doi={10.1109/FOCS.2018.00067}}

@InProceedings{daskalakis2019computationally,
  title = 	 {{Computationally and Statistically Efficient Truncated Regression}},
  author =       {Daskalakis, Constantinos and Gouleakis, Themis and Tzamos, Christos and Zampetakis, Manolis},
  booktitle = 	 {Proceedings of the Thirty-Second Conference on Learning Theory},
  pages = 	 {955--960},
  year = 	 {2019},
  editor = 	 {Beygelzimer, Alina and Hsu, Daniel},
  volume = 	 {99},
  series = 	 {Proceedings of Machine Learning Research},
  publisher =    {PMLR},
  url = 	 {https://proceedings.mlr.press/v99/daskalakis19a.html},
}

@article{Kontonis2019EfficientTS,
  title={{Efficient Truncated Statistics with Unknown Truncation}},
  author={Vasilis Kontonis and Christos Tzamos and Manolis Zampetakis},
  journal={2019 IEEE 60th Annual Symposium on Foundations of Computer Science (FOCS)},
  year={2019},
  pages={1578-1595},
  url={https://api.semanticscholar.org/CorpusID:199442432}
}

@inproceedings{
    lee2023learning,
    title={{Learning Exponential Families from Truncated Samples}},
    author={Jane H. Lee and Andre Wibisono and Manolis Zampetakis},
    booktitle={Thirty-seventh Conference on Neural Information Processing Systems},
    year={2023},
    url={https://openreview.net/forum?id=PxcWJqO3qj}
}

@misc{garrigos2023handbook,
      title={Handbook of Convergence Theorems for (Stochastic) Gradient Methods}, 
      author={Guillaume Garrigos and Robert M. Gower},
      year={2023},
      eprint={2301.11235},
      archivePrefix={arXiv},
      primaryClass={math.OC}
}

@article{xu2019accelerate,
author = {Xu, Yi and Lin, Qihang and Yang, Tianbao},
title = {Accelerate Stochastic Subgradient Method by Leveraging Local Growth Condition},
journal = {Analysis and Applications},
volume = {17},
number = {05},
pages = {773-818},
year = {2019},
doi = {10.1142/S0219530519400050},

URL = { 
    
        https://doi.org/10.1142/S0219530519400050
    
    

},
eprint = { 
    
        https://doi.org/10.1142/S0219530519400050
    
    

}
}

@article{yang2018beating,
  author       = {Tianbao Yang and
                  Qihang Lin},
  title        = {{RSG:} Beating Subgradient Method without Smoothness and Strong Convexity},
  journal      = {J. Mach. Learn. Res.},
  volume       = {19},
  pages        = {6:1--6:33},
  year         = {2018},
  url          = {https://jmlr.org/papers/v19/17-016.html},
  bibsource    = {dblp computer science bibliography, https://dblp.org}
}

@article{chevillard2012erf,
  author       = {Sylvain Chevillard},
  title        = {The Functions erf and erfc Computed with Arbitrary Precision and Explicit Error Bounds},
  journal      = {Inf. Comput.},
  volume       = {216},
  pages        = {72--95},
  year         = {2012},
  url          = {https://doi.org/10.1016/j.ic.2011.09.001},
  doi          = {10.1016/J.IC.2011.09.001},
  bibsource    = {dblp computer science bibliography, https://dblp.org}
}

@book{wainwright2019high,
  title={High-dimensional statistics: A non-asymptotic viewpoint},
  author={Wainwright, Martin J},
  volume={48},
  year={2019},
  publisher={Cambridge university press}
}
